\documentclass[preprint,10pt,numbers]{elsarticle}
\usepackage{amssymb}
\usepackage{amsmath,bm,natbib,graphics,mathrsfs,graphicx,epsfig,placeins,indentfirst,setspace,enumerate,enumitem,caption,varioref,booktabs,tabularx,color,epstopdf,multirow}
\usepackage{booktabs}
\usepackage{siunitx}
\usepackage{caption}
\usepackage{nicefrac}
\usepackage{bm}
\usepackage{subfigure}

\usepackage{tikz}
\usepackage{float}
\newlength{\fboxw}
\usetikzlibrary{shapes.geometric, arrows.meta, positioning, calc, decorations.pathreplacing, intersections, backgrounds, fit}

\usepackage[utf8]{inputenc}
\usepackage{amsfonts}
\usepackage{listings}
\usepackage{xcolor}
\lstdefinestyle{pystyle}{
    language=Python,
    frame=lines,
    framerule=0.5pt,
    basicstyle=\small\ttfamily,
    keywordstyle=\color{blue},
    commentstyle=\color{gray},
    stringstyle=\color{teal},
    showstringspaces=false,
    breaklines=true,
    tabsize=4
}
\usepackage{algorithm}
\usepackage{algpseudocode}
\usepackage{soul}


\usepackage[colorlinks=true,
            linkcolor=blue,
            citecolor=blue,
            urlcolor=blue]{hyperref}
\makeatletter \oddsidemargin  -.1in \evensidemargin -.1in

\journal{Nuclear Physics B}
\begin{document}
\newcommand{\bea}{\begin{eqnarray}}
		\newcommand{\eea}{\end{eqnarray}}
	\newcommand{\nn}{\nonumber}
	\newcommand{\bee}{\begin{eqnarray*}}
		\newcommand{\eee}{\end{eqnarray*}}
	\newcommand{\lb}{\label}
	\newcommand{\nii}{\noindent}
	
	\newcommand{\ii}{\indent}
	\newtheorem{theorem}{Theorem}[section]
	\newtheorem{example}{Example}[section]
	\newtheorem{corollary}{Corollary}[section]
	\newtheorem{definition}{Definition}[section]
	\newtheorem{lemma}{Lemma}[section]
	
	\newtheorem{remark}{Remark}[section]
	\newtheorem{proposition}{Proposition}[section]
	\numberwithin{equation}{section}
	\renewcommand{\theequation}{\thesection.\arabic{equation}}
    \newenvironment{proof}
    {\noindent\textbf{Proof. }}
    {\hfill$\square$\par}
\begin{frontmatter}

\title{\bf QGPINNs: A Physics-Informed Neural Network Framework for Nonlocal Differential Equations on Quantum Graphs}

\author[1]{Vaibhav~{\bf Mehandiratta}\corref{cor1}}
\ead{vaibhavm@goa.bits-pilani.ac.in}

\author[1]{Saket~{\bf Ramchandra}}
\ead{saketrprofessional@gmail.com}

\cortext[cor1]{Corresponding author}

\affiliation[1]{organization={Department of Mathematics}, 
            addressline={Birla Institute of Technology and Science, Pilani, K K Birla Goa Campus}, 
            city={Zuarinagar, Sancoale},
            state={Goa 403726},
            country={India}}

\begin{abstract}
We propose QGPINNs, a physics-informed neural network framework developed in PyTorch~\cite{paszke2019pytorch} for the numerical solution of nonlocal differential equations on quantum graphs. The framework is designed as a general computational implementation in which the solution on each edge of the graph is approximated by a neural network, while a unified graph-based loss function enforces the governing equations together with initial, boundary, and vertex transmission conditions. In particular, the formulation incorporates standard continuity and Kirchhoff-Neumann vertex conditions and Dirichlet boundary conditions into the learning process to couple the local edge-wise neural approximations into a global solution on the graph. The framework is developed for two representative classes of nonlinear models: multi-order fractional elliptic problems and time-fractional evolution equations on quantum graphs. To improve accuracy and training stability, QGPINNs integrates several graph-adapted learning strategies, including soft and hard constraint enforcement, dynamic loss balancing, Fourier feature embeddings, and a learnable singularity-capturing feature for weakly singular solutions arising in the considered problems. The framework also extends naturally to inverse problems, including the identification of the orders of fractional operators and physical parameters from noisy observational data.  We validate the accuracy, computational efficiency, and physical consistency of the proposed framework through numerical experiments on benchmark graph structures and real-world networks, including the IEEE 14-bus system and an open-channel agricultural drainage network.

 \end{abstract}

\end{frontmatter}
\section{Introduction}

In this paper, we establish a general framework based on the deep learning approach for solving a class of nonlocal differential equations defined on quantum graphs, and provide various examples and features of its usage and accuracy \cite{saketgithub}.\vspace{0.05cm}

Over the past few decades, differential equation models involving non-local operators have gained significant interest in the fields of applied mathematics, physics, engineering, and computational science primarily due to their ability to capture long-range (i.e., nonlocal) interactions and anomalous diffusion processes \cite{nochetto2015pde}, which cannot be adequately described by classical integer-order differential equations. Fractional differential operators, in particular, provide a powerful mathematical framework for modeling such nonlocal phenomena and have found applications in porous media flows, viscoelastic materials, biological systems transport in heterogeneous media, and many other areas. However, many complex systems encountered in modern science and engineering do not evolve on standard Euclidean domains. Instead, they are naturally defined on network-like structures, such as connected urban traffic systems in a city, branched river networks, biological neural networks, gas and power transmission systems, etc. Consequently, in recent years, research focus has shifted significantly toward the mathematical modeling, statistical analysis, and numerical solution of the differential equations posed directly on network domains \cite{Arioli2017FEM, kovacs2021stochastic, mehandiratta2021optimal, bolin2024gaussian, bolin2026statistical}. \vspace{0.05cm}

\begin{figure}
    \centering
    \includegraphics[width=1\linewidth]{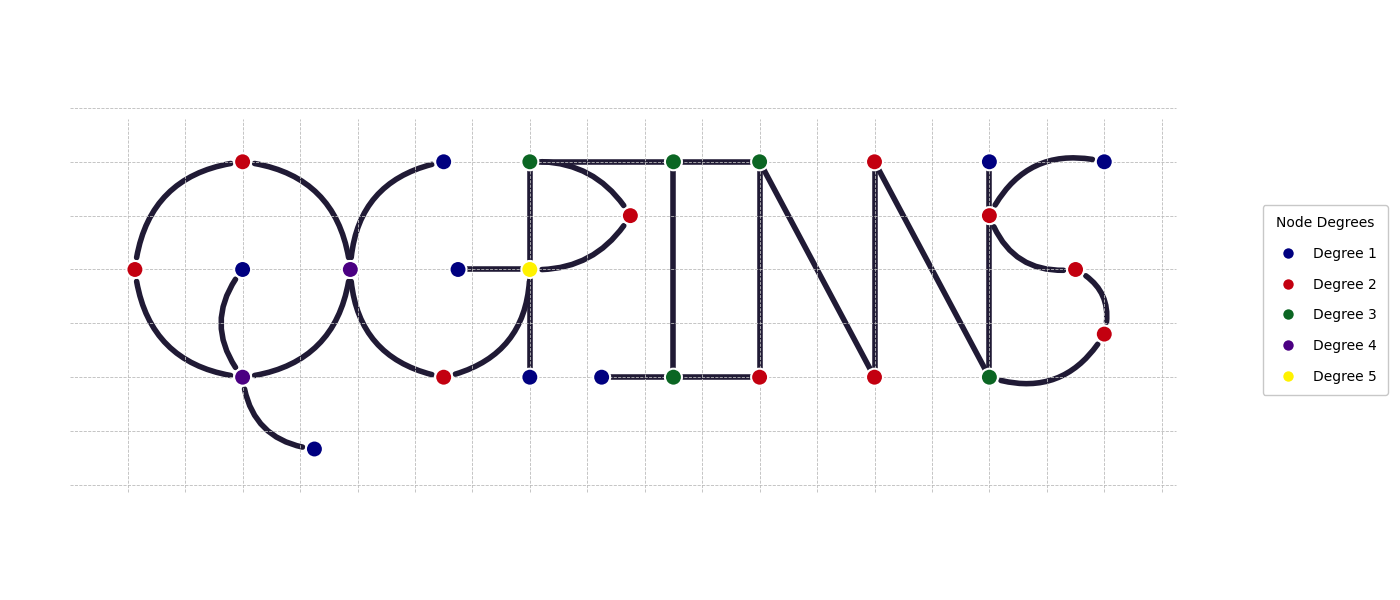}
    \caption {Quantum graph with the degree of each vertex labeled. Node colors indicate vertex degree.}
    \label{fig:QGPINNS}
\end{figure}

For such practical applications, the underlying spatial domains can be efficiently represented as compact metric graphs \cite{Berkolaiko2013}, where each edge is identified with an interval and is equipped with a metric structure. A quantum graph $\Gamma$ is a metric graph endowed with additional structures, namely, assigning the differential operator, associated function spaces, and vertex conditions that describe the interaction of the edge-wise dynamics at the junctions of the network. More precisely, we define the graph $\Gamma=(\mathcal{V}, \mathcal{E})$ consisting of a finite set of nodes or vertices $\mathcal{V}=\{v_i\}_{i=1}^{|\mathcal{V}|}$ and a set of edges $\mathcal{E}=\{e_i\}_{i=1}^{|\mathcal{E}|}$ connecting the vertices, such that each edge of the graph is identified by an interval $(0,\ell_e)$ with $0<\ell_e<\infty$. That is, on each edge $e$, connecting two vertices $v_r$ and $v_s$ ($v_r \rightarrow v_s$), we impose a coordinate $x$ that increases from $0$ to $\ell_e$ in such a way
that $x =0$ corresponds to vertex $v_r$, while $x =\ell_e$ corresponds to vertex $v_s$. Moreover, in a more general setting, the edges of the underlying graph may not be straight connections between vertices, but can be induced by any regular curve $r:[0,\ell_e]\to \mathbb{R}^{d}$, $d\in \mathbb{N}$, parametrized by arc-length. This geometric flexibility makes the quantum graph framework particularly suitable for modeling real-world networked structures, where the connections between vertices are naturally represented by continuous spatial curves rather than straight line segments. A visual representation of such a graph, the QGPINNs logo, is provided in Figure \ref{fig:QGPINNS}, along with the degree of each vertex, deg($v$), defined as the number of edges connected to it.\vspace{0.1cm}

Several theoretical works have been investigated for classical and, more recently, for fractional models on quantum graphs, including spectral properties of differential operators, well-posedness of the associated differential problems, controllability, and qualitative behaviour of solutions; see in particular \cite{von1988classical, lagnese1993control, band2017quantum, odvzak2019weyl, mehandiratta2023well} and references therein. Nevertheless, the literature on numerical methods for solving differential equation models on such domains remains relatively sparse. In particular, for fractional differential equation (FDE) models on quantum graphs, traditional finite difference and finite element methods have been employed in recent years \cite{mehandiratta2020difference, bolin2024regularity, kumari2025finite, bolin2026numerical}. However, these methods are restricted to discretization of each edge of the underlying graph together with the consistent treatment of the vertex conditions. Consequently, the traditional methods face computational bottlenecks and numerical instability, particularly for complex geometries and large-scale real networks such as those considered in this work. Moreover, for fractional or nonlocal models, the computational difficulty further increases as the underlying operators involve memory effects and long-range interactions. These challenges motivate us to look beyond the existing traditional numerical methods for solving nonlocal differential equations on real network structures modeled by quantum graphs.\vspace{0.1cm}

Deep learning methods, particularly physics-informed neural networks (PINNs), provide an alternative computational framework for solving differential equations. In this approach, the solution is approximated by a neural network, and the governing equations, together with physical constraints such as initial and boundary conditions, are incorporated into a loss function. The key effectiveness
of PINNs lies in employing automatic differentiation \cite{baydin2018automatic} to analytically derive the classical derivatives of neural network outputs. However, automatic differentiation cannot be directly employed for the fractional differential equation models due to the nonlocality of the fractional derivatives and the inapplicability of the classical chain
rule in fractional calculus. To address this issue, fractional PINNs and nonlocal PINNs have been proposed \cite{pang2019fpinns, guo2022monte, singh2024non}, where one utilises conventional numerical techniques to discretize fractional operators, and these resulting approximations are then integrated into the residual loss. We refer to \cite{sophiya2025comprehensive} for a nice comprehensive survey of PINNs with their different variants and applications. These existing nonlocal PINNs frameworks, however, are predominantly developed for problems defined on Euclidean domains and do not address the difficulties occur from graph topology and vertex conditions.

On the other hand, recent PINN-based approaches for differential equations on quantum graphs have mainly focused on integer-order models \cite{zhao2022deep, laczko2025transferable}. These works, in particular, shows that PINNs can be adapted to classical elliptic and evolution equations on graphs. However, to the best of our knowledge, a general PINN framework for nonlocal differential equations on quantum graphs has not yet been investigated. In particular, there is a need for a unified framework that simultaneously incorporates edge-wise nonlocal operators, graph topology, vertex transmission conditions, and neural-network-based training strategies. The main objective of the work is to describe and implement such a general computational framework. We propose QGPINNs, a unified physics-informed neural network framework for solving nonlocal differential equations on quantum graphs. The proposed framework is designed to handle both classical and nonlocal differential operators on quantum graphs while incorporating the vertex conditions that couple the edge-wise equations into a global graph problem. 

While we have striven to implement a general software framework for quantum graph computations using PINNs, the direction of development has been guided by two classes of nonlinear models, namely, fractional elliptic equations and time-fractional evolution equations on quantum graphs. 

\vspace{0.05cm}

\textbf{Model 1:} \textit{Multi-Order Fractional Elliptic Problem}:
\label{elip_prob_def}
The first class consists of elliptic-type fractional boundary value
problems involving two spatial fractional operators of different orders.
On each edge $e_i \in \mathcal{E}$, identified with the interval
$[0, \ell_{i}]$, we seek $u_{i} : [0, \ell_i] \to \mathbb{R}$ satisfying:
\begin{equation}
    {}_{0}^{C}D_{x}^{\gamma} u_{i}(x)
    =g\left(u_{i}(x),{}_{0}^{C}D_{x}^{\beta} u_{i}(x),\frac{du_{i}}{dx}\right) + f_{i}(x),
    \quad x \in (0, \ell_{i}),\quad i=1,2,\dotsc, \mathcal{E},
    \label{eq:fbvp}
\end{equation}
where $1 < \gamma \leq 2$, $0 < \beta \leq \gamma-1$, the function
$g : \mathbb{R}^3 \to \mathbb{R}$ encodes the possibly nonlinear reaction term, and $f_{i}$ is a prescribed forcing function on the edge $e_i$.\vspace{0.1cm}

\textbf{Model 2:} \textit{Time-Fractional Evolution Equation}:
\label{Para_problem_def}
The second class consists of nonlocal evolution equations in which the classical first-order time derivative is replaced by a Caputo fractional derivative of order $\gamma \in (0,1)$.
On each edge $e_i \in \mathcal{E}$, we seek
$u_{i} : [0, \ell_{i}] \times (0,T] \to \mathbb{R}$ satisfying:
\begin{equation}
    {}_{0}^{C}D_{t}^{\gamma}\, u_{i}(x,t)
    =  \mathcal{N}\left(u_{i}(x,t),\frac{\partial u_{i}}{\partial x}(x,t),\frac{\partial^2 u_{i}}{\partial x^2}(x,t)\right) +f_{i}(x,t), \quad (x,t)\in (0,\ell_i)\times (0,T),\quad i=1,2,\dotsc, \mathcal{E},
    \label{eq:parabolic}
\end{equation}
subject to the initial condition
\begin{equation*}
    u_{i}(x, 0) = u_{i}^{(0)}(x), \qquad x \in [0, \ell_{i}],
\end{equation*}
where $\mathcal{N}$ is a nonlinear differential operator and $f_{i}$ is the source term on edge $e_i$. For a sufficiently smooth function $h$, the nonlocal operator ${}_{0}^{C}D_{t}^{\gamma}$ of order $\gamma>0$ is defined by
\begin{equation}\label{caputodef}
    {}_{0}^{C}D_{t}^{\gamma} h(t)
    = \frac{1}{\Gamma(n-\gamma)}
      \int_{0}^{t} (t-\tau)^{n-\gamma-1} h^{(n)}(\tau)\, d\tau,\quad t\in [0,T],
\end{equation}
where $n-1<\gamma\leq n$, $n\in \mathbb{N}$, and $\Gamma(\cdot)$ denotes the Euler gamma function. The spatial fractional operators ${}_{0}^{C}D_{x}^{\gamma}$ and ${}_{0}^{C}D_{x}^{\beta}$ in \eqref{eq:fbvp} are defined analogously.\vspace{0.1cm}

The above problems, containing both classical (local) and fractional (nonlocal) operators, are defined as a collection of equations on each edge of the graph. To define the global problem on the underlying graph $\Gamma$, we equipped the edge-wise equations with appropriate boundary conditions at the vertices, known as the vertex conditions in the context of quantum graphs. To this end, for a vertex $v\in \mathcal{V}$, let $E_v$ denote the set of all edges $e$ incident to the vertex $v$. The vertex condition consists of\vspace{0.05cm}\\
\textit{Continuity condition:}
\begin{equation}\label{cont}
u_i(v)=u_j(v)~\forall e_i,e_j\in E_v.
\end{equation}
\textit{Kirchhoff–Neumann condition:}
\begin{equation}\label{kn}
\sum_{e_i\in E_{v}}\partial u_i(v)=0,
\end{equation}
where $\partial u_i(v)$ denotes the directional derivative of $u_i$ at the vertex $v$ taken in the direction pointing away from the vertex. That is, for an edge $e_i=[0,\ell_i]$, connecting the vertices $v_r \rightarrow v_s$, one has $\partial u_i(v_r)=\partial u_i(0)=u_i'(0)$, and $u_i(v_s)=\partial u_i(\ell_i)=-u_i'(\ell_i)$. The above condition \eqref{kn} is the natural extension of Neumann boundary conditions in the graph setting, as, in particular, for the boundary vertex, i.e., a vertex $v$ with deg($v$)=1, this reduces to the standard Neumann boundary condition. Moreover, this is the standard vertex condition for diffusion-type operators. For a general advection-diffusion or transport models, the derivative term in \eqref{kn} may be replaced by the corresponding problem dependent flux which leads to a generalized Kirchhoff condition at the junction \cite{berkolaiko2013introduction}. Such a flux-based junction condition has been  used in the open-channel flow example in Section~\ref{sec:p4_drain}. Finally, the \textit{Dirichlet vertex condition} is
\begin{equation}\label{dc}
u(v)\equiv u_i(v)=g_v,
\end{equation}
which we impose only at boundary vertices as an alternative to the corresponding Neumann condition there. If one imposes the Dirichlet condition \eqref{dc} for each vertex $v\in \mathcal{V}$, then it would completely decouple the edge-wise problems and essentially remove any connection between the edges adjacent to the vertex. Consequently, in this case, the topology of the graph becomes irrelevant, and the global problem on $\Gamma$ reduces identically to a collection of independent problems posed locally on individual edges.

\begin{figure}[htbp]
\centering
\begin{tikzpicture}[
  scale=0.5,
  vertex/.style={circle, draw, fill=white, minimum size=7pt, inner sep=0pt, thick},
  hub/.style={circle, draw, fill=black!80, minimum size=9pt, inner sep=0pt, thick},
  pendant/.style={circle, draw, fill=violet!30, minimum size=7pt, inner sep=0pt, thick},
  lbl/.style={font=\small},
  anno/.style={font=\scriptsize, align=center},
]

\node[hub]     (v0) at (0,0)   {};
\node[pendant] (v1) at (6, 0) {};
\node[pendant] (v2) at (5.36, 2.68) {};
\node[pendant] (v3) at (5.36, -2.68) {};

\draw[thick] (v0) -- node[above, lbl]{$e_1=[0,\ell_1]$} (v1);
\draw[thick] (v0) -- node[above left, lbl]{$e_2=[0,\ell_2]$} (v2);
\draw[thick] (v0) -- node[below left, lbl]{$e_3=[0,\ell_3]$} (v3);

\node[above=2pt of v0, lbl] {$v_0$};
\node[right=2pt        of v1, lbl] {$v_1$};
\node[above=2pt         of v2, lbl] {$v_2$};
\node[below=2pt         of v3, lbl] {$v_3$};

\node[anno, right=14pt of v1]
  {Dirichlet \\ $u_1(v_1)=0$};
\node[anno, above right=6pt and 4pt of v2]
  {Neumann \\ $\partial u_2(v_2)=0$};
\node[anno, below right=6pt and 4pt of v3]
  {Neumann \\ $\partial u_3(v_3)=0$};

\node[anno, draw, thin, inner sep=6pt, left=34pt of v0, align=left] (interior_box)
  {\textbf{Interior Vertex}\\\textbf{Continuity}:$u_1(v_0)=u_2(v_0)=u_3(v_0)$ \\[4pt]
   \textbf{Kirchhoff-Neumann}\\
   $\displaystyle\sum_{i=1}^{3}\partial u_i(v_0)=0$\\
   with $\partial u_i(v_0)=u'_i(0)$};

\draw[->, gray, thin] (interior_box.east) -- (-0.18, 0);

\end{tikzpicture}
\caption{A star graph $\mathcal{G}$ with interior vertex $v_0$ (filled) and
         pendant vertices $v_1, v_2, v_3$ (shaded).
At $v_0$, the solution must satisfy both the continuity condition
         (equal trace from all incident edges) and the Kirchhoff flux
         condition (outward derivatives sum to zero).}
\label{fig:star-graph}
\end{figure}
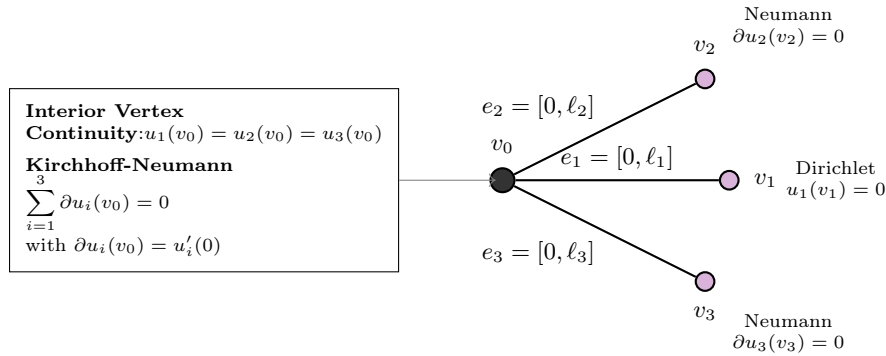

The equations \eqref{eq:fbvp} and \eqref{eq:parabolic}, together with the above-defined vertex conditions, define the fractional elliptic and the time-fractional evolution equations on quantum graphs, respectively, where, in particular, for the problem \eqref{eq:parabolic}, the vertex conditions \eqref{cont}-\eqref{dc} are imposed for every $t\in (0,T)$. In recent years, these models, particularly time-fractional diffusion equations, have gained significant interest in modeling anomalous and memory-dependent processes on network structures and have been studied from both theoretical and computational aspects; see, for instance, \cite{mehandiratta2019existence, mehandiratta2020difference, sobirov2021green, faheem2023collocation, wang2024local, vats2026time}. However, existing numerical methods for such models have been predominantly based on traditional discretization techniques, which, as mentioned above, are computationally costly for large-scale networks, as it requires the discretization of both the governing equations on each edge and the vertex conditions at the graph junctions. Therefore, these models require an efficient computational framework that is capable of handling local differentiation, nonlocal memory effects, and graph-based coupling in a unified manner.\vspace{0.08cm}

In the proposed QGPINNs framework, the solution on each edge of the graph is represented by a neural network, and the total loss function is then constructed via the residuals of the governing equations, initial conditions, boundary conditions, and vertex conditions. The local derivatives in the model are computed using automatic differentiation, whereas the nonlocal fractional derivatives are approximated using suitable numerical schemes, particularly the $L1$ and $L2-1_\sigma$ approximation schemes \cite{lin2007finite, alikhanov2015new}, which are then integrated into the resulting residual loss derived for our framework.  This strategy allows QGPINNs to preserve the flexibility of PINNs while accurately accounting for the nonlocal structure of fractional operators. In addition, the framework describes and implements various architectural and training features designed to improve accuracy and robustness. These include enforcement of hard and soft constraints, dynamic balancing of different loss components, Fourier feature embedding for the efficient representation of oscillatory solutions, and a singularity-capturing feature for the weak singular behaviour of the solutions to fractional-order models. These features provide substantial performance gains compared with standard PINNs, without introducing significant overhead in terms of runtime or memory. Moreover, in nonlocal models, the order of the operator is a key parameter because it determines the
strength and range of nonlocal interactions. Therefore, our framework also incorporates parameter estimation, particularly the estimation of the order of the fractional operators, from noisy observed data.

The main contributions of this paper are summarized as follows: 
\begin{itemize} 
\item We propose QGPINNs, a unified physics-informed neural network framework for solving local and nonlocal differential equations on quantum graphs.

\item We formulate a unified loss function that incorporates edge-wise equation residuals together with Dirichlet and Neumann boundary conditions and continuity and Kirchhoff-Neumann vertex transmission conditions.

\item We introduce and implement several training and architectural enhancements, including hard and soft constraints,  dynamic balancing of different loss components, Fourier feature embeddings, and learnable singularity features. 

\item We demonstrate the applicability of the proposed framework to both forward and inverse problems, including identification of the fractional-orders and physical parameters from noisy observations. 

\item We validate the accuracy, computational efficiency and physical consistency of the framework on several typical graph structures, including real-world network such as the IEEE 14-bus system and an open-channel agricultural drainage network. 
\end{itemize}

The rest of the paper is organised as follows. In section 2, we discuss important mathematical preliminaries including the approximation schemes used for the considered fractional operators. Section 3 introduces the QGPINNs framework, and , in particular, defines the loss function for both forward and inverse problems, the enforcement of vertex conditions, and various optimization strategies that significantly improve performance. Section 4 discusses the implementation details of the proposed framework.
In Section 5, we demonstrate the accuracy and the effectiveness of the proposed framework using numerical examples on different quantum graph structures. Finally, the conclusion of the paper is provided in Section 6.

\section{Preliminaries}

In this section, we briefly present the key mathematical tools used in the QGPINNs framework. We first describe the standard physics-informed neural network formulation for differential equation models and then provide the approximation of Caputo fractional derivatives, required for constructing the residual losses associated with nonlocal operators.
\subsection{Physics-informed neural networks}\label{pinnsection}

A neural network is a mathematical model consisting of different layers of neurons, including an input layer, several hidden layers, and an output layer, which are interconnected through nonlinear activation functions.

If \(z^0=z\in \mathbb{R}^{n_0}\) denotes the input layer, then the hidden layers are defined recursively by
\[
    z^k=\sigma\left(W^k z^{k-1}+b^k\right),
    \qquad k=1,2,\ldots,L-1,
\]
and the output layer is given by
\[
    z^L=W^L z^{L-1}+b^L\in \mathbb{R}^{n_L}.
\]
Here, \(W^k\in \mathbb{R}^{n_k\times n_{k-1}}\) and \(b^k\in \mathbb{R}^{n_k}\) denote the weight matrix and bias vector at the \(k\)-th layer, respectively, $n_k$ is the number of neurons in the $k^{th}$ layer, and \(\sigma\) is a nonlinear activation function. The collection of all trainable weights and biases is denoted by $\bm{\theta}=\{W^k,b^k\}_{k=1}^{L}$
and the neural network approximation of the unknown solution is then written as
\[
    \widehat{u}(z;\bm{\theta})=z^L.
\]

A loss function is associated with this neural network output, which essentially quantifies the discrepancies between the network prediction and the expected outcome. The underlying network is trained using backpropagation \cite{rumelhart1986learning}, together with a suitable optimization algorithm, to minimize the resulting loss function and update the trainable parameters $\bm{\theta}$.

Physics-informed neural networks (PINNs) form a well-known class of neural network methods in which the underlying physics of the system is incorporated into the training process through the loss function. In particular, the neural network is trained not only from data but also from the governing differential equations and the associated initial and boundary conditions. To introduce the basic idea, consider the general differential problem
\[
\begin{cases}
    \mathcal{L}[u](z)=f(z), & z\in \Omega,\\
    \mathcal{B}[u](z)=g(z), & z\in \partial\Omega,
\end{cases}
\]
where \(\Omega\subset \mathbb{R}^{n}\) is the spatial domain, \(\mathcal{L}[\cdot]\) represents the governing differential operator, and \(\mathcal{B}[\cdot]\) denotes the boundary operator corresponding to the boundary conditions. For time-dependent problems, an initial condition
\[
    \mathcal{I}[u](z,0)=u_0(z),~~z\in \Omega
\]
is also prescribed.

Given the neural-network approximation \(\widehat{u}(z;\bm{\theta})\), let \(\{z_r^j\}_{j=1}^{N_r}\subset\Omega\) and \(\{z_b^j\}_{j=1}^{N_b}\subset\partial\Omega\) denote the residual and boundary collocation points, respectively. Then,  

the residual and boundary losses are defined by
\[
    \mathcal{L}_{r}(\bm{\theta})
    =
    \frac{1}{N_r}
    \sum_{j=1}^{N_r}
    \left(
    \mathcal{L}[\hat{u}](z_r^j;\bm{\theta})-f(z_r^{j})
    \right)^2,
\]
and
\[
    \mathcal{L}_{b}(\bm{\theta})
    =
    \frac{1}{N_b}
    \sum_{j=1}^{N_b}
    \left(
    \mathcal{B}[\widehat{u}](z_b^j;\bm{\theta})-g(z_b^j)
    \right)^2,
\]
respectively. For time-dependent problems, the initial condition is enforced through
\[
    \mathcal{L}_{0}(\bm{\theta})
    =
    \frac{1}{N_0}
    \sum_{j=1}^{N_0}
    \left(
    \widehat{u}(z_0^j,0;\bm{\theta})-u_0(z_0^j)
    \right)^2,
\]
where \(\{z_0^j\}_{j=1}^{N_0}\subset\Omega\) denotes the set of collocation points corresponding to the initial conditions. The total loss function for PINN is then written as
\[
    \mathcal{L}(\bm{\theta})
    =
    \lambda_r \mathcal{L}_{r}(\bm{\theta})
    +
    \lambda_b \mathcal{L}_{b}(\bm{\theta})
    +
    \lambda_0 \mathcal{L}_{0}(\bm{\theta}),
\]
where \(\lambda_r,\lambda_b,\lambda_0>0\) are weights used to balance the different loss components. The trainable parameters are obtained by solving
\[
    \bm{\theta}^\ast
    =
    \operatorname*{arg\,min}_{\bm{\theta}}
    \mathcal{L}(\bm{\theta}).
\]
The trained function \(\widehat{u}(\cdot;\bm{\theta}^\ast)\) is then considered as the PINN approximation of the solution. 

During the optimization process, the derivatives of \(\widehat{u}\) appearing in the residual loss are evaluated using automatic differentiation for integer-order differential operator. However, for non-local operators $\mathcal{L}$, suitable numerical schemes are employed to approximate such operators, and the resulting approximations are incorporated into the residual loss. In both cases, the key objective of PINN framework is to determine the optimal parameters that minimize a loss function containing the governing equations, boundary conditions, initial conditions, and any available observational data.

In the proposed framework, the above standard PINN formulation has been employed edge-wise on a quantum graph. More precisely, for each edge \(e_i\in\mathcal{E}\), identified with an interval $[0,\ell_i]$, the underlying solution $u_i$ is approximated by a neural network \(\hat{u}_{i,\bm{\theta}}\), and the differential residual is evaluated separately on that edge. The edge losses are then coupled through additional vertex-condition loss functions associated with continuity, Kirchhoff-Neumann conditions, and boundary conditions at the vertices. The complete graph-based loss formulation is presented in Section~\ref{main}.

\subsection{Approximation of non-local operators}\label{nonlocapprox}
As mentioned above, the integer-order derivatives in the considered problems are computed via automatic differentiation. However, the non-local nature of the Caputo fractional derivative requires the use of suitable approximation methods to handle such operators within the neural network optimization.

While PINNs inherently provide a generalized continuous approximate solution over the entire computational domain, the training of the model is executed on a finite set of collocation points. Therefore, the choice and distribution of these data points play an important role in the accuracy and convergence efficiency of the PINN framework. Furthermore, the considered nonlocal problems do not possess a smooth solution throughout the closed domain and often exhibit weak singularities \cite{stynes2017error}. Such singularities present a significant challenge for a PINN to approximate the solution when using a standard uniform discretization mesh.

To tackle this, nonuniform meshes, in particular, graded meshes, are commonly employed. 

Specifically, let $N$ be a positive integer, then the temporal graded mesh is defined by

\begin{equation}\label{graded}
    t_n = T\!\left(\frac{n}{N}\right)^{\!r}, \quad n = 0, 1, \dots, N,
\end{equation}
where $r\geq 1$ is the mesh grading parameter chosen by the user. The graded mesh on each spatial edge $e_i=[0,\ell_i]$ is defined analogously. In particular, for $r=1$, the graded mesh reduces to a uniform mesh.

\begin{figure}[htbp]
\centering
\begin{tikzpicture}[x=0.85\textwidth, y=1cm]

\def\N{12}

\draw[->, gray!50, thick] (0, 2) -- (1.08, 2);
\node[left, font=\small] at (0, 2) {$r=1$};
\foreach \n in {0,1,...,12} {
  \pgfmathsetmacro\x{\n/\N}
  \fill[violet!70] (\x, 2) circle (2pt);
}

\draw[->, gray!50, thick] (0, 1) -- (1.08, 1);
\node[left, font=\small] at (0, 1) {$r=2$};
\foreach \n in {0,1,...,12} {
  \pgfmathsetmacro\x{(\n/\N)^2}
  \fill[teal!80!black] (\x, 1) circle (2pt);
}

\draw[->, gray!50, thick] (0, 0) -- (1.08, 0);
\node[left, font=\small] at (0, 0) {$r=4$};
\foreach \n in {0,1,...,12} {
  \pgfmathsetmacro\x{(\n/\N)^4}
  \fill[orange!80!red] (\x, 0) circle (2pt);
}

\node[below, font=\small] at (0,   0) {$0$};
\node[below, font=\small] at (1.0, 0) {$T$};
\node[below, font=\small] at (0.5,-0.45) {$t$};

\draw[gray, dashed, thin] (0, 2.3) -- (0, 2.6)
                          (1, 2.3) -- (1, 2.6);
\draw[gray, thin] (0, 2.6) -- (1, 2.6)
  node[midway, above, font=\scriptsize] {uniform spacing};
\draw[gray, dashed, thin] (0, -0.3) -- (0, -0.6)
                          (0.083, -0.3) -- (0.083, -0.6);
\draw[gray, thin] (0, -0.6) -- (0.083, -0.6)
  node[midway, below, font=\scriptsize] {dense near $t=0$};
\end{tikzpicture}
\caption{Graded mesh with $N=12$ points on time interval $[0,T]$. Higher $r$ clusters points near $t=0$, providing finer resolution of the initial singularity region.}
\label{fig:graded-mesh}
\end{figure}
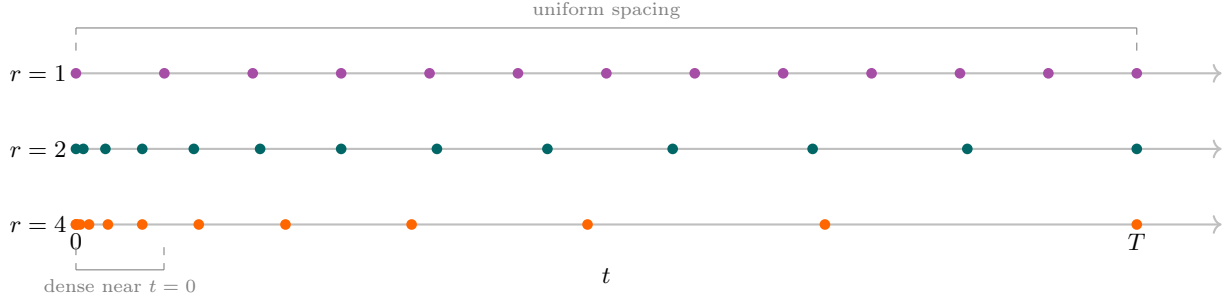

We shall employ two different discretizations, namely, the $L1$ scheme and $L2-1_{\sigma}$of scheme, for the approximation for the Caputo fractional derivatives given in \eqref{eq:fbvp} and \eqref{eq:parabolic}. In the following, we describe the approximation for the temporal Caputo fractional derivative; the corresponding spatial fractional derivative can be defined analogously on each edge.

\subsubsection{The L1 Scheme}

In view of the temporal graded mesh \eqref{graded} defined over the interval $(0,T]$, the Caputo fractional derivative \eqref{caputodef} of order $\gamma\in (0,1)$ at a mesh point $t=t_n$ can be written as

\begin{equation*}
    {}_{0}^{C}D_{t}^{\gamma} f(x, t_n)
    = \frac{1}{\Gamma(1-\gamma)} \sum_{k=0}^{n-1}
      \int_{t_k}^{t_{k+1}} (t_n - s)^{-\gamma}
      \frac{\partial f(x, s)}{\partial \tau}\, ds
\end{equation*}

Now, for $n\geq 1$, the well-known $L1$ formula for approximating Caputo fractional derivative is expressed as \cite{lin2007finite}

\begin{equation}\label{L1scheme}
\begin{aligned}
    {}_{0}^{C}D_{t}^{\gamma} f(x, t_n)
    &\approx \frac{1}{\Gamma(1-\gamma)} \sum_{k=0}^{n-1}
    \frac{f(x, t_{k+1}) - f(x, t_k)}{\tau_{k+1}}
    \int_{t_k}^{t_{k+1}} (t_n - s)^{-\gamma}\, ds\\
    &=  \frac{1}{\Gamma(2-\gamma)} \sum_{k=0}^{n-1}  \frac{f(x, t_{k+1}) - f(x, t_k)}{\tau_{k+1}} [(t_n - t_k)^{1-\gamma} - (t_n - t_{k+1})^{1-\gamma}]\\
    &= \frac{1}{\Gamma{(2-\gamma)}}[w_{n,1}f(x,t_n)-w_{n,n}f(x,0)-\sum_{k=1}^{n-1} f(x,t_{n-k})(w_{n, k}-w_{n,k+1})],
\end{aligned}
\end{equation}
where $\tau_k=t_k-t_{k-1}$ and 
\begin{equation}\label{weightl1}
    w_{n,k}:= \frac{(t_n - t_{n-k})^{1-\gamma} - (t_n - t_{n-k+1})^{1-\gamma}}
        {\tau_{n-k+1}},~~\text{for}~k=1,2,\dotsc,n.
\end{equation}
Based on the error analysis provided in \cite{stynes2017error}, the above $L1$ approximation \eqref{L1scheme} on a graded mesh satisfies the convergence estimate $\mathcal{O}(\tau^{\min (r\gamma,2-\gamma)})$, where $\tau=\displaystyle\max_{1\leq n\leq N}\tau_n$. Consequently, the optimal grading mesh parameter for achieving the maximum rate $\mathcal{O}(\tau^{2-\gamma})$ is $r=\frac{2-\gamma}{\gamma}$.

\subsubsection{The L2-1$_\sigma$ Scheme}

We now present a higher-order approximation scheme developed by Alikhanov~\cite{alikhanov2015new}, namely, the L2-1$\sigma$ scheme on nonuniform mesh.

The L2-1$_\sigma$ scheme improves on the L1 scheme by employing piecewise quadratic interpolation.
To achieve higher accuracy, the fractional derivative is evaluated at the shifted point
\[
t_{n-\sigma}=\sigma t_{n-1}+(1-\sigma)t_n,
\qquad \sigma=1-\frac{\gamma}{2},
\qquad n=1,2,\ldots,N .
\] First, we define the following local kernel components:

\begin{align*}
& A_{0}^{n}=\frac{1}{\tau_n \Gamma(1-\gamma)}\int_{t_{n-1}}^{t_{n-\sigma}} (t_{n-\sigma} - s)^{-\gamma}\, ds =\frac{(t_{n-\sigma}-t_{n-1})^{1-\gamma}}{\tau_n\Gamma{(2-\gamma)}}=\frac{(1-\sigma)^{1-\gamma}\tau_{n}^{-\gamma}}{\Gamma{(2-\gamma)}},\\
&A_{n-k}^n = \frac{1}{\tau_k \Gamma(1-\gamma)}
    \int_{t_{k-1}}^{t_k} (t_{n-\sigma} - s)^{-\gamma}\, ds=\frac{(t_{n-\sigma}-t_{k-1})^{1-\gamma}-(t_{n-\sigma}-t_k)^{1-\gamma}}{\tau_k \Gamma{(2-\gamma)}},\\
 &B_{n-k}^n= \frac{2}{\tau_k(t_{k+1} - t_{k-1})\Gamma(1-\gamma)}
    \int_{t_{k-1}}^{t_k} (t_{n-\sigma} - s)^{-\gamma}
    (s - t_{k-1/2})\, ds\\
    &= \frac{2}{\tau_k(\tau_k+\tau_{k+1})\Gamma(2-\gamma)}\left(-\frac{\tau_k}{2}  ((t_{n-\sigma}-t_{k-1})^{1-\gamma}+(t_{n-\sigma}-t_k)^{1-\gamma})+\frac{(t_{n-\sigma}-t_{k-1}))^{2-\gamma}-(t_{n-\sigma}-t_k)^{2-\gamma}}{2-\gamma}\right)
\end{align*}
where $t_{k-1/2} = (t_{k-1} + t_k)/2$ and $k=1,2,\dotsc,n-1$. Based on the above components, we construct the discrete kernels $C_{k}^{(n,\gamma)}$. For $n=1$, we set $C_{0}^{(1,\gamma)}=A_{0}^{1}$ and for $2\leq n \leq N$, we define
\begin{equation}\label{l2weight}
    C_{n-k}^{(n,\gamma)} =
    \begin{cases}
    A_{n-1}^n - B_{n-1}^n, & k=1,\\
    A_{n-k}^n + \rho_{k-1}B_{n-k+1}^n - B_{n-k}^n, & 2 \leq k \leq n-1, \\
    A_0^n + \rho_{n-1}B_1^n, & k=n,
    \end{cases}
\end{equation}
where $\rho_k = \tau_k / \tau_{k+1}$.
Therefore, the final discrete approximation for the Caputo fractional derivative of order $\gamma\in (0,1)$ at the shifted point $t_{n-\sigma}$ is:
\begin{equation}\label{L2scheme}
    {}_{0}^{C} D_{t}^{\gamma} f(x, t_{n-\sigma})
    \approx C_0^{(n,\gamma)} f(x, t_n)
    - \sum_{k=1}^{n-1}
      \bigl(C_{n-k-1}^{(n,\gamma)} - C_{n-k}^{(n,\gamma)}\bigr) f(x_, t_k)
    - C_{n-1}^{(n,\gamma)} f(x, t_0),\quad 1\leq n \leq N.
\end{equation}
The above $L2-1_{\sigma}$ scheme on nonuniform mesh achieves the convergence rate $\mathcal{O}(\tau^{min(r\gamma, 2)})$ \cite{liao2018second}, under suitable regularity assumptions.
Therefore, the optimal grading exponent for obtaining the rate $O(N^{-2})$ is
$r = 2/\alpha$.

\begin{remark}
In view of Definition \eqref{caputodef}, the Caputo fractional derivative $ {}_{0}^{C}D_{x}^{\alpha} f$ of order $1<\alpha \leq 2$ can be expressed as
\[
{}_{0}^{C}D_{x}^{\alpha} f(x)= {}_{0}^{C}D_{x}^{\alpha-1}\bigl(f'(x)\bigr).
\]
Thus, in view of above relation, the approximation of the fractional derivative in \eqref{eq:fbvp} can be obtained by first evaluating the classical derivative \(f'(x)\) through automatic differentiation, and then applying the $L1$ or $L2-1_{\sigma}$ approximation given in \eqref{L1scheme} and \eqref{L2scheme}, respectively, to the resulting quantity with fractional order \(\gamma=\alpha-1\). 

\end{remark}

\subsubsection{Unified Matrix Representation}

Both the L1 and L2-1$_\sigma$ discretizations of the Caputo fractional derivative can be efficiently implemented as a matrix-vector product. To this end,
let $
\mathbf{u}_{\theta}
=
\bigl[u_{\theta}(\zeta_0),u_{\theta}(\zeta_1),\ldots,u_{\theta}(\zeta_N)\bigr]^T
$ denote the vector of neural network outputs evaluated at the chosen mesh or collocation points $\zeta_i$, $i=0,1,\dotsc,N$. 

Then, the discrete Caputo operator based on the $L1$ scheme can be written in the operator form
\begin{equation*}
    {}_{0}^{C}D_{\zeta}^{\gamma} \mathbf{u}_{\theta} \approx W_{\gamma} \mathbf{u}_{\theta},
\end{equation*}
where $W_{\gamma} \in \mathbb{R}^{(N+1) \times (N+1)}$ is the lower triangular matrix given by 

\begin{equation}\label{matrixl1}
W_{\gamma} = \frac{1}{\Gamma{(2-\gamma)}}
\begin{pmatrix}
0 & 0 & 0 & 0 & \cdots & 0\\
-w_{1,1} & w_{1,1} & 0 & 0 & \cdots & 0\\
-w_{2,2} & w_{2,2}-w_{2,1} & w_{2,1} & 0 & \cdots & 0\\
-w_{3,3} & w_{3,3}-w_{3,2} & w_{3,2}-w_{3,1} & w_{3,1} & \cdots & 0\\

\vdots & \vdots & \vdots & \vdots & \ddots & \vdots\\
-w_{N,N} & w_{N,N}-w_{N,N-1} & w_{N,N-1}-w_{N,N-2} & \cdots & w_{N,2}-w_{N,1} & w_{N,1}
\end{pmatrix}.
\end{equation}
Here, the weights \(w_{n,k}\), \(k=1,2,\ldots,n\), are defined analogously to \eqref{weightl1} with \(t_i\) replaced by \(\zeta_i\). Similarly, for the $L2-1_{\sigma}$ scheme \eqref{L2scheme}, we have
\[
 {}_{0}^{C}D_{\zeta}^{\gamma} \mathbf{u}_{\theta}\approx A_\gamma \mathbf u_{\theta},
\]
where
\begin{equation}\label{matrixl2}
A_\gamma=
\begin{pmatrix}
0 & 0 & 0 & 0 & \cdots & 0\\
-C_{0}^{(1,\gamma)} & C_{0}^{(1,\gamma)} & 0 & 0 & \cdots & 0\\
-C_{1}^{(2,\gamma)} & C_{1}^{(2,\gamma)}-C_{0}^{(2,\gamma)} & C_{0}^{(2,\gamma)} & 0 & \cdots & 0\\
-C_{2}^{(3,\gamma)} & C_{2}^{(3,\gamma)}-C_{1}^{(3,\gamma)} & C_{1}^{(3,\gamma)}-C_{0}^{(3,\gamma)} & C_{0}^{(3,\gamma)} & \cdots & 0\\

\vdots & \vdots & \vdots & \vdots & \ddots & \vdots\\
-C_{N-1}^{(N,\gamma)} & C_{N-1}^{(N,\gamma)}-C_{N-2}^{(N,\gamma)} & C_{N-2}^{(N,\gamma)}-C_{N-3}^{(N,\gamma)} & \cdots & C_{1}^{(N,\gamma)}-C_0^{(N,\gamma)} & C_{0}^{(N,\gamma)}
\end{pmatrix},
\end{equation}
with coefficients $C_k^{(n,\gamma)}$ are given by \eqref{l2weight}, again interpreted with respect to the mesh points $\zeta_i$.

For a fixed graded mesh parameter, the above matrices can be precomputed once outside the PINN architecture. Therefore, during the training process, the matrix-vector products \(W_{\gamma}\mathbf{u}_{\theta}\) or \(A_{\gamma}\mathbf{u}_{\theta}\) are directly used to approximate the corresponding temporal or spatial fractional derivatives. This allows the nonlocal fractional operators to be incorporated efficiently into the residual loss function.

\section{Loss Formulations and Learning Framework}\label{main}

In this section, we describe the framework used to obtain the approximate solutions of the proposed nonlocal differential equations on quantum graphs. The framework consists of loss formulations on the underlying graph together with suitable training strategies for enforcing governing equations, vertex transmission conditions, boundary conditions, and initial conditions. Moreover, additional features such as hard and soft constraint enforcement, Fourier feature embeddings, and learnable singularity features are incorporated to improve the accuracy and stability of the proposed PINN solver.

\subsection{Unified Loss Formulation}\label{sec:unified_loss}
As described in Section \ref{pinnsection}, the primary distinction between a conventional neural network and a Physics-Informed Neural Network lies in the formulation of the loss function. In this section, we derive a rigorous, unified residual loss formulation for the proposed problems on a quantum graph 
$\Gamma = (\mathcal{V}, \mathcal{E})$. 

To this end, let $\widehat{u}_{i,\bm{\theta}}$ denote the neural network approximation of the solution on an edge $e_i\in \mathcal{E}$, where $\bm{\theta}$ represents the trainable parameters of the neural network. For elliptic problems \eqref{eq:fbvp}, the approximation is of the form
$
\widehat u_{i,\bm{\theta}}:[0,\ell_i]\to\mathbb R
$,
whereas for the time-dependent problems \eqref{eq:parabolic}, it maps onto the spatiotemporal domain:
$
\widehat u_{i,\bm{\theta}}:[0,\ell_i]\times[0,T]\to\mathbb R.
$.
The total loss function $\mathcal{L}_{\mathrm{total}, \bm{\theta}}$ over the quantum graph $\Gamma$ is constructed as the sum of edge-wise residual loss and the loss associated with the vertex conditions. That is,
\begin{equation}\label{totalloss}
\mathcal{L}_{\mathrm{total}, \bm{\theta}}:=\sum_{e_i\in \mathcal{E}}\mathcal{L}_{e_i, \bm{\theta}}+\lambda \mathcal{L}_{node, \bm{\theta}},
\end{equation}
where $\lambda>0$ is an adaptive parameter that balances the relative weight of the different residuals. The term $\mathcal{L}_{e_i, \bm{\theta}}$ and $\mathcal{L}_{node, \bm{\theta}}$ in \eqref{totalloss} corresponds to the edge loss and node loss, respectively, which essentially compute the discrepancies between the target data and PINN predictions along the edges and the nodes of the quantum graph. We shall now provide the mathematical expression for each term.

For the training process, we define spatial collocation points $\{x_{i,j}\}_{j=1}^{N_x} \subset (0, \ell_i)$ on each edge $e_i\in \mathcal{E}$, $i=1,2,\dotsc,|\mathcal{E}|$, and temporal points $\{t_k\}_{k=1}^{N_t} \subset (0, T]$. The edge loss function associated to elliptic and evolutionary problem on each edge is formulated as
\begin{align*}
    \mathcal{L}_{e_i, \bm{\theta}}^{\mathrm{elliptic}} &= \frac{1}{N_x} \sum_{j=1}^{N_x} \big| \mathcal{N}[\hat{u}_{i,\bm{\theta}}](x_{i,j}) - f_i(x_{i,j}) \big|^2, \\
      \mathcal{L}_{e_i, \bm{\theta}}^{\mathrm{evolution}} &= \frac{1}{N_x N_t} \sum_{k=1}^{N_t} \sum_{j=1}^{N_x} \big| \mathcal{N}[\hat{u}_{i,\bm{\theta}}](x_{i,j}, t_k) - f_i(x_{i,j}, t_k) \big|^2+\frac{1}{N_x}\sum_{j=1}^{N_x} \bigl(\hat{u}_i(x_{i,j}, 0) - u_i^{(0)}(x_{i,j})\bigr)^2,~ i=1,2,\dotsc, |\mathcal{E}|,
\end{align*}
respectively, where $f_i$ represents the source terms on edge $e_i$ and $\mathcal{N}$ denotes the governing differential operator in the considered problems containing both local and non-local derivatives of the approximate solution $\hat{u}_{i,\bm{\theta}}$.

Now, to define the loss function for the nodes or vertices, we partition the set of all vertices $\mathcal{V}$ into a set of boundary (degree-one) vertices $\Omega \subset \mathcal{V}$, and a set of interior junction vertices  $\mathcal{V} \setminus \Omega$. The node loss term aggregates the continuity and Kirchhoff-Neumann conditions in junction (interior) nodes, and the conditions on boundary nodes:
\begin{equation}\label{nodelosselliptic}
    \mathcal{L}_{node, \bm{\theta}}^{\mathrm{elliptic}} = \sum_{v \in \Omega} \mathcal{L}_{\mathrm{bc},v} + \sum_{v \in \mathcal{V} \setminus \Omega} \left( \mathcal{L}_{\mathrm{cont},v} + \mathcal{L}_{\mathrm{KN},v} \right).
\end{equation}
Here, for boundary vertices $v\in \Omega$ subject to a Dirichlet condition $g_v$ or a Neumann condition $h_v$, the boundary penalty is
\begin{equation*}
    \mathcal{L}_{\mathrm{bc},v} = \big( \hat{u}_{i,\bm{\theta}}(x_v) - g_v \big)^2 \quad \text{or} \quad \big( \hat{u}_{i,\bm{\theta}}'(x_v) - h_v \big)^2,
\end{equation*}
where the index $i$ corresponds to the single edge $e_i$, incident on $v$, and based on the convention adopted $x_v\in \{0,\ell_i\}$ is the local coordinate for the vertex. At an interior junction $v \in \mathcal{V} \setminus \Omega$ of degree $d_v$, let $\{e_{j_1}, e_{j_2}, \dots, e_{j_{d_v}}\}$ denotes the set of all edges incident to $v$. Then, the continuity condition is enforced by penalizing pairwise differences:
\begin{equation*}
    \mathcal{L}_{\mathrm{cont},v} = \sum_{1 \le a < b \le d_v} \big( \hat{u}_{j_a,\bm{\theta}}(x_v) - \hat{u}_{j_b, \bm{\theta}}(x_v) \big)^2.
\end{equation*}
The loss function corresponds to the Kirchhoff-Neumann condition at the junction node $v$ reads
\begin{equation*}
    \mathcal{L}_{\mathrm{KN},v} = \left( \sum_{k=1}^{d_v} s_k \, \hat{u}_{j_k, \bm{\theta}}'(x_v) \right)^{\!2},
\end{equation*}
where $s_k \in \{-1, +1\}$ depending on whether the direction of edge $e_{j_k}$ is incoming or outgoing with respect to the junction $v$.

For evolutionary problems, we define the node loss function by taking the temporal average of the above vertex residuals over the discrete time levels $t_k$:
\begin{equation}\label{nodelossevo}
\mathcal{L}_{\mathrm{node}, \bm{\theta}}^{\mathrm{evolution}} = \frac{1}{N_t} \sum_{k=1}^{N_t} \left[ \sum_{v \in \Omega} \mathcal{L}_{\mathrm{bc},v}(t_k) + \sum_{v \in \mathcal{V} \setminus \Omega} \left( \mathcal{L}_{\mathrm{cont},v}(t_k) + \mathcal{L}_{\mathrm{KN},v}(t_k) \right) \right],
\end{equation}
where each constituent is explicitly defined as
\begin{equation*}
\begin{cases}\mathcal{L}_{\mathrm{bc},v}(t_k) = \big( \hat{u}_{i,\bm{\theta}}(x_v, t_k) - g_v(t_k) \big)^2 \ \ \text{or} \ \ \big( \hat{u}_{i,\bm{\theta}}'(x_v, t_k) - h_v(t_k) \big)^2, & \text{for } v \in \Omega, \\
\mathcal{L}_{\mathrm{cont},v}(t_k) = \sum_{1 \le a < b \le d_v} \big( \hat{u}_{j_a,\bm{\theta}}(x_v, t_k) - \hat{u}_{j_b, \bm{\theta}}(x_v, t_k) \big)^2, & \text{for } v \in \mathcal{V} \setminus \Omega, \\
\mathcal{L}_{\mathrm{KN},v}(t_k) = \left( \sum_{m=1}^{d_v} s_m \hat{u}_{j_m, \bm{\theta}}'(x_v, t_k) \right)^{2}, & \text{for } v \in \mathcal{V} \setminus \Omega.\end{cases}
\end{equation*}

Finally, we provide the loss formulation for inverse problems where the objective is to estimate governing system parameters from sparse and noisy observations. This is achieved by augmenting the above-defined loss function with an explicit empirical data residual term, defined as
\begin{equation}\label{data loss}
    \mathcal{L}_{\mathrm{data},i} = \frac{1}{N_d} \sum_{m=1}^{N_d} \big| \hat{u}_{i,\bm{\theta}}(x_{i,m}) - u_{\mathrm{data}}(x_{i,m}) \big|^2,~~i=1,2,\dotsc,|\mathcal{E}|,
\end{equation}
where $\{x_{i,m}\}_{m=1}^{N_d}$ denotes the set of discrete observation data points on edge $e_i$ and $u_{\mathrm{data}}$ denotes the observed values. The total per-edge loss is then directly modified by including the data-driven loss term \eqref{data loss}, which allows the total loss function for the inverse problems to retain the structure of the forward problem formulation, that is
\begin{equation}\label{eq:inverse_total_loss}
\begin{aligned}
    \mathcal{L}_{\mathrm{total}}^{\mathrm{inverse}}:&= \sum_{i \in \mathcal{E}} \mathcal{L}_{i} + \lambda \mathcal{L}_{\mathrm{node},\bm{\theta}}\\
    &= \lambda_d\sum_{i \in \mathcal{E}} \mathcal{L}_{\mathrm{data},i} + \sum_{i\in\mathcal{E}}\mathcal{L}_{e_i,\bm{\theta}}+\lambda \mathcal{L}_{\mathrm{node},\bm{\theta}}.
    \end{aligned}
\end{equation}
Note that $\lambda_d$ represents the adaptive weight of the data-driven loss term and its computation has been done identically to the adaptive node weight $\lambda$. During the training phase, the unknown physical parameters are treated as learnable variables and are optimized along with the neural network parameters using back-propagation. Notably, for estimating the fractional order of the considered nonlocal operators, the corresponding operational matrices derived in \eqref{matrixl1} and \eqref{matrixl2} must be continuously differentiable with respect to the fractional order parameter. This requirement is addressed by initializing the discrete matrix coefficients as differentiable analytical tensors\cite{pang2019fpinns}.

\subsection{Hard and Soft Constraint Enforcement}
\label{sec:HS-cons}
The proposed framework supports two modes for enforcing conditions at the graph vertices. The first mode corresponds to soft constraints, in which the residuals associated with the vertex conditions are incorporated into the loss function and penalized during training. The second mode is hard constraints, introduced in the context of neural-network-based differential equation solvers by Lagaris et al. \cite{lagaris1998artificial}, which modify the network architecture such that the prescribed boundary conditions are satisfied by construction. Therefore, in the latter case, the corresponding boundary residuals need not be included in the loss functional.

Soft constraints utilize the raw neural network prediction $\hat{u}_{i,\bm{\theta}}$ without any modification and treat it directly as the approximate solution on each edge $e_i$. The boundary conditions at the degree-one vertices, together with the continuity and Kirchhoff-Neumann conditions at the interior vertices, are imposed through the node loss term as described in \eqref{nodelosselliptic} and \eqref{nodelossevo}. This approach is the default operational mode in our framework since it applies uniformly to different types of graph constraints.
The hard constraint mode is formulated only for Dirichlet conditions imposed at boundary vertices.  Thus, rather than using raw network predictions, the trial solution is modified by means of suitable lifting functions to ensure that prescribed boundary values are satisfied identically. Consequently, the corresponding Dirichlet residuals are excluded from $\mathcal{L}_{\mathrm{bc},v}$ in this mode.  The modification for hard constraint mode in our framework is obtained by adapting the boundary-distance concept of Sukumar and Srivastava~\cite{sukumar2022exact} to the localized graph setting through a one-sided lifting.

To this end, let $\mathcal{F}_{i,\bm{\theta}}$ denote the raw neural network output on an edge $e_i$, and let $x_v\in \{0,\ell_i\}$ denotes the local coordinate for a boundary vertex $v\in \Omega$ incident with edge $e_i$. Since the other vertex of the same edge connects to an interior node, the hard constraint enforcement in the graph setting is one-sided, that is, it leaves the other endpoint of the edge unconstrained. For an elliptic problem, let $g_v$ be the prescribed Dirichlet value at a boundary vertex $x_v$. Then, the constrained network output is defined as

\begin{equation}\label{constelliptic}
\hat{u}_{i,\bm{\theta}}(x) = \left( \frac{|x - (\ell_i - x_v)|}{\ell_i} \right) g_v + \left( \frac{|x - x_v|}{\ell_i} \right) \mathcal{F}_{i,\bm{\theta}}(x), \quad i=1,2,\dotsc, |\mathcal{E}|.
\end{equation}
Upon evaluating the above expression, it is evident that at the boundary vertex $x=x_v$, the ansatz gives $\hat{u}_{i,\bm{\theta}}(x_v)=g_v$, which ensures that Dirichlet conditions are satisfied by construction. Moreover, at the other vertex, i.e., the interior junction where $|x-x_v|=\ell_i$, the output $\hat{u}_{i,\bm{\theta}}$ reduces to the raw neural network prediction $\mathcal{F}_{i,\bm{\theta}}$, which remains unconstrained.

In case of evolutionary problems with time-varying Dirichlet boundaries $g_v(t)$, the construction must accommodate the initial profile $u_i^{0}(x)$ and boundary data simultaneously. The exact mathematical ansatz chosen depends on the compatibility of the initial and boundary conditions at $t=0$. If $g_v(0) = u_{i}^0(x_v)$, then the spatiotemporal extension naturally takes the form:

\begin{equation}\label{constevol}
\hat{u}_{i,\bm{\theta}}(x,t) = \left( \frac{|x - (\ell_i - x_v)|}{\ell_i} \right) g_v(t) + \left( \frac{|x - x_v|}{\ell_i} \right) \mathcal{F}_{i, \bm{\theta}}(x,t).
\end{equation}

When the conditions are incompatible, i.e., $g_v(0) \neq u_{i}^0(x_v)$, then the initial and boundary data cannot be jointly satisfied by a single continuous hard constraint lifting function. In this case, we employ a time-decaying weight function:

\begin{equation*}
w(t) = e^{-\kappa t}, \quad \kappa = 20,
\end{equation*}
which provides the corresponding spatiotemporal ansatz
\begin{equation*}
\hat{u}_{i,\bm{\theta}}(x,t) = w(t)u_i^{0}(x) + \bigl(1 - w(t)\bigr)\left[ \left( \frac{|x - (\ell_i - x_v)|}{\ell_i} \right) g_v(t) + \left( \frac{|x - x_v|}{\ell_i} \right) \mathcal{F}_{i,\bm{\theta}}(x,t) \right].
\end{equation*}
Under this formulation, at $t = 0$ the weight $w(0) = 1$ recovers the initial condition $u_{i}^{0}(x)$ exactly. As \(t\) increases, \(w(t)\) decays to zero and the ansatz approaches the boundary-constrained lifting \eqref{constevol}. Thus, in this case, the above construction provides a transition from the initial data to the  boundary constrained representation, rather than enforcing both constraints exactly at $t=0$. 

\begin{remark}
In our framework, the hard constraint formulation is restricted only to Dirichlet conditions at boundary vertices. The Neumann conditions at boundary vertices, along with continuity and Kirchhoff-Neumann conditions at junction nodes, are more naturally imposed as soft constraints. Although hard-constrained formulation for Neumann data at boundary vertices can be constructed using higher-order lifting functions in the spirit of \eqref{constelliptic} and \eqref{constevol}, such constructions are less natural on metric graphs because they may affect the behavior of the function at the opposite endpoint, i.e., at an interior junction. This can interfere with the simultaneous enforcement of continuity and Kirchhoff--Neumann vertex conditions across incident edges. Therefore, these conditions are retained as soft constraints to maintain a flexible and stable optimization formulation.
\end{remark}

The advantage of hard constraints is that they enforce prescribed boundary data exactly and may provide better accuracy. However, their performance depends on the choice of lifting function and is highly sensitive to the distribution of collocation points, especially for nonlocal fractional models where graded meshes are used to resolve weak singularities. Soft constraints, on the other hand, provide a more flexible and highly robust formulation, since the same loss formulation can be used for different types of boundary conditions, graph topologies, and mesh distributions.

We compare the hard and soft constraints and demonstrate the limitations of hard constraints empirically through an example. To quantify the error in the estimation of the approximate solution provided by our framework, we define the relative discrete $L^2$ error for elliptic problems on the graph $\Gamma$ as
\begin{equation}\label{errorelliptic}
\frac{
\|\widehat{u}_{\bm{\theta}}-u\|_{L^2(\Gamma)}
}{
\|u\|_{L^2(\Gamma)}
}
\approx
\left( \frac{\sum_{i=1}^{|\mathcal{E}|} \sum_{m=0}^{N_i-1} \left| \hat{u}_{i,\bm{\theta}}(x_{i,m}) - u_{i}(x_{i,m}) \right|^2}{\sum_{i=1}^{|\mathcal{E}|} \sum_{m=0}^{N_i-1} \left| u_{i}(x_{i,m}) \right|^2} \right)^{1/2},
\end{equation}
where $x_{i,m}$ represents the discrete spatial collocation points along edge $e_i$, and $N_i$ represents the total number of spatial validation points sampled per edge, $\hat{u}_{i,\bm{\theta}}$ is the predicted solution by QGPINNs on edge $e_i$ and $u_i$ represents the exact solution on that edge. To ensure edges of varying lengths are considered equally, the number of validation points per unit length is uniform across all edges. This ensures that longer edges contain proportionally more validation points. Moreover, the $N_i$ validation points are randomly distributed throughout the spatial domain $e_i=[0,\ell_i]$ to ensure that the $L^2$ errors are not biased towards the points used in network training.

\begin{example}\label{eq:hard_prob}
We consider the fractional elliptic problem
\begin{equation*}
    {}_{0}^{C}D_{x}^{\alpha} u_i(x) + {}_{0}^{C}D_{x}^{\beta} u_i(x)=f_{i},\quad  i=1,2,3,
\end{equation*}
on a metric star graph with 3 edges (outward pointing) as given in Figure \ref{fig:star-graph} with vertex conditions \eqref{cont}-\eqref{kn} at node $v_0$ and Dirichlet vertex conditions \eqref{dc} at boundary nodes $v_i$, $i=1,2,3$. We take $\alpha=1.5$ and $\beta=0.5$, and the source terms $f_i$ are chosen such that
\[u_1(x)=x^3-2x,~~u_2(x)=x^{1.2}+x~~\text{and}~~u_3(x)=x^2+x\]
forms the exact solution of the problem.
\end{example}

\begin{table}[htbp]
    \centering
    \footnotesize
    \caption{Effect of the mesh grading factor $r$ on the relative discrete $L^2$ error \eqref{errorelliptic} for soft and hard constraints on a metric star graph.}
    \label{tab:grading_factor_comparison}
    \begin{tabular}{@{}lcc@{}}
        \toprule
        \textbf{Mesh Grading Factor} & \textbf{Soft Constraints} & \textbf{Hard Constraints} \\
        \midrule
        $r = 1$ & $1.78 \times 10^{-3}$ & $1.76 \times 10^{-3}$ \\
        $r = 2$         & $2.17 \times 10^{-3}$     & $3.35 \times 10^{-3}$ \\
        $r = 3$     & $1.66 \times 10^{-4}$     & $1.01 \times 10^{-3}$ \\
        \bottomrule
    \end{tabular}
\end{table}

Table~\ref{tab:grading_factor_comparison} compares the global relative $L^2$ error \eqref{errorelliptic} obtained using hard and soft constraint enforcement for different mesh grading factors. For the uniform mesh $r=1$, both approaches produce nearly identical errors. However, as the grading factor increases, the accuracy of the hard constraints framework deteriorates. In particular, for high grading factor $r=3$, the soft-constraint formulation gives much smaller error than the hard-constraint formulation. This indicates that, although hard constraints can be effective for simple geometries and weakly graded meshes, their accuracy may deteriorate when the mesh is strongly redistributed to capture weak singular behaviour in nonlocal models. The soft-constraint formulation is therefore more stable with respect to mesh grading in this example and is better suited as the default enforcement strategy in the proposed framework.

We also emphasise that the above comparison can only be interpreted as an empirical rather than a theoretical guarantee. In general, hard constraints may still be advantageous when the boundary data are simple and the lifting function is well adapted to the structure of the graph. However, for nonlocal problems on metric graphs, where boundary, continuity, and Kirchhoff--Neumann conditions must be handled simultaneously, the soft formulation provides a more uniform and robust implementation.

\subsection{Dynamic Balancing of Loss Function Weight}
\label{sec:lamb}

The choice of the weighting parameter \(\lambda\) in the total loss function \eqref{totalloss} plays an important role in balancing the loss residuals and significantly affects the efficacy of the PINN. For small \(\lambda\), the neural network may reduce the governing equation residual while violating the vertex constraints, and for large $\lambda$, the optimization process may overemphasize the residuals associated with boundary and vertex conditions at the expense of the governing equation. Therefore, a fixed chosen value of the weight parameter \(\lambda\), in general,  may not be equally effective during the training process. To
mitigate the task of manually tuning the parameter, our framework implements methods to dynamically update the value of $\lambda$ during training. We describe two distinct adaptive strategies: boundary data matching strategy and gradient pathology balancing, which are then integrated into a mixed dual update rule.

Let \(\lambda^{(n)}\)denotes the value of the weight parameter at the current training epoch \(n\), where \(n=0,1,\dots,E\), and \(E\) is the total number of training epochs, such that the normalized training progress is denoted as $p_n=\frac{n}{E}$. Each adaptive strategy produces a target weight \(\lambda^{(n)}_{target}\). However, a noisy loss floor may cause fluctuations in the target weight; thus, to avoid abrupt changes in \(\lambda\), we use a log-space exponential moving average to define a smoothed target weight
$$\overline{\lambda}_{\mathrm{target}}^{(n)} = \exp \left( \beta \ln \big(\overline{\lambda}_{\mathrm{target}}^{(n-1)}\big) + (1 - \beta) \ln \big(\lambda_{\mathrm{target}}^{(n)}\big) \right)$$
where $\beta\in(0,1)$ is a user-defined smoothing coefficient with a default value of $0.95$.

Given the smoothed target weight, we introduce a user-defined update rate $\rho$ (default set to $\rho=0.1$). To prevent unbounded growth of the loss weight $\lambda$, we apply strict bounds and express the updated value of $\lambda$ as
$$\lambda^{(n+1)} = \Pi_{[\lambda_{\min}, \lambda_{\max}]} \left( \exp \left( (1 - \rho) \ln \big(\lambda^{(n)}\big) + \rho \ln \big(\overline{\lambda}_{\mathrm{target}}^{(n)}\big) \right) \right)$$
where $\Pi_{[a,b]}(x)=max(a,min(x,b))$ is the standard clipping operator mapped to $[\lambda_{min},\lambda_{max}]$.
The above mechanism ensures that the relative adaptive weights of the loss function remain stable. We shall now provide the details to compute the target weight \(\lambda^{(n)}_{target}\) from both the strategies.

\subsubsection*{Boundary Data Matching Mechanism (BDMM)}

The boundary data matching strategy balances the objective function components by driving the loss associated with graph vertices towards a self calibrating target function. The aim is to prevent the vertex constraint contribution from becoming either negligible or excessively dominant during training. Let
\[
 \mathcal{L}_{\mathrm{edge},\bm{\theta}}
=
 \sum_{e_i\in\mathcal{E}}\mathcal{L}_{e_i,\bm{\theta}},
 \qquad
\mathcal{L}_{\mathrm{node},\bm{\theta}}
=
\sum_{v\in\Omega}\mathcal{L}_{\mathrm{bc},v}
+
\sum_{v\in\mathcal V\setminus\Omega}
\left(
\mathcal{L}_{\mathrm{cont},v}
+
\mathcal{L}_{\mathrm{KN},v}
\right)
\]
denote the total edge residual loss, and the total node residual loss, respectively, as defined in Section \ref{sec:unified_loss}. Then, at epoch \(n\), we define the instantaneous loss ratio as
\[
\lambda^{(n)}_{\mathrm{target}}
=
\frac{
\mathcal{L}_{\mathrm{node},\bm{\theta}}^{(n)}
}{
max(\mathcal{L}_{\mathrm{edge},\bm{\theta}}^{(n)},\varepsilon)
},
\]
where \(\varepsilon>0\) is a small stabilization parameter which has been used to avoid division by zero. The target weight is then applied to the smooth update rule described above.

\subsubsection*{Gradient Pathology Balancing}

The second strategy is adapted from the gradient balancing methodology proposed by Wang et al.~\cite{wang2021understanding}. In PINNs, different components of the loss may generate gradients of substantially different magnitudes, which may lead to slower convergence or can cause the optimization to prioritize certain components of the loss over the others. To reduce this effect, we dynamically adapt \(\lambda\) using the relative magnitudes of the gradients associated with the edge residual and the node residual across the network layers.

Let $\theta$ the set of all trainable parameters in the neural network. At epoch $n$, we define the gradient-based target weight by

$$\lambda_{\mathrm{target}}^{(n)} = \frac{ \sqrt{ \sum_{p \in \bm{\theta}} \left\|\nabla_{p} \mathcal{L}_{\mathrm{edge},\bm{\theta}}^{(n)} \right\|_2^2 } }{ \max \left( \sqrt{ \sum_{p \in \bm{\theta}} 
\left\|\nabla_{p} \mathcal{L}_{\mathrm{node},\bm{\theta}}^{(n)} \right\|_{2}^2}, \varepsilon \right) }$$
where $\Vert{}\cdot\Vert{}_2$ denotes the standard $L_2$ norm. The numerator captures the global gradient contribution from the edge residual, whereas the denominator provides the scale for the gradients generated by the node residual. This target weight is again used in the same smoothed update rule for $\lambda$.

\subsubsection*{Mixed Dual Strategy}

The above-defined adaptive strategies exhibit complementary regularizing characteristics. The BDMM monitors the relative size of the boundary residual, whereas the gradient pathology balancing accounts for the relative magnitude of the optimization gradients governed by different loss components. By leveraging their complementary properties, we introduce a four-phase mixed dual strategy that will be used in the proposed framework.

To this end, let \(E\) be the total number of training epochs and let \(\tau\in(0,1)\) denote the epoch fraction after which the adaptive update is frozen.

\begin{enumerate}
    \item \textit{Phase 0} \((0\leq n<0.02E)\):
    The value of the weight parameter \(\lambda\) is kept fixed at its initial value, and no adaptive update is applied during this stage.

    \item \textit{Phase 1} \((0.02E\leq n<0.1E)\):  
    The BDMM controller is active, and the target weight is chosen as the BDMM target.

    \item \textit{Phase 2} \((0.1E\leq n<\tau E)\):  
    The gradient pathology controller is active, and the target weight is chosen as the gradient-based target.

    \item \textit{Phase 3} \((\tau E\leq n\leq E)\):  
    Both controllers are inactive, and therefore \(\lambda\) remains fixed during the final part of the first-order optimization stage.
\end{enumerate}

Notably, after the completion of the first-order Adam optimization stage, the balancing parameter \(\lambda\) is fixed to the median of its values over the final \(N_{\mathrm{median}}\) Adam iterations. This value is then remains static throughout the subsequent L-BFGS refinement stage. In our implementation, we choose \(N_{\mathrm{median}}=1000\).\vspace{0.1cm}

We now demonstrate the comparative analysis of these strategies through a numerical example. For evolutionary problems, the relative discrete $L^2$ error on the graph $\Gamma$ is defined as
\begin{equation}\label{errorevol}
\frac{
\|\widehat{u}_{\bm{\theta}}-u\|_{L^2(0,T;\Gamma)}
}{
\|u\|_{L^2(0,T;\Gamma)}
}
\approx \left( \frac{\sum_{i=1}^{|\mathcal{E}|} \sum_{m=0}^{N_i-1} \sum_{k=0}^{N_t-1} \left| \hat{u}_{i,\bm{\theta}}(x_{i,m}, t_k) - u_{i}(x_{i,m}, t_k) \right|^2}{\sum_{i=1}^{|\mathcal{E}|} \sum_{m=0}^{N_i-1} \sum_{k=0}^{N_t-1} \left| u_{i}(x_{i,m}, t_k) \right|^2} \right)^{1/2},
\end{equation}
where $t_k$ denotes the $k$-th temporal validation point distributed along the temporal domain, and $N_t$ represents the total number of temporal validation points sampled across the temporal domain $[0, T]$. The other notations are the same as in \eqref{errorelliptic}.

\begin{example}\label{dynamic}
We consider the following time-fractional evolution equation on the tree-like metric graph given in Figure \ref{fig:tree-graph}:
\begin{equation}
\label{lambda_problem}
 {}_{0}^{C}D_{t}^{\gamma}\, u_{i}(x,t) + u_i(x,t)\frac{\partial u_i}{\partial x}(x,t) - \nu \frac{\partial^2 u_i}{\partial x^2}(x,t) + c u_i(x,t) = f_i(x,t), \quad x \in (0, 1), \quad t \in (0, 1],
\end{equation}
where the fractional order is chosen as $\gamma = 0.5$, the kinematic viscosity (diffusion coefficient) is $\nu = 0.5$, and the linear reaction coefficient is $c = 5.0$. The initial condition is $u_i(x,0) = 0$, for $x \in [0, 1]$ and homogeneous Dirichlet boundary conditions are applied at the boundary nodes $v_0, v_2, v_4$. The source terms $f_i(x,t)$ are chosen such that exact solution on the edge $i$ is:
\begin{equation}
u_i(x,t) = A_i \sin(k x) t^2, \quad \text{with } k = 4\pi, \text{ and } A = [2.0, 1.0, 1.0, 1.0],
\end{equation} 
where the coefficients $A_i$ have been obtained by enforcing the continuity \eqref{cont} and Kirchhoff-Neumann \eqref{kn} constraints at the node $v_1$ and $v_3$.
\end{example}

\begin{figure}[htbp]
  \centering
  
  \begin{minipage}[c]{0.48\textwidth}
    \centering
    \begin{tikzpicture}[
      scale=0.6,
      vertex/.style={circle, draw, fill=white, minimum size=7pt, inner sep=0pt, thick},
      hub/.style={circle, draw, fill=black!80, minimum size=9pt, inner sep=0pt, thick},
      pendant/.style={circle, draw, fill=violet!30, minimum size=7pt, inner sep=0pt, thick},
      lbl/.style={font=\small},
      edge_lbl/.style={font=\small, text=blue!80!black}
    ]
  
    \node[pendant, label={[lbl]left:$v_0$}]  (v0) at (-3.5, 0)   {};
    \node[pendant, label={[lbl]right:$v_2$}] (v2) at ( 2.5, 2.2) {};
    \node[pendant, label={[lbl]right:$v_4$}] (v4) at ( 5.5,-1.8) {};

    \node[hub, label={[lbl]above:$v_1$}]     (v1) at (-0.5, 0)   {};
    \node[hub, label={[lbl]above:$v_3$}]     (v3) at ( 2.5,-1.8) {};

    \draw[thick] (v0) -- node[above, edge_lbl] {$e_0$} (v1);
    \draw[thick] (v1) -- node[above left, edge_lbl] {$e_1$} (v2);
    \draw[thick] (v1) -- node[below left, edge_lbl] {$e_2$} (v3);
    \draw[thick] (v3) -- node[above, edge_lbl] {$e_3$} (v4);
    
    \end{tikzpicture}
    \captionof{figure}{A four-edge cascading tree metric graph.}
    \label{fig:tree-graph}
  \end{minipage}
  \hfill 
  \begin{minipage}[c]{0.48\textwidth}
    \centering
    \begin{tabular}{@{} lc @{}}
        \toprule
        \textbf{Strategy}  & \textbf{Relative $L^2$ Error}\\
        \midrule
        $\lambda=1$      & $1.26 \times 10^{-2}$ \\
        $\lambda=30$      & $5.40 \times 10^{-3}$ \\
        Gradient Pathology                 & $4.86 \times 10^{-3}$\\
        BDMM     & $1.19 \times 10^{-2}$\\
        Dual         & $4.36 \times 10^{-3}$\\
        \bottomrule
    \end{tabular}
    \captionof{table}{Relative \(L^2\) error \eqref{errorevol} for Example~\ref{dynamic} obtained using different weighting strategies.}
    \label{tab:lambda_strategies}
  \end{minipage}

\end{figure}

\begin{figure}[h]
    \centering
    \includegraphics[width=0.7 \linewidth]{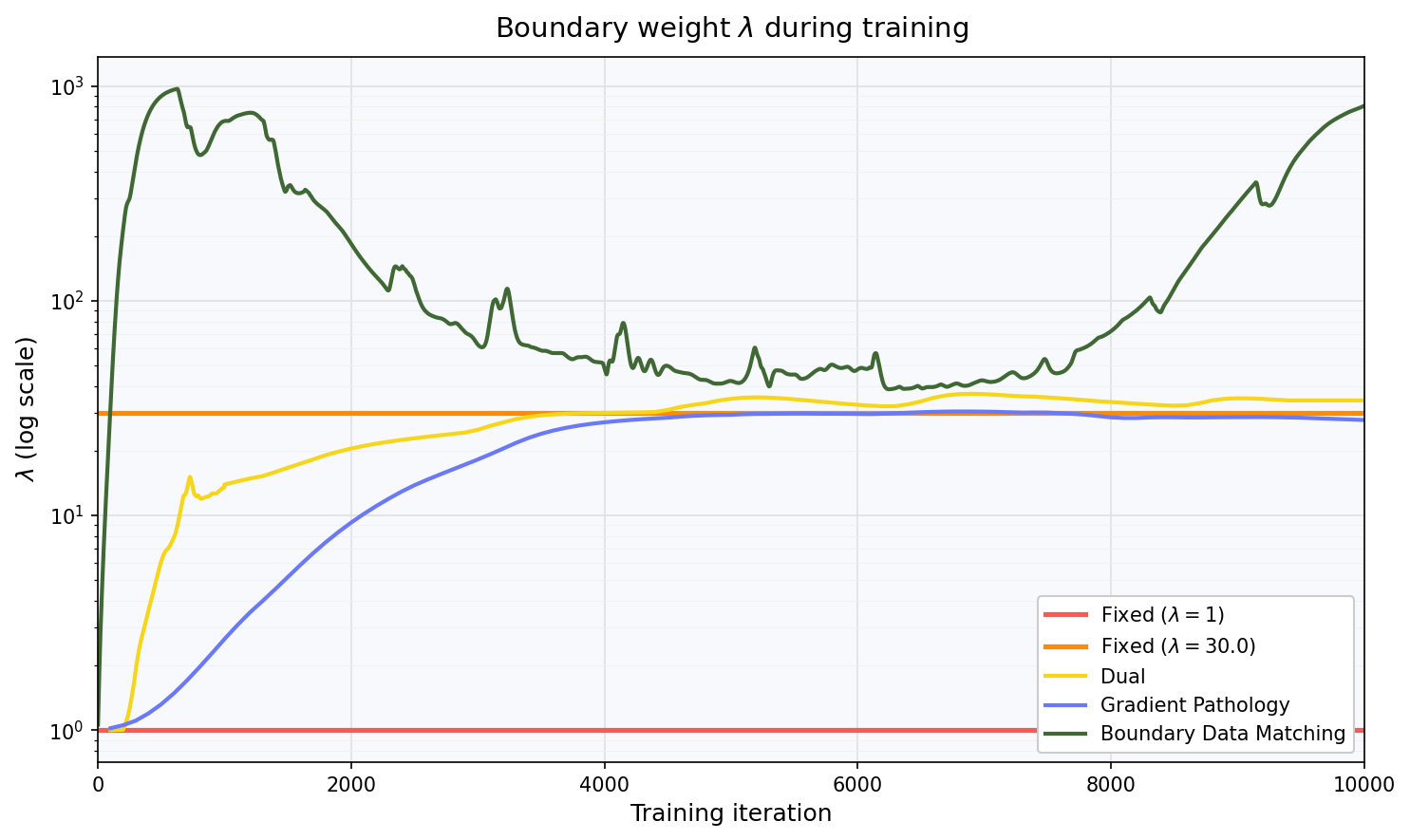}
    \caption{Dynamic evolution of the balancing parameter $\lambda$ during training for different adaptive strategies.}
    \label{fig:lamb_strat}
\end{figure}

For this comparison, we utilize a configuration of 4 hidden layers with 128 neurons each. The Adam optimization phase is run for $10000$ epochs followed by a L-BFGS refinement phase. All factional derivatives were computed using the $L2-1_{\sigma}$ scheme. Table~\ref{tab:lambda_strategies} compares the relative  \(L^2\) error \eqref{errorevol} obtained using a fixed value of \(\lambda\) and the proposed adaptive balancing strategies for fractional order $\gamma=0.5$. The adaptive strategies outperformed the fixed $\lambda=1$ and $\lambda=30$ variants without requiring any problem-specific calibration. Figure~\ref{fig:lamb_strat} shows the evolution of the weight \(\lambda\) during training for the adaptive strategies. These results support the use of dynamic balancing in the proposed framework, since it reduces sensitivity to a fixed choice of the loss-weighting parameter.

Therefore, the above numerical results show that dynamic balancing of \(\lambda\) can improve the performance of the proposed PINN framework as compared to a fixed manually chosen loss weight, where the proposed dual strategy provided the most accurate result for this example. Nevertheless, the relative performance of the individual adaptive strategies may depend on the structure of the problem, the type of boundary and vertex conditions, and the relative strength of the edge and node residuals. For more problems and discussion, we refer to \cite{saketgithub}.

\subsection{Spectral Bias and Fourier Feature Embedding}
\label{sec:fourier}
Rather than utilizing only a conventional multi-layer perceptron (MLP) architecture, as described in Section \ref{pinnsection}, our framework also incorporates random Fourier feature embeddings. The underlying feature is implemented to mitigate the phenomenon of spectral bias, which is inherent in standard neural networks, as described by Rahaman et al.~\cite{rahaman2019spectral}.
Spectral bias describes the tendency of MLPs to preferentially learn low-frequency components of a target function over the high-frequency components. Consequently, standard PINNs may fail to accurately resolve oscillatory solutions or sharp variations during the early stages of training.

To address this issue, we project the raw spatial or spatio-temporal input coordinates into a higher-dimensional feature space using a Fourier feature mapping adapted from Tancik et al.~\cite{tancik2020fourier}. Let \(\mathbf{x}\in\mathbb{R}^d\) denote the input coordinate vector, which for elliptic problems represents the spatial coordinate, and for evolutionary problems includes both spatial and temporal coordinates. The Fourier feature map is defined by
\begin{equation*}
    \gamma(\mathbf{x}) =
    \bigl[\cos(2\pi \mathbf{B}\mathbf{x}),\,
           \sin(2\pi \mathbf{B}\mathbf{x})\bigr]^\top
\end{equation*}
where $\mathbf{B} \in \mathbb{R}^{m\times d}$ represents a static frequency matrix whose entries are sampled once prior to training from a Gaussian distribution $\mathcal{N}(0,\sigma^2I)$. The parameter \(\sigma>0\) controls the range of frequencies introduced into the input representation.

In the Fourier feature PINN framework, the embedded coordinate \(\gamma(\mathbf{x})\) is passed through the MLP instead of the raw coordinate \(\mathbf{x}\). This allows the network to represent oscillatory components more effectively and reduces the low-frequency preference of standard MLP-based PINNs. From the neural tangent kernel (NTK) perspective, Fourier feature mappings modify the effective kernel associated with the MLP. As shown by Tancik et al.~\cite{tancik2020fourier}, this transformation can convert the effective NTK into a stationary kernel with a tunable bandwidth, which improves the learning of high-frequency components. This motivates the use of Fourier embeddings in our framework for nonlocal problems on metric graphs, where  the solution may exhibit oscillatory behaviour, weak singularities, or multiscale spatial structures.

\setlength{\fboxw}{0.20\textwidth}
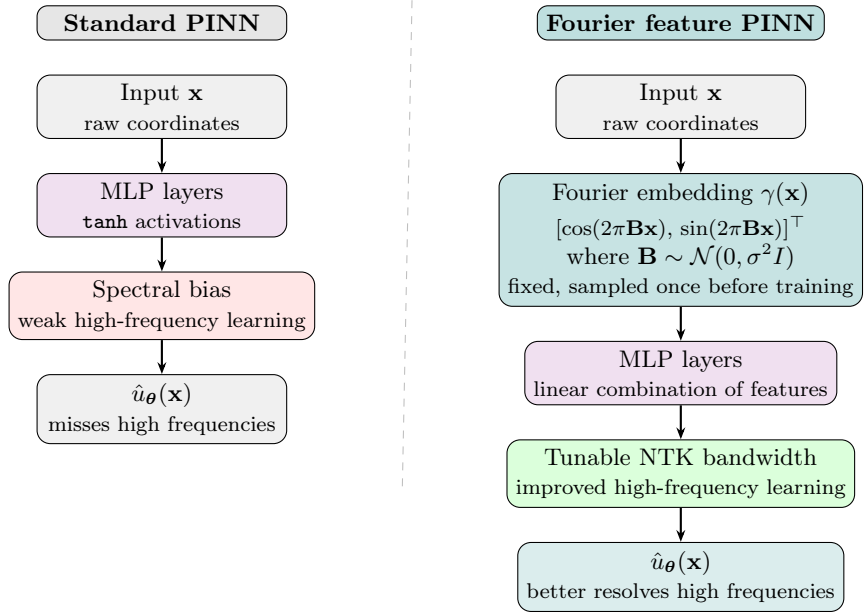
\begin{figure}[htbp]
\centering
\begin{tikzpicture}[
  node distance=0.45cm,
  box/.style={rectangle, rounded corners=5pt, draw,
              minimum width=\fboxw, minimum height=0.7cm,
              align=center, font=\small},
  graybox/.style={box, fill=gray!12},
  purpbox/.style={box, fill=violet!12},
  redbox/.style ={box, fill=red!10},
  tealbox/.style={box, fill=teal!14},
  grnbox/.style ={box, fill=green!14},
  embbox/.style ={box, fill=teal!22, minimum width=\fboxw},
  arr/.style={-{Stealth[length=4.5pt]}, thick},
  header/.style={rectangle, rounded corners=4pt, draw,
                 minimum width=\fboxw, align=center, font=\small\bfseries},
]

\node[header, fill=gray!20]          (hl) {Standard PINN};
\node[graybox, below=of hl]          (l1) {Input $\mathbf{x}$ \\ \footnotesize raw coordinates};
\node[purpbox, below=of l1]          (l2) {MLP layers \\ \footnotesize \texttt{tanh} activations};
\node[redbox,  below=of l2]          (l3) {Spectral bias \\ \footnotesize weak high-frequency learning};
\node[graybox, below=of l3]          (l4) {$\hat{u}_{\bm{\theta}}(\mathbf{x})$ \\ \footnotesize misses high frequencies};
\draw[arr] (l1)--(l2); \draw[arr] (l2)--(l3); \draw[arr] (l3)--(l4);

\node[header, fill=teal!25,
      right=0.20\textwidth of hl]    (hr) {Fourier feature PINN};
\node[graybox, below=of hr]          (r1) {Input $\mathbf{x}$ \\ \footnotesize raw coordinates};
\node[embbox,  below=of r1]          (r2) {Fourier embedding $\gamma(\mathbf{x})$ \\[2pt]
                                            \footnotesize $[\cos(2\pi\mathbf{B}\mathbf{x}),\,\sin(2\pi\mathbf{B}\mathbf{x})]^\top$
      \\where $\mathbf{B} \sim \mathcal{N}(0,\sigma^2I)$ \\ \footnotesize fixed, sampled once before training};
\node[purpbox, below=of r2]          (r3) {MLP layers \\ \footnotesize linear combination of features};
\node[grnbox,  below=of r3]          (r4) {Tunable NTK bandwidth \\ \footnotesize improved high-frequency learning};
\node[tealbox, below=of r4]          (r5) {$\hat{u}_{\bm{\theta}}(\mathbf{x})$ \\ \footnotesize better resolves high frequencies};
\draw[arr] (r1)--(r2); \draw[arr] (r2)--(r3);
\draw[arr] (r3)--(r4); \draw[arr] (r4)--(r5);

\draw[dashed, gray!50, thin]
  ($ (hl.east)!0.5!(hr.west) $) ++(0,0.3)
  -- ($ (l4.east)!0.5!(r5.west) $) ++(0,-0.3);
\end{tikzpicture}
\caption{Framework comparison between a standard PINN and a Fourier feature PINN. The Fourier embedding \(\gamma(\mathbf{x})\) modifies the input representation by introducing oscillatory features and thus, mitigating spectral bias and improving the learning of high-frequency components.}

\label{fig:spectral-bias}
\end{figure}

To quantify the performance enhancement driven by this Fourier mapping under non-local dynamics, we consider the following time-fractional diffusion equation on the metric star graph given in Figure~\ref{fig:star-graph}:

\begin{equation*}
    {}_{0}^{C}D_{t}^{\gamma} u_i(x, t) = \frac{\partial^2 u_i}{\partial x^2}(x,t) + f_i(x,t), \quad \text{for } i = 1, 2, 3,
\end{equation*}
subject to the homogeneous initial condition $u_i(x,0) = 0$ for all $i \in \{1, 2, 3\}$. On boundary nodes $v_1$ and $v_2$, the Neumann conditions are imposed as
$\frac{\partial u_1}{\partial x}(1, t) = 24\pi t^2$ and
$\frac{\partial u_2}{\partial x}(1, t) = 0.3\pi \cos(0.3\pi) t^2$, while on $v_3$, the Dirichlet condition $ u_3(1, t) = -5.4 t^2$ has been imposed. The source terms $f_i$ has been chosen such that

\begin{equation*}
    u_i(x,t) = c_i t^2 \sin(\omega_i x), \quad \text{for } i = 1, 2, 3,
\end{equation*} 
forms the exact solution. The values of $c_i$ are chosen to satisfy continuity and Kirchhoff-Neumann conditions at the junction node $v_0$. The edge-wise spatial frequencies are:
$\omega_1=6\pi;\omega_2=0.3\pi;\omega_3=4.5\pi$
with coefficients $c_1=4.0; c_2=1.0; c_3=-5.4$. 

We apply our optimization strategies on the fractional order $\gamma=0.5$ and use a configuration of 4 hidden layers with 128 neurons each. The Fourier embedding variant has 128 dimensions with $\sigma=1$. The Fourier dimensions represents the number of waves sampled, hence $128$ Fourier dimensions translate to $64$ sine and $64$ cosine waves. We use the dual scheme and train the Adam phase for $10000$ iterations. All fractional derivatives are computed using the $L2-1_{\sigma}$ scheme.

\begin{table}[htbp]
    \centering
    \footnotesize
    \caption{Performance comparison of an embedded Fourier feature PINN with a standard PINN.}
    \label{tab:fourier_comparison}
    \begin{tabular}{@{}lc@{}}
        \toprule
        \textbf{Strategy} & \textbf{Relative $L_2$ Error} \\
        \midrule
        Fourier Disabled  & $1.90 \times 10^{-1}$ \\
        Fourier Enabled          & $5.92 \times 10^{-3}$ \\
        \bottomrule
    \end{tabular}
\end{table}

\begin{figure}[htbp]
    \centering
    \includegraphics[width=1.0\linewidth]{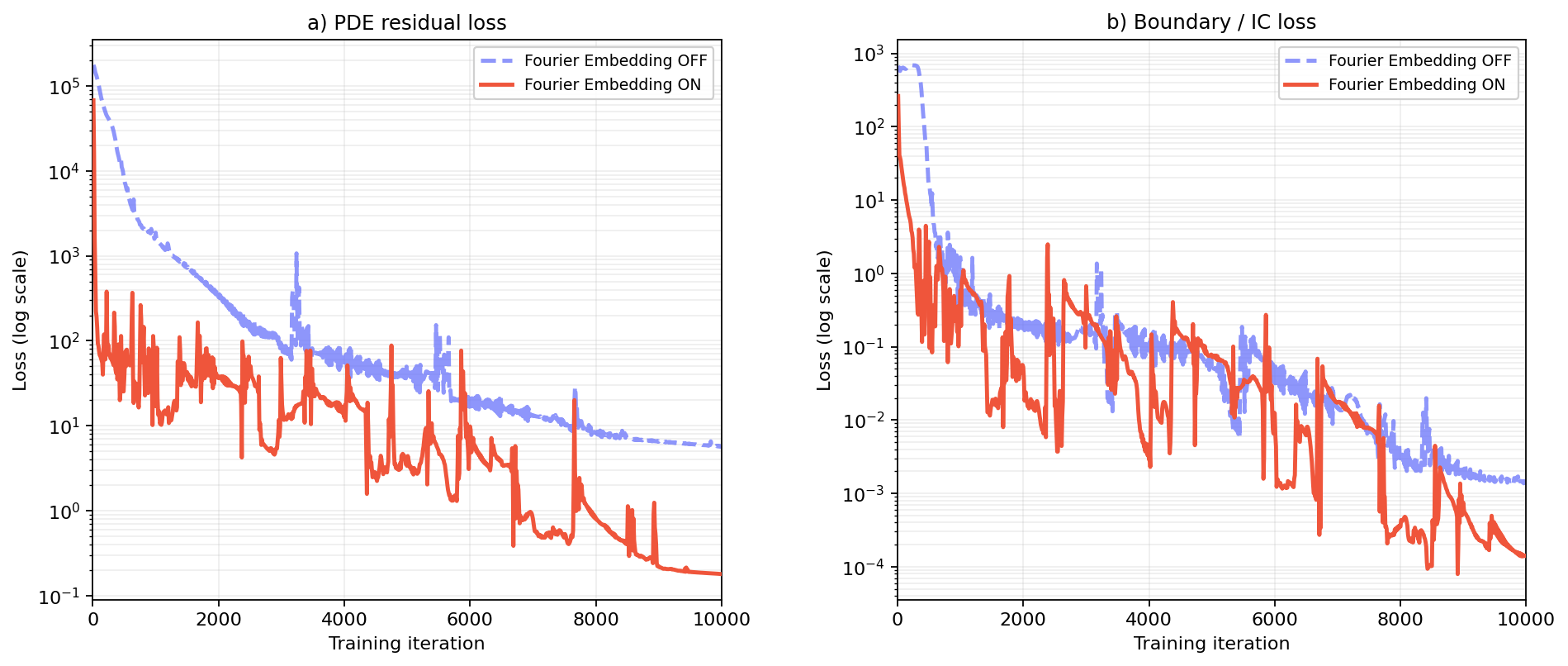}
    \caption{Performance of the network with Fourier embedding enabled and disabled. a) represents the edge residual loss, b) shows the node residual loss.}
    \label{fig:fourier_loss}
\end{figure}

The numerical results  in Table~\ref{tab:fourier_comparison} demonstrates how Fourier feature embedded significantly reduce the relative $L_2$ approximation error between the exact and predicted solution. This confirms that the the standard MLP-based PINN struggles to resolve the oscillatory components of the solution, while the Fourier feature architecture  provides a more suitable input parametrization for high-frequency behavior. Further, Figure ~\ref{fig:fourier_loss} shows this improvement by comparing the training behavior with and without Fourier embedding. The Fourier-enabled model significantly reduces the edge residual loss and node loss, and thus,

support the use of Fourier feature embeddings as a default component in the proposed framework.

\subsection{Singularity-Capturing Feature}
\label{sec:singularity}

While non-uniform graded meshes are employed to minimize truncation errors in traditional discrete schemes, their direct implementation within PINNs may create severe optimization bottlenecks. Graded meshes heavily concentrate collocation points near the singular point of the domain, such as \(x=0\) or \(t=0\), which may cause the coefficients of the discrete operators defined in \eqref{weightl1} and \eqref{l2weight} to become large and highly uneven. Moreover, the smooth structure of the activation functions in standard MLPs makes it difficult to accurately capture singular derivatives near such points.

To mitigate this behaviour, we incorporate a singularity-capturing feature in the spirit of Li and Tan~\cite{li2026novel}. Rather than forcing the network to approximate singular derivatives at the origin using only raw spatial or spatiotemporal coordinates, we introduce an explicit, non-smooth auxiliary singularity feature defined by
\begin{equation*}
    \mathcal{Z}(c) = c^\xi,
\end{equation*}
where \(\xi>0\) is the singularity-capturing parameter, and $c$ corresponds to the singular coordinate. In particular, \(c=t\) for time-fractional evolution problems, while \(c=x\) may be used for space-fractional elliptic problems on an edge parametrized from the singular endpoint.

Although graded meshes increase the resolution near the singular point, they do not by themselves change the approximation space of the neural network. The singularity-capturing feature complements mesh grading by enriching the input space with a non-smooth coordinate \(\mathcal{Z}(c)=c^\xi\), which allows the network to represent fractional-type singular behaviour more naturally. In our graph-based framework, particularly for time-dependent problems, the neural network on each edge is evaluated using the augmented input
\[
\widehat{u}_{i,\bm{\theta}}(x,t)
=
\mathcal{F}_{i,\bm{\theta}}(x,t,\mathcal{Z}(t)).
\]

The mathematical utility of this feature becomes evident from the classical chain rule. For instance, 

\begin{equation*}
    \frac{\partial \widehat{u}_{i,\bm{\theta}}}{\partial t}
    =
    \frac{\partial \mathcal{F}_{i,\bm{\theta}}}{\partial t}
    +
    \frac{\partial \mathcal{F}_{i,\bm{\theta}}}{\partial \mathcal{Z}}
    \frac{d\mathcal{Z}}{dt}
    =
    \frac{\partial \mathcal{F}_{i,\bm{\theta}}}{\partial t}
    +
    \frac{\partial \mathcal{F}_{i,\bm{\theta}}}{\partial \mathcal{Z}}
    \xi t^{\xi-1}.
\end{equation*}
which is singular at \(t=0\) whenever \(0<\xi<1\). Therefore, the augmented input allows the neural network to represent solution components whose derivatives exhibit fractional-type singularities near the initial time.

The exponent \(\xi\) may be prescribed when the expected singularity order is known. However, in practical
computational settings, especially with inverse parameter estimation, the analytical order of the initial
singularity is unknown. Therefore, in our implementation, \(\xi\) is treated as a trainable parameter and optimized together with the network parameters \(\bm{\theta}\). This allows the network to adapt the strength of the singular feature during training. We initialize \(\xi\) using the corresponding fractional order and update it dynamically via backpropagation. 

We compare the performance of the proposed singularity-capturing PINN with a standard PINN by considering the following problem defined on the tree graph in Figure~\ref{fig:tree-graph}.
\begin{equation}
\label{z(t)_problem}
 {}_{0}^{C}D_{t}^{\gamma}\, u_{i}(x,t) + u_i(x,t)\frac{\partial u_i}{\partial x}(x,t) - \nu \frac{\partial^2 u_i}{\partial x^2}(x,t) + c u_i(x,t) = f_i(x,t), \quad x \in (0, 1), \quad t \in (0, 1],
\end{equation}
where the fractional order is chosen as $\gamma = 0.5$, the kinematic viscosity (diffusion coefficient) is $\nu = 0.5$, and the linear reaction coefficient is $c= 5.0$. The initial condition is $u_i(x,0) = 0$, for $x \in [0, 1]$ and zero Dirichlet boundary conditions are applied at the boundary nodes $v_0, v_2, v_4$ such that $u(x,t) = 0, \quad \text{for } t \in (0, 1]$. The source terms $f_i(x,t)$ are chosen such that exact solution on the edge $i$ is:

\begin{equation}
u_i(x,t) = A_i \sin(k x) t^\gamma, \quad \text{with } k = 4\pi, \text{ and } A = [2.0, 1.0, 1.0, 1.0].
\end{equation} 
For this comparison we utilize a configuration of 4 hidden layers with 128 neurons each. The Adam optimization phase is run for $10000$ epochs followed by a L-BFGS refinement phase. We choose the dual adaptive weighting and Fourier feature embedding with 128 dimensions. All fractional derivatives were computed using the $L2-1_{\sigma}$ scheme. 

\begin{table}[htbp]
    \centering
    \footnotesize
    \caption{Effect of mesh grading and singularity-capturing feature on the relative \(L^2\) error }
    \label{tab:singularity_comparison}
    \begin{tabular}{@{}lccc@{}}
        \toprule
        \textbf{Mesh grading factor} 
        & \textbf{\(\mathcal Z(t)\) disabled} 
        & \textbf{\(\mathcal Z(t)\) enabled} \\
        \midrule
        \(r=1\) & $2.1\times10^{-3}$& $9.71\times10^{-4}$ \\
        \(r=2\) & $2.00\times10^{-3}$ & $7.65\times10^{-4}$ \\
        \(r=3\) & $2.38\times10^{-3}$& $1.31\times10^{-3}$\\
        \(r=4\) & $2.78\times10^{-3}$ & $1.44\times10^{-3}$ \\
        \bottomrule
    \end{tabular}
\end{table}

\begin{figure}[htbp]
    \centering
    \includegraphics[width=1\linewidth]{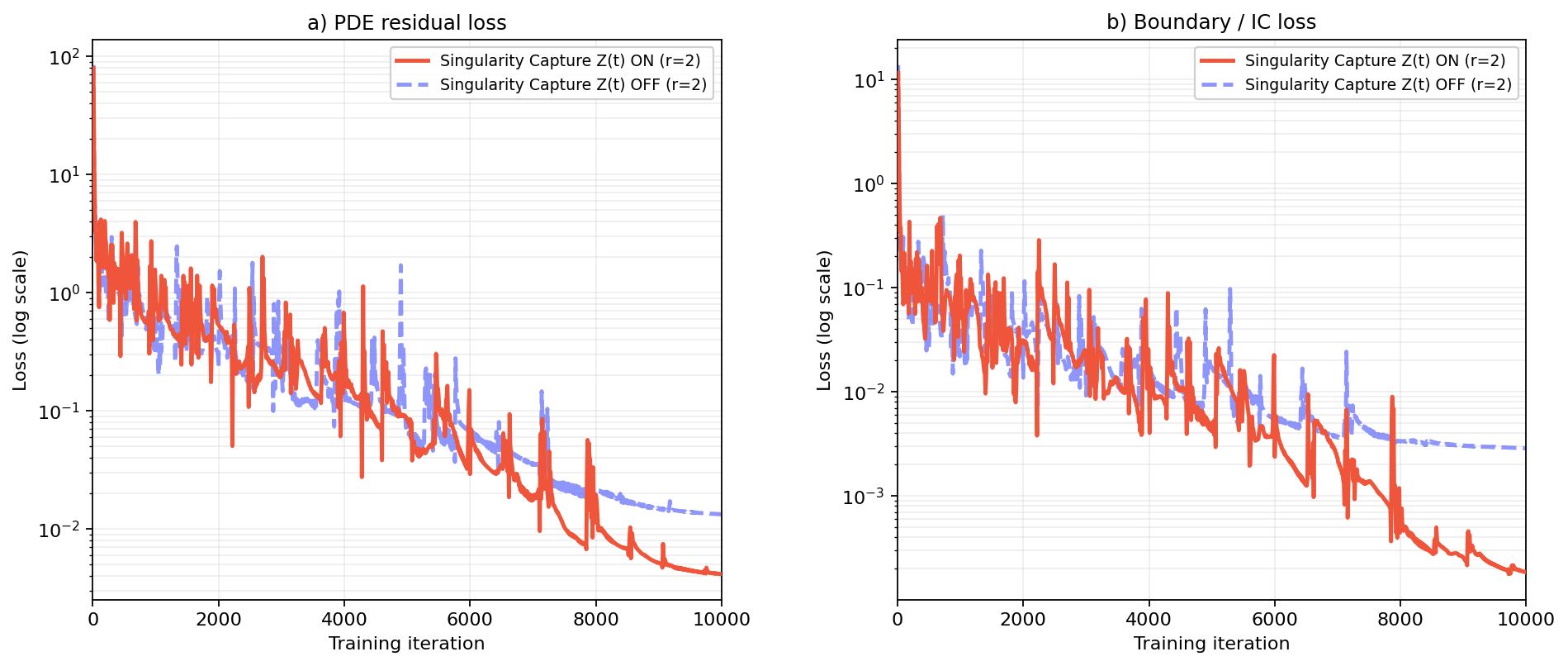}
    \caption{Comparison of PINN approximations with and without the singularity-capturing feature \(Z(t)=t^\xi\).}
    \label{fig:z(t)_loss}
\end{figure}

Table~\ref{tab:singularity_comparison} highlights the benefits of the singularity capturing feature when the governing system has an initial singularity. The ability of the PINN to represent the solution as a function of $t^\xi$ has resulted in a significant improvement in accuracy and lower PDE and boundary residuals, as seen in figure~\ref{fig:z(t)_loss}. The above ablation comparing the engine at various grading parameters highlights an interesting behaviour of our collocation methodology. Our network uses a unified mesh approach where the graded mesh \eqref{graded} used for the approximation of the Caputo fractional derivative directly serves as the neural network training points. Graded meshes concentrate points towards the initial portion of the domain. While this reduces the truncation error of the approximation schemes (minimum at $2/\alpha$ for $L2-1_{\sigma}$), it also starves later portions of the domain of collocation points. Therefore, for such implementations, the choice of the grading parameter $r$ is a trade-off between minimizing the truncation error and ensuring that the entire domain is well represented with collocation points.

\section{Framework Architecture and Implementation Blueprint}
\label{sec:architecture}

To bridge the mathematical methods established in Section \ref{main} with an efficient computational framework, we present a modular implementation architecture for QGPINNs. The implementation is designed to separate the graph topology, the nonlocal physical operators, and the neural network optimization components. This separation allows the framework to handle different graph configurations without requiring structural modifications to the underlying deep learning architecture.

\subsection{Neural Network Structure on a Metric Graph}
\label{subsec:geometric_allocation}

In case of Euclidean domains, a standard neural network typically takes the coordinates as input through a single MLP. However, in the context of metric graphs, such a global configuration may become inefficient as physical coefficients, edge lengths, and solution gradients can vary significantly across different edges of a complex network. Consequently, a single set of weights have to represent substantially different local behaviour across the graph, which can affect the convergence. This effect can be further amplified by the nonlocality of the fractional differential operators. 

Therefore, in our framework, we assign an independent edge-wise sub-network \(\mathcal{N}_i(\cdot;\bm{\theta}_i)\) to each edge $e_i \in \mathcal{E}$, where
\(\bm{\theta}_i\) denotes the trainable parameters of the subnetwork on the edge \(e_i\). 

Although the subnetworks are parametrized edge-wise, they are trained through a global graph loss. Thus, the approximation on each edge are coupled during optimization through the continuity, Kirchhoff--Neumann, and boundary residuals defined at the graph vertices.

The user specifies the graph topology as an edge list
\begin{equation}
    \mathcal{G}_{\text{list}} = \left\{ \left(u_i, v_i, \ell_i\right) \right\}_{i=1}^{|E|},
\end{equation}
where $u_i, v_i \in \mathcal{V}$ denote the vertices of the edge $e_i$ with $\ell_i \in \mathbb{R}^+$ denotes its length. During initialization, the computational engine builds
a \texttt{networkx} graph object and allocates an independent sub-network to each edge.
 
We illustrate a minimal code snippet to create a graph object below. 

\begin{lstlisting}
class StarGraph:
    nodes = [0, 1, 2, 3]
    edges = [(0, 1, L), (0, 2, L), (0, 3, L)]
\end{lstlisting}

Algorithm~\ref{alg:graph-construction} describes the graph construction and edge-wise subnetwork allocation process.

\begin{algorithm}[htbp]
\caption{Graph Construction and Sub-Network Allocation}
\label{alg:graph-construction}
\begin{algorithmic}[1]
\Require Edge list $\mathcal{G}_{\text{list}} = \{(u_i, v_i, \ell_i)\}_{i=1}^{|\mathcal{E}|}$
\State $G \gets \text{BuildGraph}(\mathcal{G}_{\text{list}})$
\Comment{\texttt{networkx} graph object}
\For{each edge $e_i = (u_i, v_i, \ell_i) \in \mathcal{E}$}
    \State $\mathcal{N}_i \gets \text{InitializeMLP}(\text{hidden\_layers}, \text{hidden\_dim})$
    \If{use\_fourier\_features}
        \State $\mathcal{N}_i \gets \text{PrependFourierEmbedding}(\mathcal{N}_i, \text{embed\_dim}, \sigma)$
    \EndIf
    \State AssignSubNetwork($e_i$, $\mathcal{N}_i$)
    \Comment{edge-wise parameters coupled through the global graph loss}
\EndFor
\State \Return $G$, $\{\mathcal{N}_i\}_{i=1}^{|E|}$
\end{algorithmic}
\end{algorithm}

\subsection{Separation of the Physics Layer from the Solver Engine}
\label{subsec:operator_separation}

Frameworks that define the network architecture and governing physics in a single script are difficult to generalize across different graph topologies and differential operators. We therefore separate the implementation into two components: a solver engine and a run-script. The solver engine handles all neural network architecture, while in run-script the user specifies only the graph topology and problem-specific information, such as the governing equation on each edge. In particular, the engine handles the creation of the fractional differentiation matrices and loss function definitions internally based on the parameters of the run script. The following code snippet describes the creation of the physics object.

\label{sec:creating_phys_obj}
\begin{lstlisting}
# The physics module for a time-fractional Burgers equation on edges
        def F(self, x, t, u, u_x, u_xx, dt_alpha_u, f):
            return dt_alpha_u + u * u_x - self.nu * u_xx - f
    
        def get_ic(self, x): #initial condition if necessary
            return torch.sin(math.pi * x)
\end{lstlisting}

The above separation allows the same solver engine to be reused for different equations and graph topologies. Once the graph topology and governing equations are defined, the user configures the solver object using object attributes. Table~\ref{tab:solver-config} summarizes the main configuration attributes and their roles.

\begin{table}[h]
\centering
\caption{Important solver attributes. The framework also provides a secondary set of attributes to allow granular control over network architecture}
\label{tab:solver-config}
\begin{tabularx}{\textwidth}{@{}lX@{}}
\toprule
\textbf{Stage} & \textbf{Role} \\
\midrule
\texttt{set\_constraints()}     & Specify boundary condition type (soft/hard) and values per vertex \\
\texttt{set\_frac\_scheme()}    & Select fractional discretization (L1 or L2-1$\sigma$) \\
\texttt{set\_mesh()}            & Choose spatial/temporal collocation density and grading \\
\texttt{set\_architecture()}   & Configure sub-network width, depth, and Fourier features \\
\texttt{set\_singularity\_capture()} & Configure the singularity capturing feature.\\
\texttt{compile()}               & Initialise the network architecture\\
\texttt{train()}                 & Choose the weight balancing strategy, training iterations and begin optimization \\
\bottomrule
\end{tabularx}
\end{table}

\begin{figure}[h]
\centering
\resizebox{0.79\textwidth}{!}{
\begin{tikzpicture}[
    font=\sffamily\footnotesize,
    node distance = 5mm and 8mm,
    >={Stealth[length=2.0mm]},
    every path/.style={line width=0.8pt},
    stage/.style={rectangle, rounded corners=2pt, draw=black!70, line width=0.8pt,
                  minimum width=45mm, minimum height=7mm, align=center,
                  inner sep=2pt, fill=white},
    io/.style={stage, rounded corners=5pt, fill=gray!12},
    global/.style={stage, fill=blue!8, draw=blue!55!black},
    local/.style={stage, fill=orange!10, draw=orange!70!black},
    solver/.style={stage, fill=violet!10, draw=violet!60!black, minimum width=96mm},
    config/.style={stage, fill=green!8, draw=green!45!black, minimum width=96mm},
    train/.style={stage, fill=red!8, draw=red!55!black, minimum width=96mm},
    output/.style={io, minimum width=96mm, fill=gray!18},
    lbl/.style={font=\sffamily\bfseries\scriptsize, text=black!65},
    arr/.style={-> , draw=black!70},
]

\node[io] (input) {$\mathcal{G}_{\text{list}} = \{(u_i, v_i, \ell_i)\}_{i=1}^{|\mathcal{E}|}$ \\ \footnotesize (user-defined graph topology)};

\node[global, below left=12mm and -6mm of input] (nxgraph) {\texttt{networkx} Graph Object \\ \footnotesize vertices $\mathcal{V}$, edges $\mathcal{E}$, lengths $\ell_i$};
\node[global, below=of nxgraph] (subnet) {Per-Edge Sub-Network Allocation \\ \footnotesize edge-wise MLP $\mathcal{N}_i(x,t;\bm{\theta}_i)$ per edge \\ \footnotesize (+ optional Fourier feature embedding)};

\node[local, below right=12mm and -6mm of input] (physics) {Physics Definition (run script) \\ \footnotesize \texttt{F(x,t,u,u\_x,u\_xx,dt\_alpha\_u)} \\ \footnotesize \texttt{get\_ic(x)}, boundary data};
\node[local, below=of physics] (disc) {Fractional Discretization \\ \footnotesize Caputo $\partial_t^\gamma u$ via L1 / $L2-1_{\sigma}$ scheme \\ \footnotesize non-local assembly matrix $W_\gamma / A_{\gamma}$};

\node[lbl, above=1mm of nxgraph] {Global structure (topology)};
\node[lbl, above=1mm of physics] {Local structure (edge physics)};

\node[solver, below=14mm of $(subnet)!0.5!(disc)$] (solver) {Solver Instantiation \\ \footnotesize \texttt{ParabolicPINNSolver(graph, physics\_list)} \quad / \quad \texttt{EllipticPINNSolver(\ldots)}};

\node[config, below=of solver] (config) {
  Configuration of neural network parameters and various mathematical schemes.
};

\node[train, below=of config] (train) {
  Training \\[2pt]
  \footnotesize dual-phase optimization (Adam $\to$ L-BFGS)\\
  \footnotesize singularity-capturing feature, adaptive $\lambda$ balancing \\
  \footnotesize continuity and Kirchhoff-Neumann coupling enforced at internal junction nodes
};

\node[output, below=of train] (out) {Trained QGPINN Solution \\
\footnotesize edge-wise approximations \(\widehat{u}_{i,\bm{\theta}}\) on \(e_i\) \\
\footnotesize globally coupled through vertex conditions};

\draw[arr] (input.south) -| (nxgraph.north);
\draw[arr] (input.south) -| (physics.north);
\draw[arr] (nxgraph) -- (subnet);
\draw[arr] (physics) -- (disc);
\draw[arr] (subnet.south) |- ($(subnet.south)!0.5!(solver.north)$) -| (solver.north west);
\draw[arr] (disc.south) |- ($(disc.south)!0.5!(solver.north)$) -| (solver.north east);
\draw[arr] (solver) -- (config);
\draw[arr] (config) -- (train);
\draw[arr] (train) -- (out);

\begin{scope}[on background layer]
\node[fit=(nxgraph)(subnet), draw=blue!40, dashed, rounded corners=4pt, inner sep=5mm, line width=0.7pt] (fitL) {};
\node[fit=(physics)(disc), draw=orange!60!black, dashed, rounded corners=4pt, inner sep=5mm, line width=0.7pt] (fitR) {};
\end{scope}

\end{tikzpicture}
} 
\caption{System architecture and pipeline flow of QGPINNs.}
\label{fig:pipeline-flow}
\end{figure}
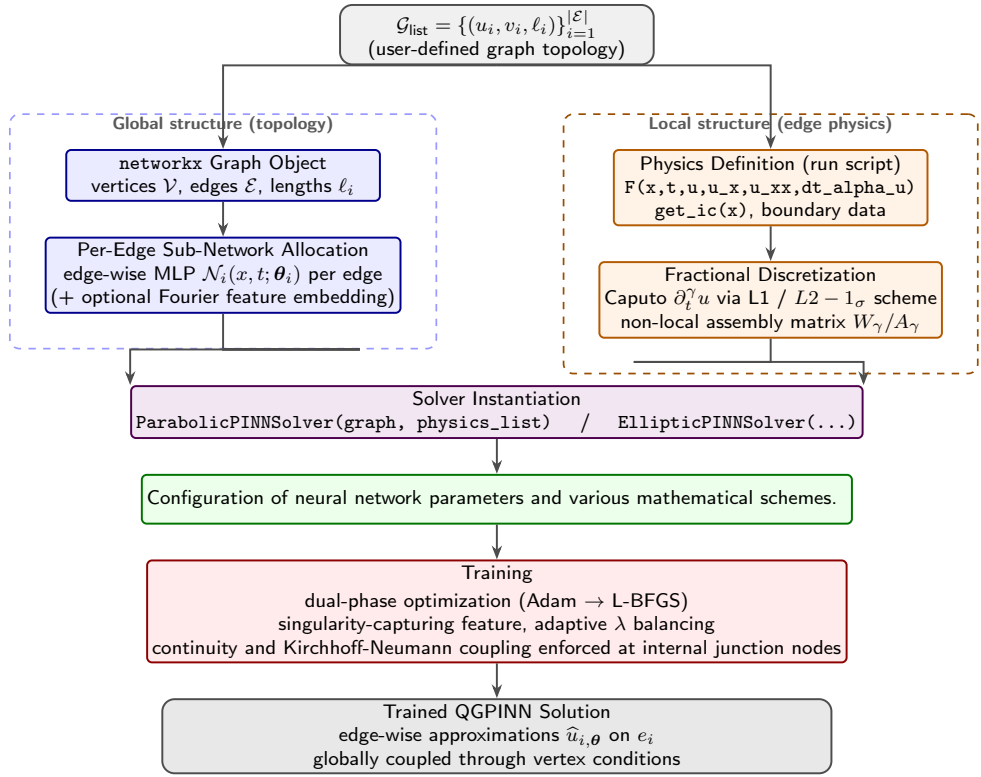

Boundary conditions are specified at boundary vertices by assigning the condition type
(\texttt{dirichlet}/\texttt{neumann}) and the corresponding value or time-dependent function. the vertex transmission conditions at junction nodes are handled separately by the engine from the graph topology and enforced through the continuity and Kirchhoff--Neumann losses defined in \eqref{cont} and \eqref{kn}. The following code snippet describes the syntax used to define the boundary condition.

\begin{lstlisting}
    bc_types = {1: 'neumann', 2: 'neumann', 3: 'dirichlet'}
    bc_values = {
        1: lambda t: -math.pi * 5*torch.pow(t.clamp(min=1e-10), ALPHA),
        2: 0.0,
        3: lambda t: 5*torch.pow(t.clamp(min=1e-10), ALPHA)
    }
    solver.set_constraints('soft', bc_types=bc_types, bc_values=bc_values)
\end{lstlisting}

After graph construction and solver configuration, the training procedure assembles the total loss \eqref{totalloss}. Depending on the selected configuration, the framework may also incorporate Fourier feature embeddings, hard constraint enforcement for Dirichlet boundary data, dynamic loss balancing, and singularity-capturing inputs.

The optimization-phase follows a dual-stage schedule. First, the model is trained using the Adam optimizer together with the dynamic balancing strategy to update the loss weight \(\lambda\). After the first-order stage, the value of \(\lambda\) is frozen, and the network parameters are further refined using L-BFGS. 

The trained model provides edge-wise approximations \(\widehat{u}_{i,\bm{\theta}}\) on all edges, which together define the approximate solution on the quantum graph. Figure~\ref{fig:pipeline-flow} provides a complete overview of the implementation pipeline of QGPINNs, from graph input and physics definition to solver configuration, training, and output generation.

The above modular and structured design of QGPINNs allows the framework to handle different graph topologies, fractional operators, boundary conditions, and training configurations without changing the core solver architecture.   The next section demonstrates these capabilities through numerical experiments on various graphs, including real network structures.
\section{Numerical Experiments}\label{numerical}

This section evaluates the proposed framework against standard implementations and physical topologies to assess its accuracy and computational performance. Problem 1 is defined as a nonlinear fractional elliptic system on a tadpole graph. Tadpole graphs are well-studied in the quantum graph literature and are considered to be building blocks
for more sophisticated graphs. We utilize the problem to provide insight into the efficacy of the various optimization schemes introduced in Section \ref{main}.  For problem 2, we consider a time-fractional diffusion equation on the same tadpole graph. We compare the relative discrete $L^2$ error \eqref{errorevol} based on two different discretizations, namely, an auxiliary graded mesh with distributed training points~\cite{li2026novel}, and a unified mesh, where the graded mesh \ref{graded}, used for the approximation of the Caputo fractional derivative, directly serves as the neural network training points. Problem 3 describes a time fractional Burger's equation defined on an open channel drainage system. The presence of heavy vegetation along the banks may induce non-local wave propagation behaviour, for which the considered Caputo derivative is suitable for the modeling. Problem 4 describes a time-fractional telegraph equation on the well-known bus network system. High-frequency voltage events are susceptible to memory effects, this implies that the usage of the Caputo derivative is better suited to describe such a phenomenon. Problems 3 and 4 do not contain an exact solution, thus, we employ secondary physical constraints to test the approximation's validity. To mitigate initialization bias and ensure reproducibility, each problem is benchmarked across three fixed random seeds for $500$ epochs. The optimal seed initialization is then selected for complete network training. Examples with an exact solution are tested for numerical accuracy based on the relative discrete $L^2$ error given in \eqref{errorelliptic} and \eqref{errorevol}. All experiments were executed on NVIDIA T4 and P100 GPUs.

\subsection{Non-linear elliptic fractional differential equation on a Tadpole graph}\label{ellipnum}

This problem evaluates the accuracy and computational overhead of the optimization schemes described in Section 3. The optimization schemes are successively enabled on a problem to identify their performance benefits. Consider the following elliptic differential equation defined on the tadpole graph given in Figure \ref{fig:tadpole_graph}:

\begin{equation*}
    {}_{0}^{C}D_{x}^{\alpha}u_{i}(x) + V \left(u_{i}(x)\frac{du_{i}}{dx}\right) - \left({}_{0}^{C}D_{x}^{\beta}u_{i}(x)\right)^2 = f_{i}(x), \quad x \in (0, L_{i}), \quad i \in \{0, 1, 2, 3\},
\end{equation*}
where $\alpha = 1.5$, and $\beta = 0.5$ and we choose the advection velocity coefficient $V = 1.0$. The topology has Dirichlet Constraints at boundary vertices $v_3$ and $v_4$ such that $u_3(v_3)=u_3(1)=4$ and $u_4(v_4)=u_4(1)=1.6 \cdot 2^{0.6}$. The forcing terms $f_i(x)$ are chosen to allow the following solution on each edge:
 \begin{align*}
    u_1(x) &= x^{1.6} - 2^{0.6}x, && \text{for } x \in e_1 \text{ (Loop)} \\
    u_2(x) &= x^{1.6} + x, && \text{for } x \in e_2 \text{ (Inner Tail)} \\
    u_3(x) &= x^{1.6} + A_3 x^2 + B_3 x + 2, && \text{for } x \in e_3 \text{ (Outer Tail)} \\
    u_4(x) &= x^{1.6} + B_4 x, && \text{for } x \in e_4 \text{ (Branch)}
\end{align*}
The parameters $A_3$, $B_3$, and $B_4$ have been obtained by enforcing the continuity \eqref{cont} and Kirchhoff-Neumann \eqref{kn} constraints at the nodes $v_1$ and $v_2$.

\begin{center}
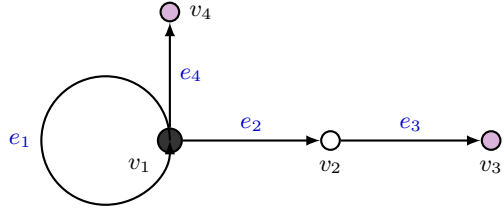

    \nopagebreak
    \begin{tikzpicture}[
        scale=0.85,
        vertex/.style={circle, draw, fill=white, minimum size=7pt, inner sep=0pt, thick},
        hub/.style={circle, draw, fill=black!80, minimum size=9pt, inner sep=0pt, thick},
        pendant/.style={circle, draw, fill=violet!30, minimum size=7pt, inner sep=0pt, thick},
        lbl/.style={font=\small},
        edge_lbl/.style={font=\small, text=blue!80!black}
    ]
  
    \node[hub, label={[lbl]below left:$v_1$}]     (v1) at (0,0)   {};
    \node[vertex, label={[lbl]below:$v_2$}]       (v2) at (2.5,0) {};
    \node[pendant, label={[lbl]below:$v_3$}]      (v3) at (5.0,0) {};
    \node[pendant, label={[lbl]right:$v_4$}]      (v4) at (0, 2.0) {};

    \draw[thick, -latex] (v1) arc[start angle=0, end angle=360, radius=1.0cm] node[midway, left, edge_lbl] {$e_1$};

    \draw[thick, -latex] (v1) -- node[above, edge_lbl] {$e_2$} (v2);

    \draw[thick, -latex] (v2) -- node[above, edge_lbl] {$e_3$} (v3);

    \draw[thick, -latex] (v1) -- node[right, edge_lbl] {$e_4$} (v4);
    
    \end{tikzpicture}
    \captionof{figure}{Tadpole graph representation. We choose the length of edges as $L_1=2.0$; $L_2=1.0$; $L_3=1.0$ and $L_4=1.0$, where $L_i$ is the length of the $i^{th}$ edge.} 
    \label{fig:tadpole_graph}
\end{center}

We compare the performance and accuracy of the proposed method by considering several variants with different features. The neural network parameters for each variant are defined in Table~\ref {tab:p2_param_burg}. Moreover, to quantify the computational load of each variant, we compare the peak GPU memory (MB) and a relative utilization metric defined as
\begin{equation*}
   \text{GPU-seconds} = \sum_{i} \Delta t_i \left( \frac{\text{Util}_i}{100} \right) 
\end{equation*}

\begin{center}
    \captionof{table}{Training and Mesh Parameters}
    \label{tab:p2_param_burg}
    \small
    \begin{tabular}{l cccc}
        \toprule
        \textbf{Parameter} & \textbf{Variant 1} & \textbf{Variant 2} & \textbf{Variant 3} & \textbf{Variant 4} \\
        \midrule
        Spatial point density $(N_x/L_i$)& $100$ & $100$ & $100$ & $100$ \\
        Spatial Grading (r) & $2$ & $2$  & $2$ & $2$  \\
        Training Iterations   & $10000$ & $10000$ & $10000$ & $10000$ \\
        Fractional scheme & $L2-1_{\sigma}$ & $L2-1_{\sigma}$ & $L2-1_{\sigma}$ & $L2-1_{\sigma}$\\
        Neurons per layer & 128 & 128 &128&128\\
        Hidden Layers& 4 & 4 &4 &4 \\
        Constraint type & Soft & Soft & Soft&Soft\\
        Adaptive Weight Strategy & Fixed & Dual & Dual &Dual\\
        Singularity Capture & Disabled & Disabled & Enabled & Enabled\\
        Fourier Dimension& 0 & 0 & 0 & 128\\
        Fourier $\sigma$ & - & - & -& 1\\
        \bottomrule
    \end{tabular}
\end{center}

\begin{figure}[h]
	\begin{center}
		\centering
		\subfigure[]{%
        \label{fig:computational_p1}
			\includegraphics[scale=0.20]{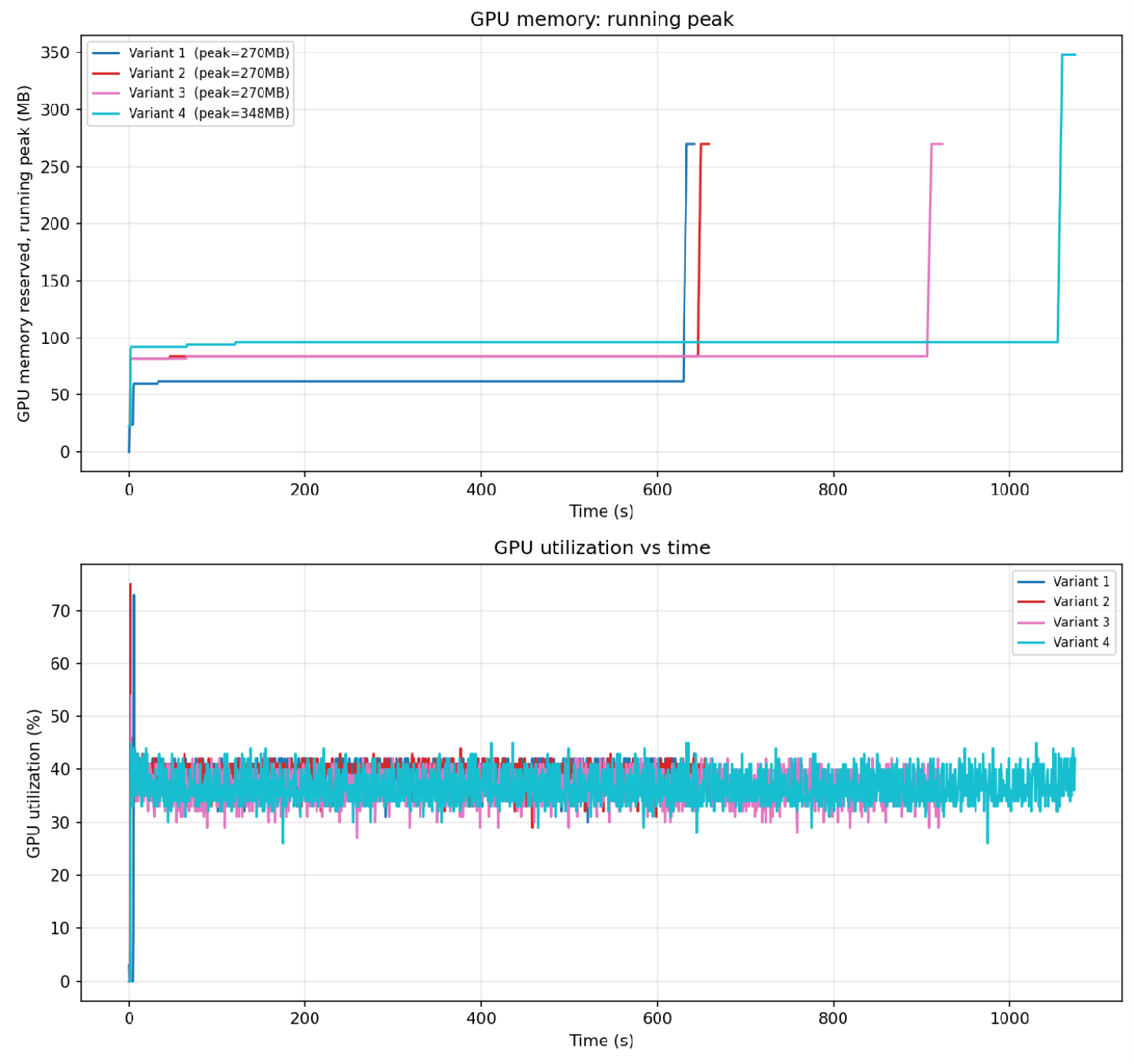}}
		\subfigure[]{%
        \label{fig:p1_dirich}
		\includegraphics[scale=0.24]{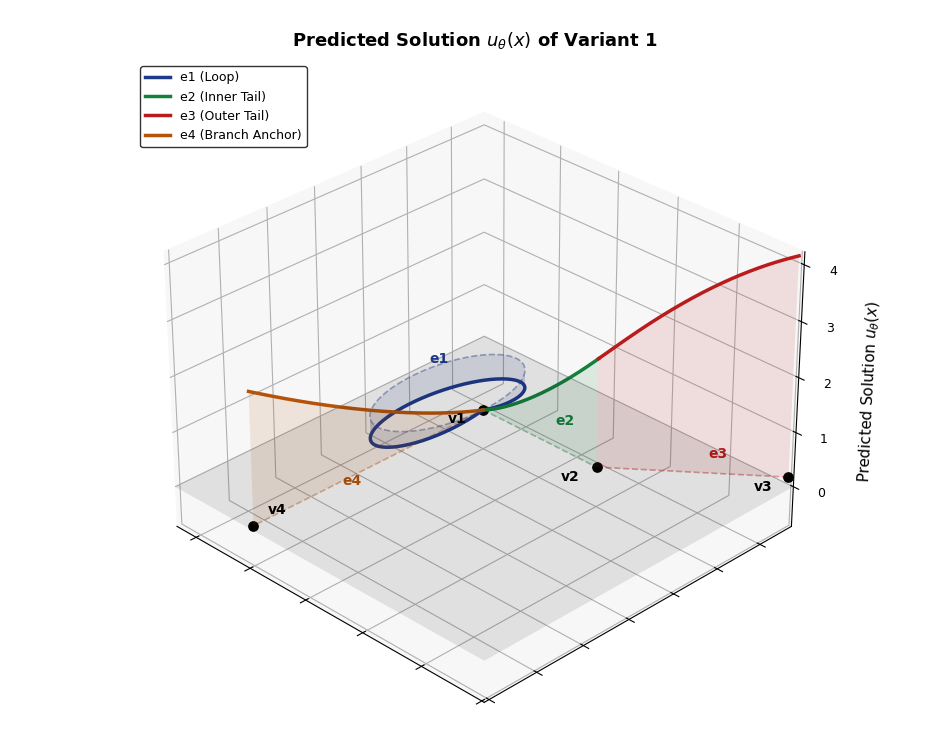}}
         \caption{(a) Computational demand of each variant. (b) A 3-Dimensional plot of the predicted solution.}
	\end{center}
\end{figure}

Table~\ref{tab:p1_comp} shows the relative performance and computational load of each variant. The enabling of the singularity capturing feature in variant 3 improved accuracy significantly, pointing towards the capability of such a feature to efficiently and accurately capture singularities. Enabling Fourier feature embedding further in Variant 4 improved performance and reduced the global $L^2$ error to the order of $10^{-5}$. Figure~\ref{fig:computational_p1} illustrates the computational load of each variant during training. The computational cost of each variant highlights the relative efficiency of these methods. While such numerical accuracy trends can also be obtained by increasing point densities and neural network sizes, the above optimization strategies are significantly more efficient. We observe that Variant 4 required an additional 141 MB of memory, however it improves the accuracy significantly as compared to other variants. Figure~\ref{fig:p1_dirich} depicts the predicted solution on the tadpole graph for Variant 1, where one can observe that the predicted solution satisfies the continuity constraint \eqref{cont} at the junction nodes $v_1$ and $v_2$.

\begin{center}
    \captionof{table}{Performance comparison of variants.}
    \label{tab:p1_comp}
    \footnotesize
    \begin{tabular}{@{}lccc@{}}
        \toprule
        \textbf{Variant} & \textbf{Global Relative $L_2$ Error} & \textbf{GPU Memory (MB)} &\textbf{GPU-S} \\
        \midrule
        Variant 1  & $2.62 \times 10^{-2}$ &$270$ &$250$\\
        Variant 2      & $2.51 \times 10^{-2}$ & $270$&$256$\\
        Variant 3      & $1.33 \times 10^{-4}$ &$270$ & $340$\\
        Variant 4      & $8.50 \times 10^{-5}$ &$348$ &$391$\\
        \bottomrule
    \end{tabular}
\end{center}

\begin{figure}[htbp]
    \centering
    \includegraphics[width=1\linewidth]{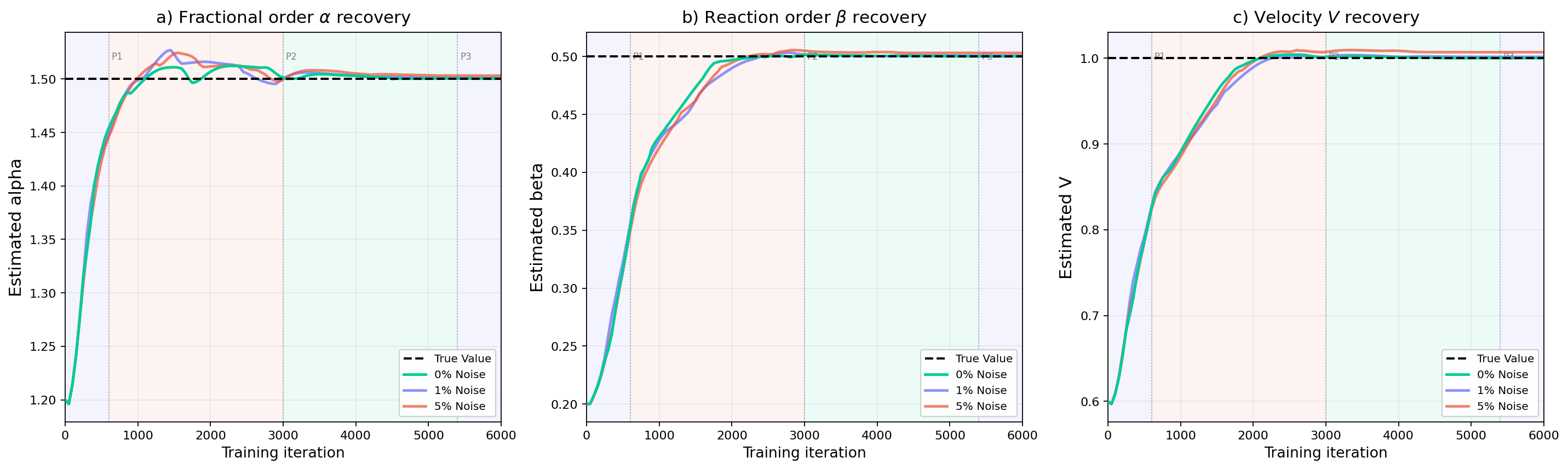}
    \caption{Convergence of Parameters for each noise level}
    \label{fig:p1_inv}
\end{figure}

A major benefit of PINNs over traditional solvers is the ability to estimate system parameters using sparse and noisy data. Our framework supports the estimation of parameters using the \texttt{set\_inverse()} method. For this problem, we choose $200$ data points for an edge. Since the above system has an exact solution, these points are chosen at random from each edge. For realistic settings, we systemically add Gaussian noise to the data set. For our experiment, we utilize the network parameters of Variant 4 and run the model for 6000 iterations with an initial guesses of $\hat\alpha=1.2$, $\hat\beta=0.2$ and $\hat V=0.6$. Table \ref{tab:p1_parameter_recovery} highlights the ability of the proposed framework to estimate system parameters and Figure \ref{fig:p1_inv} illustrates the convergence of the parameters from their initial guess to their final values. 

\begin{table}[htbp]
    \centering
    \caption{Simultaneous Parameter Recovery Profile under Synthetic Noise}
    \label{tab:p1_parameter_recovery}
    \begin{tabular}{lccc}
        \toprule
        \textbf{Noise ($\sigma$)} & \textbf{Estimated $\hat{\alpha}$} & \textbf{Estimated $\hat{\beta}$} & \textbf{Estimated $\hat{V}$} \\
        \midrule
        0\%          & $1.4954$ & $0.5032$ & $1.0014$ \\
        1\%        & $1.4929$ & $0.4997$ & $0.9983$ \\
        5\%        & $1.4947$ & $0.4961$ & $0.9938$ \\
        \bottomrule
    \end{tabular}
\end{table}

\subsection{Time-fractional diffusion equation on a Tadpole graph}
We consider the following time-fractional diffusion equation on a tadpole graph:
\begin{equation*}
    {}_{0}^{C}D_{t}^{\gamma}u_{i}(x,t)  - \frac{\partial^2u_i}{\partial x^2} = f_{i}(x,t), \quad x \in (0, L_{i}), \quad i \in \{1, 2, 3, 4\},~ t \in [0,1].
\end{equation*}
The graph topology employs Neumann constraints at the boundary nodes $v_3$ ($x=1$) and $v_4$ ($x=1$) such that $\frac{\partial u_2}{\partial x}(1,t) = \pi t^\gamma$ and $\frac{\partial u_3}{\partial x}(1,t) = \pi t^\gamma$, along with the continuity and Kirchhoff-Neumann constraints at node $v_1$ and $v_2$. The initial condition across all edges is set to $u_i(x, 0) = 0$. The forcing terms $f_i(x, t)$ has been chosen such that

\begin{align*}
    u_1(x, t) &= t^{\gamma} \left(\sin(\pi x) + \cos(\pi x)\right), && \text{for } x \in e_1 \text{ (Loop)} \\
    u_2(x, t) &= t^{\gamma} \left(\sin(\pi x) + \cos(\pi x)\right), && \text{for } x \in e_2 \text{ (Inner Tail)} \\
    u_3(x, t) &= t^{\gamma} \left(-\sin(\pi x) - \cos(\pi x)\right), && \text{for } x \in e_3 \text{ (Outer Leaf)} \\
    u_4(x, t) &= t^{\gamma} \left(-\sin(\pi x) + \cos(\pi x)\right), && \text{for } x \in e_4 \text{ (Branch Leaf)}
\end{align*}
forms the exact solution of the problem.

In this problem, we compare two different collocation strategies for approximating the Caputo fractional derivative. The first strategy follows the approach of Li et al.~\cite{li2026novel}, in which each training point is associated with an auxiliary graded mesh which is particularly used for approximating the fractional derivative. The primary training points and the auxiliary discretization are therefore decoupled, which allows the Caputo derivative to be computed using a locally constructed graded mesh at every collocation point. The second strategy is the propose unified mesh where a single graded mesh is used for both the neural network collocation points and the discretization of the fractional derivative. Consequently, the corresponding Caputo differentiation matrix is assembled only once during the initialization stage and reused throughout the optimization process. 

\begin{center}
    \captionof{table}{Training and Mesh Parameters}
    \label{tab:p1_param_para}
    \small
    \begin{tabular}{l c}
        \toprule
        \textbf{Parameter} & \textbf{Value} \\
        \midrule
        Spatial point density $(N_x/L_i)$ & $80$ \\
        Temporal points $(N_t)$ & $50$\\
        Temporal Grading (r) & $2$ \\
        Training Iterations   & $10000$\\
        Fractional scheme & $L2-1_{\sigma}$ \\
        Neurons per layer & 128 \\
        Hidden Layers& 4 \\
        Constraint type & Soft \\
        Adaptive Weight Strategy & Dual \\
        Singularity Capture & Enabled\\
        Fourier Dimension& 128\\
        Fourier $\sigma$ & 2\\
        \midrule
    \end{tabular}
\end{center}
The usage of auxiliary graded meshes poses significant challenges when adapting the approach to governing systems defined on quantum graphs. The computation of the Caputo derivative matrix at every training point introduces significant computational bottlenecks, especially when defining problems on complex real-world topologies. Therefore, in our framework we prioritize the usage of a single unified graded mesh where the Caputo fractional matrix is computed once during initialization as described in Section \ref{nonlocapprox}. We compare these two strategies to assess their influence on the accuracy and computational performance of the proposed framework. The neural network parameters for both collocation methodologies are given in Table \ref{tab:p1_param_para}.

As the auxiliary mesh implementation requires each training point to be supplemented with a unique auxiliary mesh, we choose 2 variations. Variant 1 supplements each training point with 50 auxiliary points, variant 2 supplements each training point with 25 auxiliary points. Each auxiliary mesh is graded with a grading factor of $r=4$.
\begin{center}
    \captionof{table}{Performance comparison of variants for $\gamma=0.5$}
    \label{tab:p2_comp}
    \footnotesize
    \begin{tabular}{@{}lccc@{}}
        \toprule
        \textbf{Variant} & \textbf{Global Relative $L_2$ Error} & \textbf{GPU Memory (MB)} &\textbf{GPU-S} \\
        \midrule
        Unified Mesh  & $7.73 \times 10^{-4}$ & $1146$& $473.6$ \\
        Auxiliary Mesh (50 points)     & $5.83 \times 10^{-4}$ &$6814$ &$2909.3$\\
        Auxiliary Mesh (25 points)     & $7.48 \times 10^{-4}$ & $3932$&$1694.0$\\
        \bottomrule
    \end{tabular}
\end{center}

\begin{center}
\captionof{table}{Loss values at Epoch 10000}
\label{loss_logs_p2}

\centering
\begin{tabular}{lcc}
\hline
\textbf{Mesh Type} & \textbf{PDE Loss} & \textbf{BC Loss} \\
\hline
50 Auxiliary Points & $7.63 \times 10^{-5}$ & $5.49 \times 10^{-5}$ \\
25 Auxiliary Points & $8.07 \times 10^{-5}$ & $8.52 \times 10^{-5}$ \\
Unified Mesh        & $3.36 \times 10^{-4}$ & $4.23 \times 10^{-5}$ \\
\hline
\end{tabular}

\end{center}

\begin{figure}[h]
	\begin{center}
		\centering
		\subfigure[]{%
        
			\includegraphics[scale=0.20]{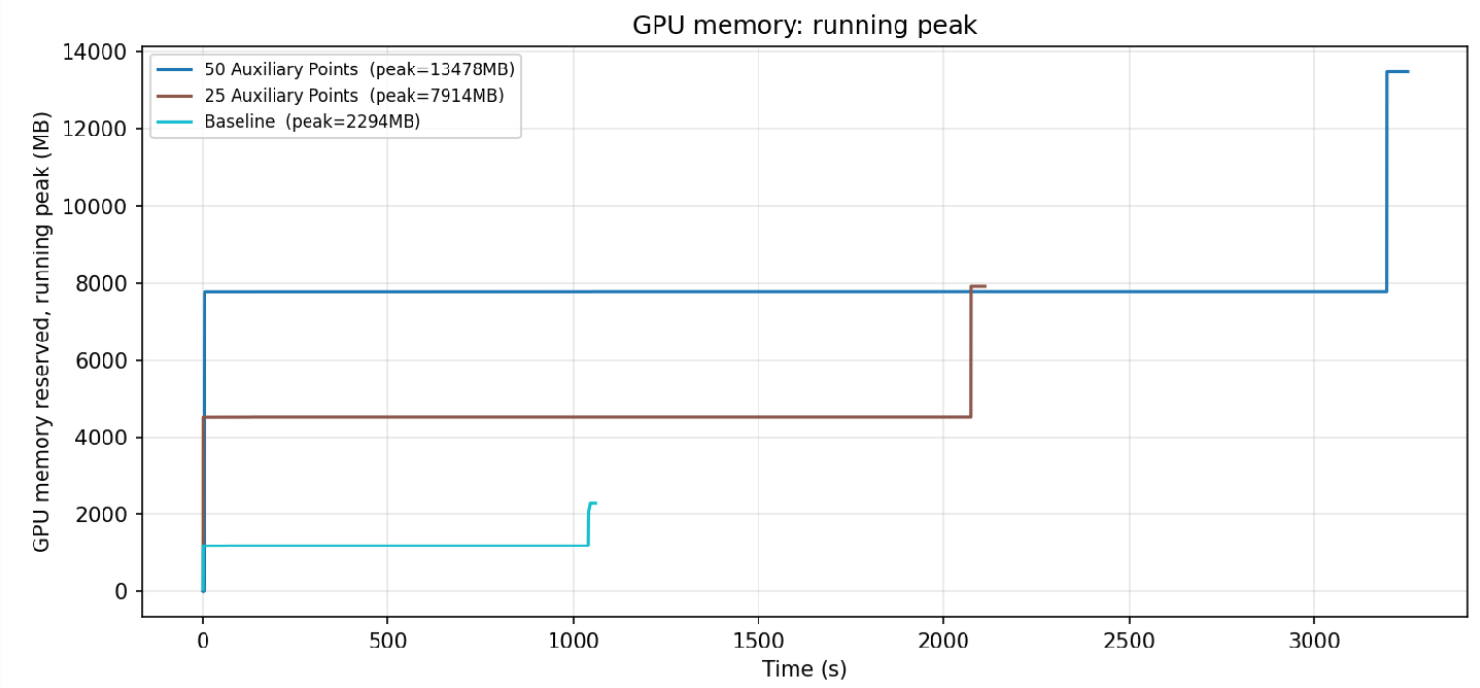}}
		\subfigure[]{%
        \label{fig:p2_usage}
		\includegraphics[scale=0.20]{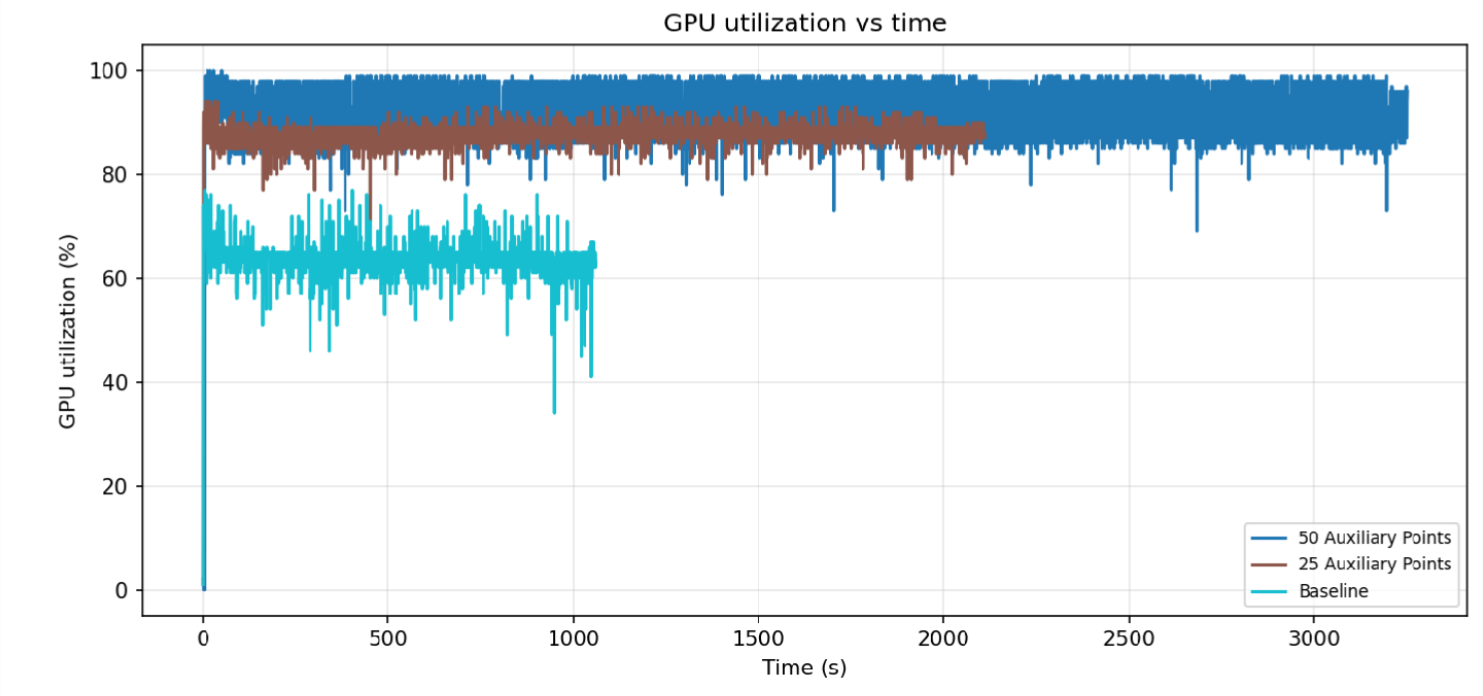}}
         \caption{Computational cost of each collocation methodology}
           \label{fig:p2_computational}
	\end{center}
\end{figure}

\begin{figure}[h]
    \centering
    \includegraphics[scale=0.32]{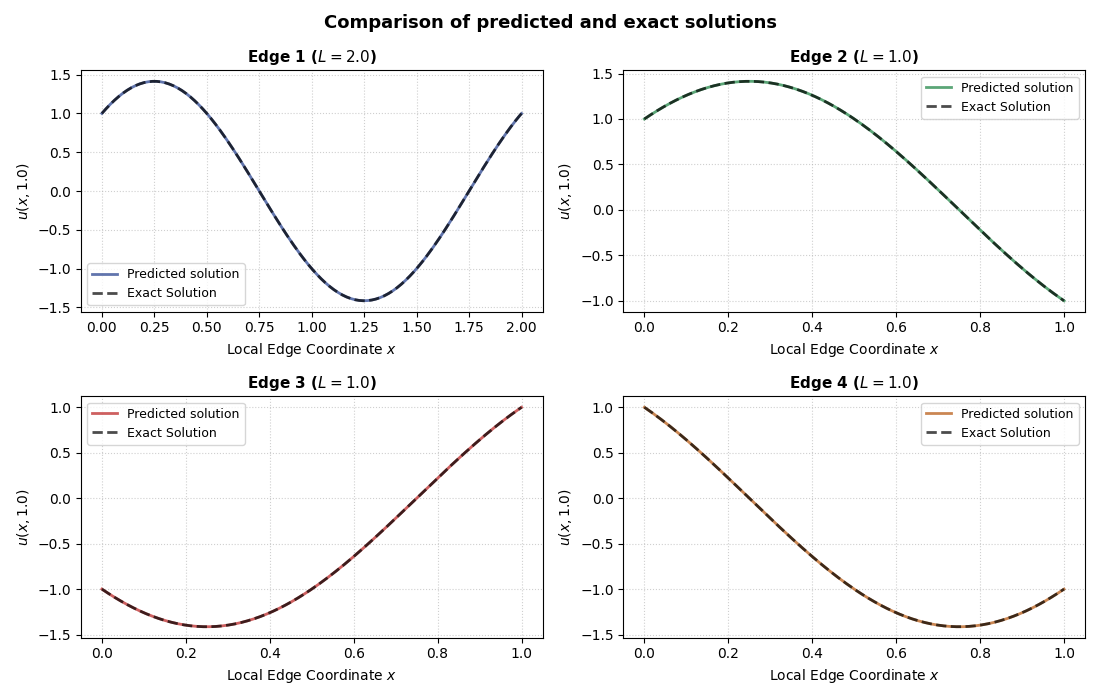}
    \caption{Predicted vs exact solution on each edge at $t=1s$.}
    \label{fig:p2_per_edge}
\end{figure}

Table~\ref{tab:p2_comp} provides the relative $L^2$ accuracy and computational cost of each collocation strategy. The results highlight the drawbacks of using auxiliary meshes in our framework. They provide very minimal performance benefits while requiring significantly more computational resources. Figure~\ref{fig:p2_computational} highlights the computational resource use of each variant. Their inability to provide superior accuracy is due to mesh decoupling. Computing the Caputo derivative using localized auxiliary meshes results in the network attempting to overfit to the local derivative computation. This is evident from the loss data shown in \ref{loss_logs_p2} where the auxiliary mesh with 50 points had a significantly lower PDE loss but slightly higher boundary condition loss. However, this lower PDE loss did not translate to an effective reduction in the approximation error. This is a consequence of the derivative overfitting as described above. This becomes more evident when comparing the auxiliary mesh with 25 points and the unified mesh approach.  We therefore promote the use of a single unified mesh which serves as an ideal tradeoff between approximation accuracy and computational efficiency. Figure~\ref{fig:p2_per_edge} illustrates the comparison of exact and predicted solution on each edge of the graph at $t=1$. It is evident that the predicted solution obtained from the proposed framework matches closely with the exact solution of the considered problem. 

Furthermore, the exact solution allows us to test the framework’s ability to estimate parameters from noisy and sparse data. We choose a spatial point density as $\frac{N_x}{L_i}=100$ and $N_t=200$ temporal points. We incorporate 200 data points from each edge into the loss function, with initial guesses of \(\hat{\gamma}=0.35\) and \(\hat{V}=0.6\). Similar to the elliptic problem \ref{ellipnum}, we simulate real world data for time-fractional diffusion problem by adding varying levels of Gaussian noise. Using the unified mesh strategy, we compare the accuracy of the $L1$ and $L2-1_{\sigma}$ fractional schemes, while all other neural network parameters remain identical to the forward problem. Table~\ref{tab:p2_parameter_recovery} demonstrates the framework’s ability to extract the governing parameters from noisy data. The neural network using the higher-order $L2-1_{\sigma}$ scheme achieved superior accuracy. Figure~\ref{fig:p2_parameter_history} illustrates the convergence of the estimated parameters to their final values during training for each fractional scheme.

\begin{table}[htbp]
    \centering
    \caption{Simultaneous Parameter Recovery Profile under Synthetic Noise}
    \label{tab:p2_parameter_recovery}
    \begin{tabular}{l cccc}
        \toprule
        & \multicolumn{4}{c}{\textbf{Fractional Scheme Used}} \\
        \cmidrule(lr){2-5}
        & \multicolumn{2}{c}{$L1$} & \multicolumn{2}{c}{$L2-1_\sigma$} \\
        \cmidrule(lr){2-3} \cmidrule(lr){4-5}
        \textbf{Noise ($\sigma$)} & \textbf{Estimated $\hat{\gamma}$} & \textbf{Estimated $\hat{V}$} & \textbf{Estimated $\hat{\gamma}$} & \textbf{Estimated $\hat{V}$} \\
        \midrule
        0\% & $0.4920$ & $1.0003$  & $0.4974$ & $1.0010$ \\
        1\% & $0.4918$ & $0.9855$ & $0.4979$ & $0.9925$ \\
        5\% & $0.4866$ & $0.9611$ & $0.4871$ & $0.9680$ \\
        \bottomrule
    \end{tabular}
\end{table}

\begin{figure}[htbp]
    \centering
    \subfigure[Estimated parameter value vs. iterations ($L1$ scheme)]{%
        \label{fig:p2_L1}
        \includegraphics[width=0.48\linewidth]{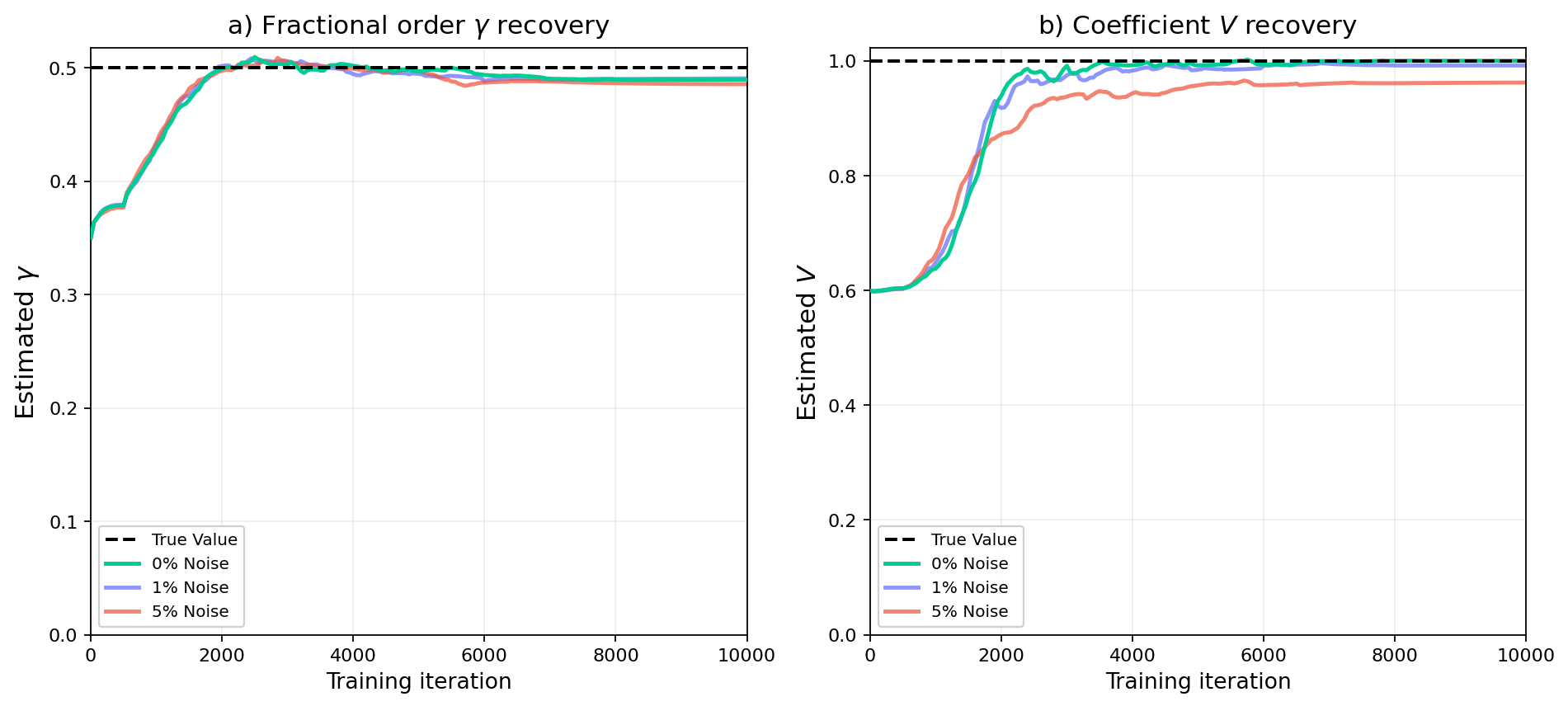}%
    }
    \hfill
    \subfigure[Estimated parameter value vs. iterations ($L2-1_\sigma$ scheme)]{
        \label{fig:p2_L21}
        \includegraphics[width=0.48\linewidth]{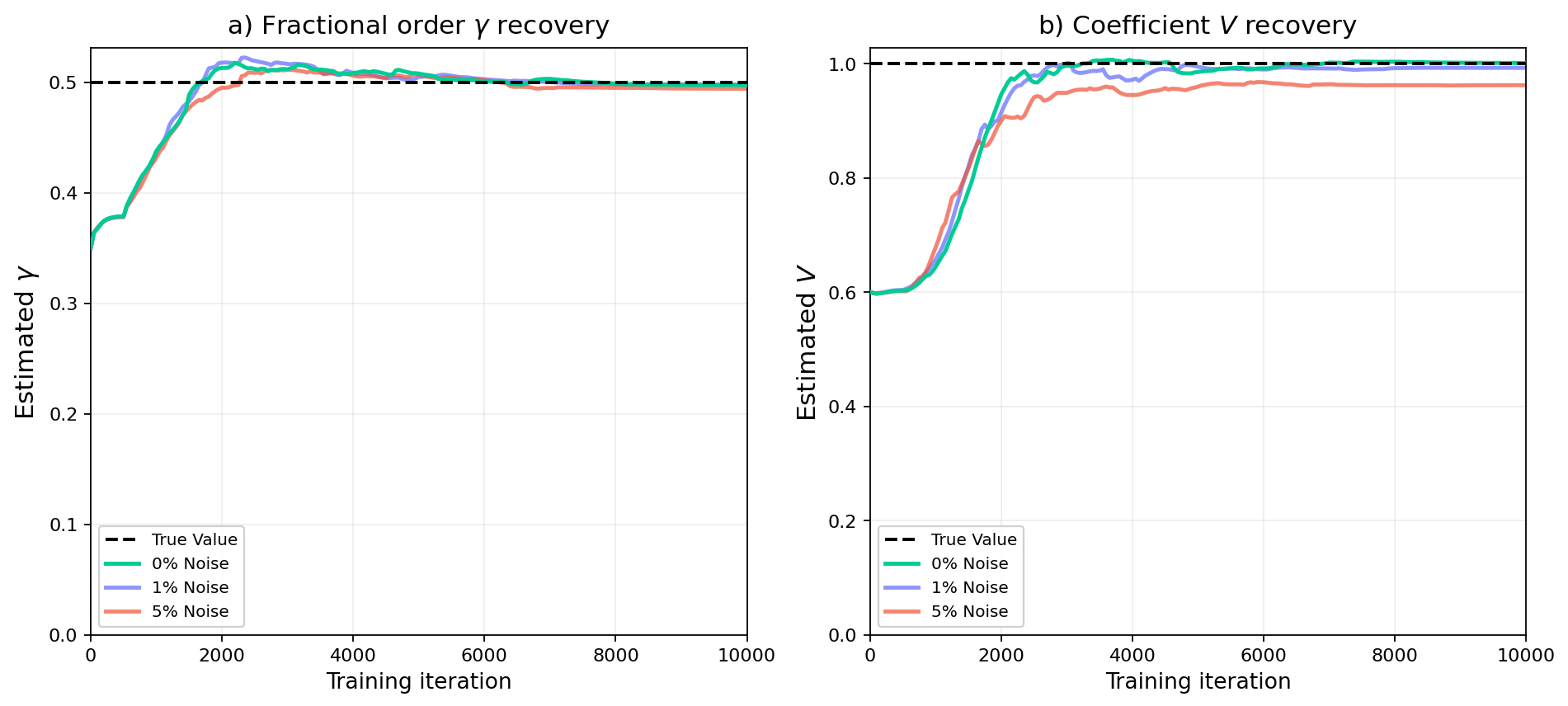}
    }
    \caption{Parameter convergence history.}
    \label{fig:p2_parameter_history}
\end{figure}

\subsection{Time-Fractional Burgers' Type Equation for Open-Channel Flow Routing}
\label{sec:p4_drain}

We simulate sub-diffusive water wave propagation in an open channel drainage system in Japan, following the network topology structured by Yoshioka et al. \cite{yoshioka2014burgers}. The system models a real world drainage system present in Japan as shown in Figure~\ref{fig:real_drainage}. The presence of heavy vegetation along the banks of the drainage system introduces anomalous wave propagation and memory effects that the Caputo derivative is suited to handle. The problem is described by embedding the real world topology on a connected metric graph~\ref{fig:drainage_topology} consisting of six boundary vertices and five edges (reaches), normalized over a physical length $L_i = 20.0\,\text{m}$ and a temporal window $T_{\max} = 10.0\,\text{s}$.

\begin{figure}[htbp]
  \centering

  \begin{minipage}[c]{0.48\textwidth}
    \centering
    \includegraphics[width=\textwidth]{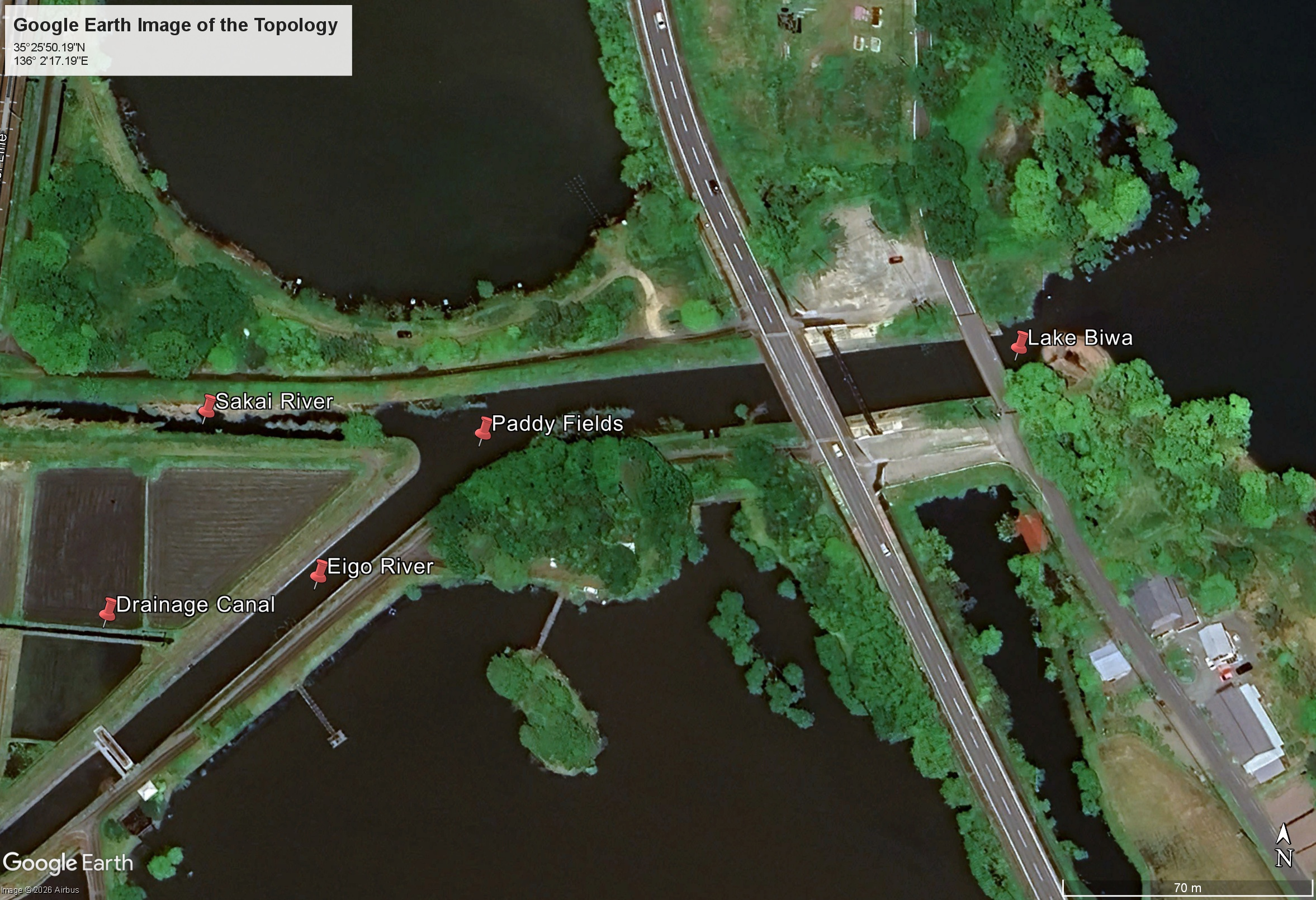}
    \caption{Satellite imagery of the  topology}
    \label{fig:real_drainage}
  \end{minipage}%
  \hfill%
  \begin{minipage}[c]{0.47\textwidth}
    \centering
    \resizebox{\textwidth}{!}{%
      \begin{tikzpicture}[
          vertex/.style={circle, draw, fill=white, minimum size=7pt, inner sep=0pt, thick},
          hub/.style={circle, draw, fill=black!80, minimum size=9pt, inner sep=0pt, thick},
          pendant/.style={circle, draw, fill=violet!30, minimum size=7pt, inner sep=0pt, thick},
          line/.style={draw, thick, -latex}, 
          lbl/.style={font=\small},
          edge_lbl/.style={fill=white, inner sep=1.2pt, font=\scriptsize, text=blue!80!black}
      ]
      
      \node[pendant, label={[lbl]above:$B$}] (1) at (0, 4) {};
      \node[pendant, label={[lbl]below:$C$}] (2) at (0, 0) {};
      \node[vertex, label={[lbl]above:$E$}]  (3) at (2.5, 2) {};
      \node[pendant, label={[lbl]above:$A$}] (0) at (5.0, 4) {};
      \node[vertex, label={[lbl]below:$F$}]  (4) at (5.0, 2) {};
      \node[hub, label={[lbl]right:$D$}]     (5) at (7.5, 2) {};

      \draw[line] (1) -- node[edge_lbl, above=1pt] {$E_0$} (3);
      \draw[line] (2) -- node[edge_lbl, below=1pt] {$E_1$} (3);
      \draw[line] (3) -- node[edge_lbl, above=1pt] {$E_2$} (4);
      \draw[line] (0) -- node[edge_lbl, right=1pt] {$E_3$} (4);
      \draw[line] (4) -- node[edge_lbl, above=1pt] {$E_4$} (5);
      
      \end{tikzpicture}%
    }
    \caption{Agricultural drainage network topology from Japan. Inlets are located at vertices $A$ (Sakai River), $B$ (Drainage Canal), and $C$ (Eigo River). The network terminates at downstream outlet $D$, which discharges into Lake Biwa.}
    \label{fig:drainage_topology}
  \end{minipage}

  \vspace{10pt}
  \addtocounter{figure}{-2} 
  \label{fig:combined-main}
\end{figure}
The governing differential equation for the depth fluctuation $u_i(x,t) = h_i(x,t) - h_0$ over a steady equilibrium depth $h_0 = 5.0\,\text{m}$ on edge $e_i$ is given by:
\begin{equation}
    {}_{0}^{C}D_{t}^{\gamma} u_{i}(x, t) + \frac{\partial}{\partial x} \Big[ V(u_i) u_i \Big] - D \frac{\partial^2 u_{i}}{\partial x^2} = 0, \quad x \in (0, L_i), \quad t \in (0, T_{\max}],
\label{eq:burgers_dimensional}
\end{equation}
where $\gamma = 0.85$ models sub-diffusive porous interactions along heavily vegetated banks, $D = 25.0\,\text{m}^2/\text{s}$ is the physical diffusion coefficient, and $g = 9.81\,\text{m}/\text{s}^2$. The advection parameter $V(h_i)$ is defined by \cite{yoshioka2014burgers}:
\begin{equation*}
    V(h_i) = \frac{2}{3}\sqrt{g}\left[(h_i + h_0)^{\frac{3}{2}} - h_0^{\frac{3}{2}}\right].
\end{equation*}

To map the spatial and temporal coordinates into unit intervals $[0, 1]$, we introduce the scaling variables $\hat{x} = x/L_i \in [0, 1]$ and $\hat{t} = t/T_{\max} \in [0, 1]$, where $L_i = 20.0\,\text{m}$ and $T_{\max} = 10.0\,\text{s}$. Substituting these transformation variables into Eq.~\eqref{eq:burgers_dimensional} and multiplying through by $(T_{\max})^\gamma$ yields the non-dimensional residual used for training:
\begin{equation}
    {}_{0}^{C}D_{\hat{t}}^{\gamma} u_{i}(\hat{x}, \hat{t}) + \Lambda_{\text{adv}} \frac{\partial}{\partial \hat{x}} \Big[ V(u_i) u_i \Big] - \Lambda_{\text{diff}} \frac{\partial^2 u_{i}}{\partial \hat{x}^2} = 0, \quad \hat{x} \in [0, 1], \quad \hat{t} \in (0, 1],
\label{eq:burgers_nd}
\end{equation}
where the advection and diffusion coefficients $\Lambda_{\text{adv}}$ and $\Lambda_{\text{diff}}$, respectively, preserve the physical scale:
\begin{equation*}
    \Lambda_{\text{adv}} = \frac{(T_{\max})^\gamma}{L_i}, \quad \Lambda_{\text{diff}} = \frac{(T_{\max})^\gamma D}{L_i^2}.
\label{eq:burgers_lambda_groups}
\end{equation*}

At $\hat{t}=0$, the channel network is initially at rest with zero water depth fluctuation, $u_i(\hat{x},0) = 0$. Dynamic high-energy boundary surges are applied at $\hat{t} > 0$ via time-dependent Dirichlet conditions at all inlets ($A, B, C$), that is
\begin{equation}
    u_3(0,\hat{t}) = u_0(0,\hat{t}) = u_1(0,\hat{t}) = 2.0 \sin^2\left(\pi \hat{t}\right).
\end{equation}
A Neumann boundary condition ($\frac{\partial u}{\partial \hat{x}} = 0$) is enforced at the downstream outlet vertex $D$. As our open channel drainage system is governed by advection, general flux conservation does not capture the true physical system. Therefore, we define a custom conservative flux operator $\hat{Q}_i(\hat{x}, \hat{t})$ that combines both the non linear advection transport and the diffusive wave motion along edge $i$:

\begin{equation}
    \hat{Q}_i(\hat{x}, \hat{t}) = \Lambda_{\text{adv}} V(u_i) u_i - \Lambda_{\text{diff}} \frac{\partial u_i}{\partial \hat{x}}
\end{equation}

The system satisfies the Kirchhoff-Neumann junction conditions using this complete flux operator. At any internal junction node $v$, the total flux in the incoming edges $E_{in}(v)$ must be equal to the flux in the outgoing edges $E_{out}(v)$. That is,

\begin{equation}
    \sum_{i \in E_{in}(v)} \hat{Q}_i(1, \hat{t}) = \sum_{j \in E_{out}(v)} \hat{Q}_j(0, \hat{t})
\end{equation}

To define such a custom flux, our framework uses the inbuilt \texttt{get\_flux} method. This allows the user to define a flux operator. The neural network parameters for this problem are given in table~\ref{tab:p3_parameters}

\begin{center}
\captionof{table}{Training and mesh parameters for time-fractional Burgers' equation.}
\label{tab:p3_parameters}
\begin{tabular}{l c}
\toprule
\textbf{Parameter} & \textbf{Value} \\
\midrule
Network topology & 6 nodes, 5 edges \\
Spatial point density($\frac{N_x}{L_i}$) & $200$ \\
Temporal points($N_t)$ & $200$ \\
Temporal grading (r) & $2$ \\
Training iterations & $15000$ \\
Fractional scheme & $L2-1_{\sigma}$ \\
Neurons per layer & $128$ \\
Hidden layers & $4$ \\
Constraint type & Soft \\
Adaptive weight strategy & Dual \\
Singularity capture & Enabled \\
Fourier dimension & $128$ \\
\bottomrule
\end{tabular}
\end{center}

Since the system does not have an exact solution, we verify the numerical solution using secondary physical constraints. In classical (integer-order) systems where the instantaneous flux is defined by $Q_{net}=\frac{dM}{dT}$, the following conservation law holds:
\begin{equation*}    M(t)-M(0)=\int_0^tQ(\tau)d\tau
\end{equation*}
However, the considered system is sub-diffusive, thus, the rate of mass accumulation is not instantaneous due to the presence of the Caputo derivative. This models the phenomenon of the vegetative banks absorbing, storing and gradually releasing water over time. Assuming water density and channel width are constant with time, the above quantity can serve as a proxy for global mass fluctuation, and hence 
satisfies the global fractional integral conservation balance:
\begin{equation}\label{fraccons}
\hat{M}(\hat{t}) - \hat{M}(0) = \mathcal{I}_{\hat{t}}^\gamma (\hat{Q}_{\text{net}})=\frac{1}{\Gamma({\gamma})}\int_{0}^{\hat{t}}(\hat{t}-\tau)^{\gamma-1}\hat{Q}_{\text{net}}(\tau) d\tau,
\end{equation}
where $\hat{Q}_{\text{net}}(\hat{t}) = \big(\hat{Q}_0(0,\hat{t}) + \hat{Q}_1(0,\hat{t}) + \hat{Q}_3(0,\hat{t})\big) - \hat{Q}_4(1,\hat{t})$ represents the net boundary flux into the system. The global stored volume fluctuation per unit width $M(t)$ is defined as
\begin{equation*}
    \hat{M}(\hat{t}) = \sum_{i=0}^{4} \int_{0}^{1} u_i(\hat{x}, \hat{t}) \, d\hat{x}.
\end{equation*}

\begin{table}[htbp]
    \centering
    \caption{Open-System Mass Conservation, $\gamma=0.85$}
    \label{tab:p3_mass_conservation}
    \begin{tabular}{lccc}
        \toprule
        \textbf{Time} & \textbf{$\hat{M}(\hat{t})-\hat{M}(0)$} & \textbf{$\mathcal{I}^\gamma[\hat{Q}_{\text{net}}](\hat{t})$} & \textbf{Rel. Error (\%)} \\
        \midrule
        $\hat{t}=0.20$ & $5.9106 \times 10^{-1}$ & $5.9061 \times 10^{-1}$ & $0.077$ \\
        $\hat{t}=0.50$ & $8.6204 \times 10^0$ & $8.6190 \times 10^0$ & $0.016$ \\
        $\hat{t}=0.80$ & $7.4173 \times 10^0$ & $7.3766 \times 10^0$ & $0.549$ \\
        $\hat{t}=1.00$ & $4.0244 \times 10^0$ & $4.0561 \times 10^0$ & $0.790$ \\
        \bottomrule
    \end{tabular}
\end{table}

\begin{table}[htbp]
    \centering
    \caption{Junction Continuity Residuals, $\gamma=0.85$}
    \label{tab:p3_junction_residuals}
    \begin{tabular}{llccc}
        \toprule
        \textbf{Junction} & \textbf{Time} & \textbf{$\hat{Q}_{in}$} & \textbf{$\hat{Q}_{out}$} & \textbf{Rel. Residual (\%)} \\
        \midrule
        $E$  & $\hat{t}=0.20$ & $2.0283 \times 10^{-2}$ & $2.1648 \times 10^{-2}$ & $6.308$ \\
        $E$  & $\hat{t}=0.50$ & $1.6337 \times 10^1$ & $1.6341 \times 10^1$ & $0.026$ \\
        $E$ & $\hat{t}=0.80$ & $7.8457 \times 10^0$ & $7.8419 \times 10^0$ & $0.048$ \\
        $E$ & $\hat{t}=1.00$ & $2.0642 \times 10^0$ & $2.0688 \times 10^0$ & $0.224$ \\
        $F$  & $\hat{t}=0.50$ & $1.1534 \times 10^1$ & $1.1523 \times 10^1$ & $0.093$ \\
        $F$  & $\hat{t}=0.80$ & $1.4285 \times 10^1$ & $1.4284 \times 10^1$ & $0.006$ \\
        $F$  & $\hat{t}=1.00$ & $5.1354 \times 10^0$ & $5.1481 \times 10^0$ & $0.248$ \\
        \bottomrule
    \end{tabular}
\end{table}
The derivatives and integrals involved in the above expressions are evaluated using automatic differentiation and \texttt{scipy.integrate} respectively. Table~\ref{tab:p3_mass_conservation} describes the mass conservation metrics. Since deviations from these conservation laws are not explicitly penalized in the QGPINNs framework during training, the small observed errors indicate that the framework is able to recover physically consistent behaviour from the governing residuals and graph constraints. Moreover, the right-hand side of the expression \eqref{fraccons} requires fractional integration of order $\gamma$, and therefore provides an additional diagnostic for verifying fractional-order dynamics captured by the model. In Table~\ref{tab:p3_junction_residuals}, we examine the Kirchhoff-Neumann and continuity constraints at the intermediate junctions $E$ and $F$ by measuring flow residuals across various temporal levels. 

The significant deviation at junction $E$ at $\hat t=0.2$ can be observed as the wave has not propagated to the junction, therefore, the small absolute residual \((\widehat Q_{\mathrm{out}}(0.2)-\widehat Q_{\mathrm{in}}(0.2))=1.36\times 10^{-3}\) appears to be a substantial large when measured relative to the total flux. Once the wave propagates to the junction and reaches non-negligible levels, the underlying residual is reduced to below $0.3\%$. 

\begin{figure}[htbp]
	\begin{center}
		\centering
		\subfigure[]{%
       \label{fig:water_fluc}
			  \includegraphics[width=0.36\linewidth]{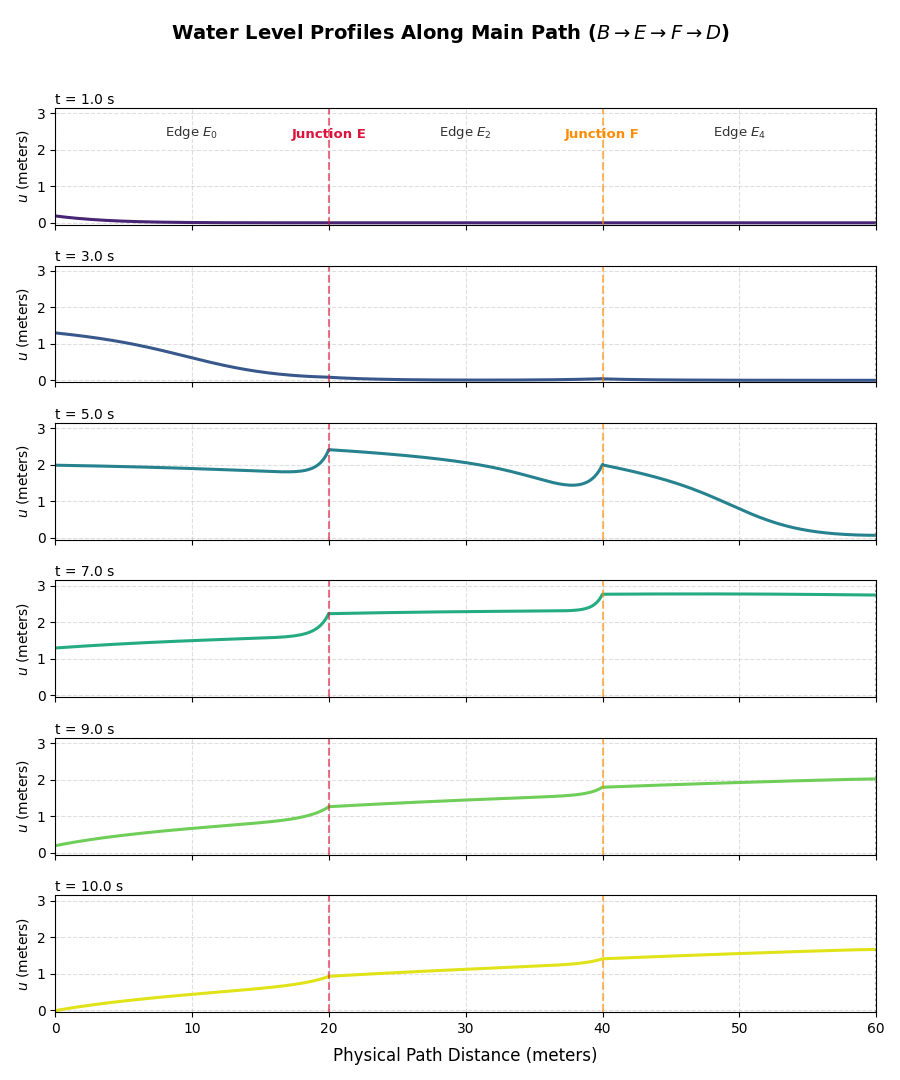}}
		\subfigure[]{%
        \label{fig:heatmap_p3}
		 \includegraphics[width=0.46\linewidth]{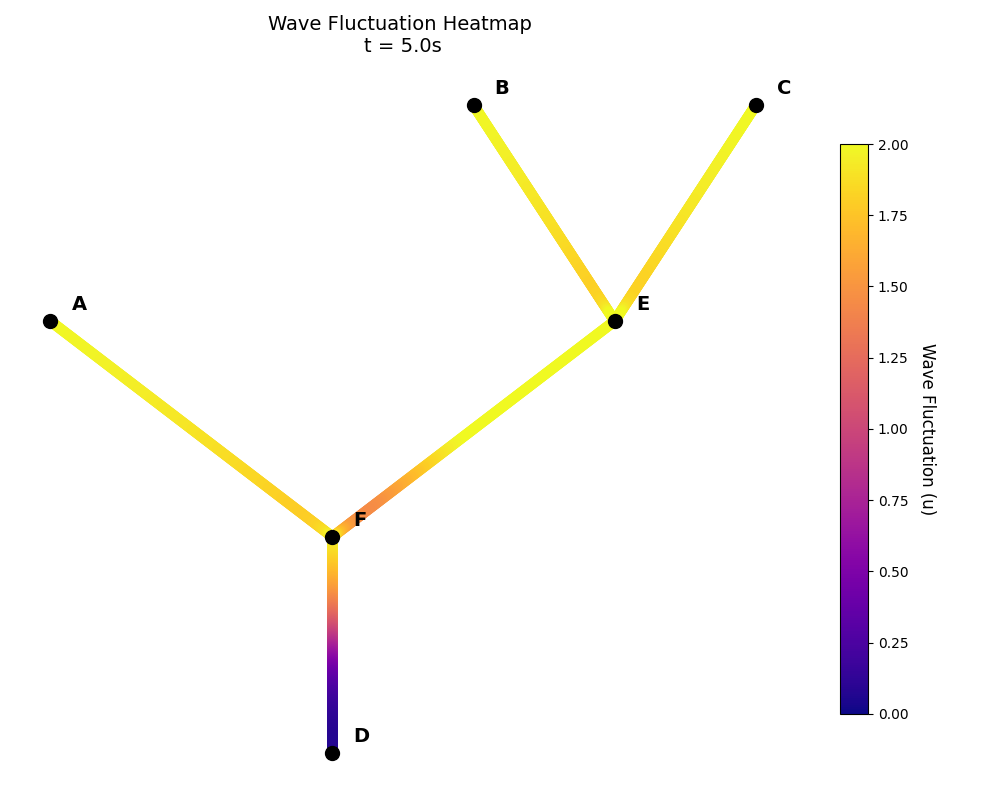}}
         \caption{(a) Water level profile along the path $B\to E \to F \to D$ at time $t$ after reversing the normalization. (b) Heatmap representing the water height fluctuation at time $t=5s$.}
	\end{center}
\end{figure}

Figure~\ref{fig:water_fluc} illustrates the water level fluctuation along the path $B\to E \to F \to D$. The normalization is reversed in order to interpret the solution on the original real-world topology. The behaviour of the predicted solution is consistent with the imposed junction constraints. The sharp upward kinks, for example at junction \(E\) when \(t=5s\), are the consequence of the periodic influx of water from multiple inlet branches, which causes a change in slope across the junction. The same qualitative behaviour is observed at junction F, where the two incoming branches combine to supply the downstream reach. This problem demonstrates the capability of the framework to simulate complex real-world network dynamics in the absence of an exact solution. Figure~\ref{fig:heatmap_p3} shows the water height fluctuation at $t=5s$, where the qualitative effect of the Kirchhoff--Neumann flux balance can be observed across the network.

\subsection{Fractional Telegraph Model for High-Frequency Surge Propagation on a Transmission Network}
\label{sec:p5_telegraph}

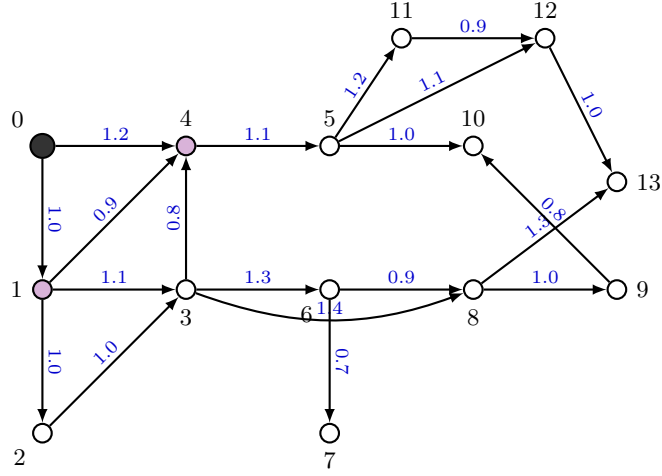
\begin{figure}[h]
    \centering
    \begin{tikzpicture}[
        scale=0.95,
        vertex/.style={circle, draw, fill=white, minimum size=7pt, inner sep=0pt, thick},
        hub/.style={circle, draw, fill=black!80, minimum size=9pt, inner sep=0pt, thick},
        pendant/.style={circle, draw, fill=violet!30, minimum size=7pt, inner sep=0pt, thick},
        line/.style={draw, thick, -latex},
        lbl/.style={font=\small},
        edge_lbl/.style={font=\scriptsize, text=blue!80!black, sloped, above, inner sep=2pt}
    ]

    \node[hub, label={[lbl]above left:$0$}]    (0) at (0, 4) {};
    \node[pendant, label={[lbl]left:$1$}]      (1) at (0, 2) {};
    \node[vertex, label={[lbl]below left:$2$}] (2) at (0, 0) {};
    \node[vertex, label={[lbl]below:$3$}]      (3) at (2, 2) {};
    \node[pendant, label={[lbl]above:$4$}]      (4) at (2, 4) {};
    \node[vertex, label={[lbl]above:$5$}]      (5) at (4, 4) {};
    \node[vertex, label={[lbl]below left:$6$}] (6) at (4, 2) {};
    \node[vertex, label={[lbl]below:$7$}]      (7) at (4, 0) {};
    \node[vertex, label={[lbl]below:$8$}]      (8) at (6, 2) {};
    \node[vertex, label={[lbl]right:$9$}]      (9) at (8, 2) {};
    \node[vertex, label={[lbl]above:$10$}]     (10) at (6, 4) {};
    \node[vertex, label={[lbl]above:$11$}]     (11) at (5, 5.5) {};
    \node[vertex, label={[lbl]above:$12$}]     (12) at (7, 5.5) {};
    \node[vertex, label={[lbl]right:$13$}]     (13) at (8, 3.5) {};

    \draw[line] (0) -- node[edge_lbl] {1.0} (1);
    \draw[line] (0) -- node[edge_lbl] {1.2} (4);
    
    \draw[line] (1) -- node[edge_lbl] {1.0} (2);
    \draw[line] (1) -- node[edge_lbl] {1.1} (3);
    \draw[line] (1) -- node[edge_lbl] {0.9} (4);
    
    \draw[line] (2) -- node[edge_lbl] {1.0} (3);
    
    \draw[line] (3) -- node[edge_lbl] {0.8} (4);
    \draw[line] (3) -- node[edge_lbl] {1.3} (6);
    \draw[line] (3) to[bend right=20] node[edge_lbl] {1.4} (8); 
    
    \draw[line] (4) -- node[edge_lbl] {1.1} (5);
    
    \draw[line] (5) -- node[edge_lbl] {1.0} (10);
    \draw[line] (5) -- node[edge_lbl] {1.2} (11);
    \draw[line] (5) -- node[edge_lbl] {1.1} (12);
    
    \draw[line] (6) -- node[edge_lbl] {0.7} (7);
    \draw[line] (6) -- node[edge_lbl] {0.9} (8);
    
    \draw[line] (8) -- node[edge_lbl] {1.0} (9);
    \draw[line] (8) -- node[edge_lbl] {1.3} (13);
    
    \draw[line] (9) -- node[edge_lbl] {0.8} (10);
    
    \draw[line] (11) -- node[edge_lbl] {0.9} (12);
    
    \draw[line] (12) -- node[edge_lbl] {1.0} (13);

    \end{tikzpicture}
    \caption{IEEE 14-Bus System Topology (0-Indexed)}
    \label{fig:ieee14_network}
\end{figure}

This problem aims to test the robustness of our framework on complex real-world topologies. We simulate the propagation of a high frequency, finite speed wave across an electric transmission network. Such phenomena are characteristic of voltage surges due to lightning strikes and other such events. Such systems undergo frequency attenuation, where the high frequency characteristics dampen over time. Standard integer-order wave models fail to capture these power-law dissipation dynamics. We consider the well-known IEEE-14 bus network as the metric graph benchmark shown in Figure \ref{fig:ieee14_network}. The governing differential equation is a time-fractional telegraph equation that models the surge voltage $u_i(x,t)$ on edge $e_i$ . Each edge present in the IEEE-14 system has a corresponding wave $c_i>0$.

\begin{equation}
\frac{\partial^2 u_i}{\partial t^2}(x,t) 
+ \eta_i \, {}_{0}^{\mathrm{C}}\mathcal{D}_t^{\gamma} u_i(x,t) 
- c_i^2 \frac{\partial^2 u_i}{\partial x^2}(x,t) = 0, 
\qquad x \in (0,L_i), \; t \in (0,T].
\label{eq:telegraph_network}
\end{equation}
We choose the fractional order $\gamma=0.5$ and similar to problem considered in Section \ref{sec:p4_drain}, the system is normalized before training. The spatial coordinate is handled by the engine's per-edge input normalization $\hat x = x/L_i$ while time is mapped to $\hat t = t/T \in [0,1]$, with $T=5.0\,\text{s}$. Substituting into Eq.~\eqref{eq:telegraph_network} and multiplying through by $T^2$ gives the residual used in training,
\begin{equation}
\frac{\partial^2 u_i}{\partial \hat t^2} + \Lambda_{\mathrm{damp},i}\, {}_{0}^{\mathrm C}\mathcal D_{\hat t}^\gamma u_i - \Lambda_{\mathrm{wave},i} \frac{\partial^2 u_i}{\partial x^2} = 0,
\label{eq:telegraph_network_nd}
\end{equation}
where
\begin{equation*}
\Lambda_{\mathrm{damp},i} = \eta_i\, T^{\,2-\gamma}, \qquad \Lambda_{\mathrm{wave},i} = c_i^2\, T^2.
\label{eq:telegraph_lambda_groups}
\end{equation*}

To simulate a lightning strike, we impose a spatial initial surge profile centered at vertex $v_0$ (node 0) such that for all edges $e_k$ incident to $v_0$ we specify
\begin{equation}
u_k(x,0) = A \exp\left[-\left(\frac{x}{\lambda_i}\right)^2\right], \qquad \frac{\partial u_i}{\partial t}(x,0) = 0,
\label{eq:telegraph_ic_spike}
\end{equation}
where $A = 1$ is the peak surge amplitude and $\lambda_i= 0.20$ controls the spatial penetration length along edge $e_i$. The value of these constants are chosen to ensure that the pulse quickly decays to machine precision within the spatial domain of the initial edge. Therefore, edges $e_j$ not incident to $v_0$ are initialized with homogeneous initial conditions, that is, $u_j(x,0) = 0$ and $\frac{\partial u_j}{\partial t}(x,0) = 0$. At all the junction nodes, Kirchhoff-Neumann and continuity conditions have been imposed to ensure that the solution is mathematically valid. The governing system contains a classical second-order time derivative. While our engine does not natively support such derivatives, the addition of a run script helper class, which computes and passes the required derivative using \texttt{torch.autograd()}, served as enough of a modification, which also highlights the flexibility of the given framework. 

However, we emphasise that the singularity capturing feature describe in Section \ref{sec:singularity} is incompatible with the considered problem. The term $z(t)=t^{0.5}$, when passed through the automatic differentiator in time, results in $t^{-\frac{3}{2}}$. This resulting term blows up to incredibly high values near $t=0$, creating artificial gradients. Therefore, specifically for this problem, we disable the singularity capturing feature. The neural network parameters for the problem are given in Table \ref{tab:p5_telegraph_parameters}.

\begin{center}
\captionof{table}{Training and mesh parameters for the fractional Telegraph problem.}
\label{tab:p5_telegraph_parameters}
\begin{tabular}{l c}
\toprule
\textbf{Parameter} & \textbf{Value} \\
\midrule
Network topology & IEEE 14-bus, 20 edges \\
Spatial point density ($\frac{N_x}{L_i}$) & $60$ \\
Temporal points ($N_t$) & $80$ \\
Temporal grading & $2$  \\
Training iterations & $15000$ \\
Fractional scheme & $L2-1_{\sigma}$ \\
Neurons per layer & $128$ \\
Hidden layers & $4$ \\
Constraint type & Soft \\
Adaptive weight strategy & Dual \\
Singularity capture & Disabled \\
Fourier dimension & $128$ \\
Fourier $\sigma$ & $1$ \\
\bottomrule
\end{tabular}
\end{center}

\begin{figure}[htbp]
    \centering
    \includegraphics[width=0.7\linewidth]{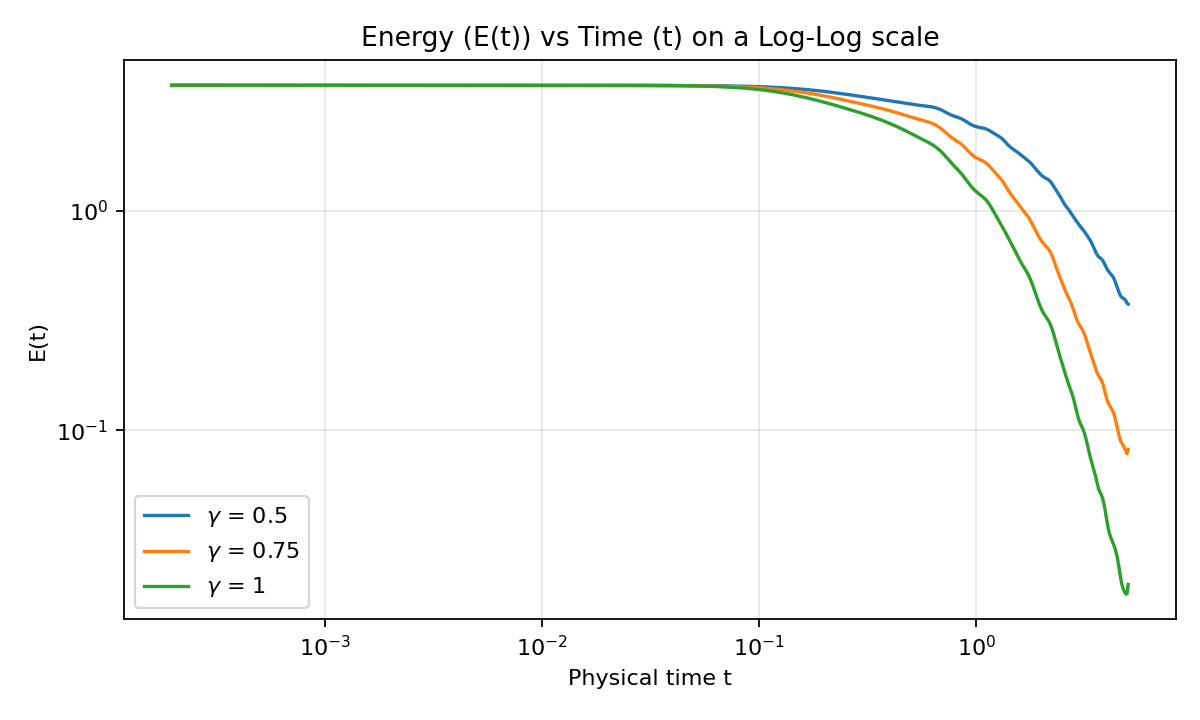}
    \caption{Plot of energy vs time in Log-Log axes}
    \label{fig:loglogp4}
\end{figure}

To verify physical consistency, we track the total mechanical-type network energy functional:
\begin{equation}
E(t) = \frac{1}{2} \sum_{i=1}^{20} \int_{0}^{L_i}
\left[
\left(\frac{\partial u_i}{\partial t}(x,t)\right)^2
+
c_i^2
\left(\frac{\partial u_i}{\partial x}(x,t)\right)^2
\right] dx.
\label{eq:telegraph_energy}
\end{equation}

\begin{table}[htbp]
    \centering
    \caption{Comprehensive Physical-Consistency and Energy Diagnostics Across Fractional Orders $\gamma$.}
    \label{tab:p5_global_verification}
    \begin{tabular}{lccc}
        \toprule
        \textbf{Metric / Quantity} & \textbf{$\gamma = 0.50$} & \textbf{$\gamma = 0.75$} & \textbf{$\gamma = 1.00$} \\
        \midrule
        \multicolumn{4}{l}{\textbf{Initial-Boundary Conditions}} \\
        Initial Pulse Relative RMS Error & $0.388\%$ & $0.558\%$ & $0.467\%$ \\
        Initial Pulse $L_2$ Error & $4.206 \times 10^{-4}$ & $6.038 \times 10^{-4}$ & $5.057 \times 10^{-4}$ \\
        Initial Velocity $L_2$ Error & $1.516 \times 10^{-3}$ & $2.488 \times 10^{-3}$ & $2.324 \times 10^{-3}$ \\
        \midrule
        \multicolumn{4}{l}{\textbf{Network Transmission Constraints}} \\
        Junction Continuity $L_2$ Error & $3.707 \times 10^{-3}$ & $3.872 \times 10^{-3}$ & $2.493 \times 10^{-3}$ \\
        Physical Kirchhoff Flux $L_2$ Residual & $9.375 \times 10^{-4}$ & $1.494 \times 10^{-3}$ & $1.109 \times 10^{-3}$ \\
        Grounded Terminal (Node 0) $L_2$ Residual & $1.598 \times 10^{-3}$ & $1.804 \times 10^{-3}$ & $1.300 \times 10^{-3}$ \\
        \midrule
        \multicolumn{4}{l}{\textbf{Global Energy Physics}} \\
        Initial Network Energy $E(0)$ & $3.728$ & $3.726$ & $3.737$ \\
        Final Network Energy $E(T_{\text{max}})$ & $0.376$ & $0.082$ & $0.020$ \\
        Energy Fraction Remaining ($E_{\text{final}} / E_{\text{initial}}$) & $10.09\%$ & $2.21\%$ & $0.53\%$ \\
        Energy Monotonicity Fraction & $98.74\%$ & $98.11\%$ & $98.74\%$ \\
        Late-Time Log-Log Slope & $-1.564$ & $-2.599$ & $-3.523$ \\
        \midrule
        \multicolumn{4}{l}{\textbf{Pulse Propagation}} \\
        Peak Surge Amplitude ($t = 0.0\,\text{s}$) & $1.001$ & $1.001$ & $1.002$ \\
        Peak Surge Amplitude ($t = 0.05\,\text{s}$) & $0.492$ & $0.487$ & $0.483$ \\
        Peak Surge Amplitude ($t = 0.10\,\text{s}$) & $0.463$ & $0.446$ & $0.427$ \\
        Peak Surge Amplitude ($t = 0.20\,\text{s}$) & $0.412$ & $0.373$ & $0.335$ \\
        Peak Surge Amplitude ($t = 1.00\,\text{s}$) & $0.116$ & $0.068$ & $0.043$ \\
        \bottomrule
    \end{tabular}
\end{table}

The energy is computed after reversing the normalization, and the
spatial integrals are computed numerically using
\texttt{scipy.integrate}. Since the underlying system is dissipative, the total network energy is expected to be non-increasing in time due to frequency attenuation. We therefore plot \(E(t)\) as a diagnostic quantity. In addition to the energy behaviour, we also verify junction continuity and Kirchhoff-Neumann flux conditions over the
temporal validation grid. We also compute several global physical quantities and report their deviations in Table~\ref{tab:p5_global_verification}. Similar to problem~\ref{sec:p4_drain}, the majority of these quantities were not explicitly minimized as part of the loss function during training. Hence, their small deviations depicted in Table \ref{tab:p5_global_verification} shows that the learned solution from QGPINNs is physically consistent. The framework captures the initial driving pulse with high accuracy, while maintaining the continuity and Kirchhoff--Neumann constraints across the large graph topology. A particularly important attribute is the global energy decay. As shown in Figure~\ref{fig:loglogp4}, the expected non-increasing behaviour of \(E(t)\) is observed over \(98.74\%\) of the temporal domain. The same figure also compares the late-time log--log decay slopes for the fractional orders \(0.5\) and \(0.75\) and the integer-order case \(1.0\). The slower energy dissipation observed for the fractional-order models indicates that QGPINNs captures an algebraic power-law decay rather than the standard exponential decay associated with integer-order dynamics~\cite{mainardi2022fractional}.

\begin{figure}[htbp]
    \centering
    \includegraphics[width=1\linewidth]{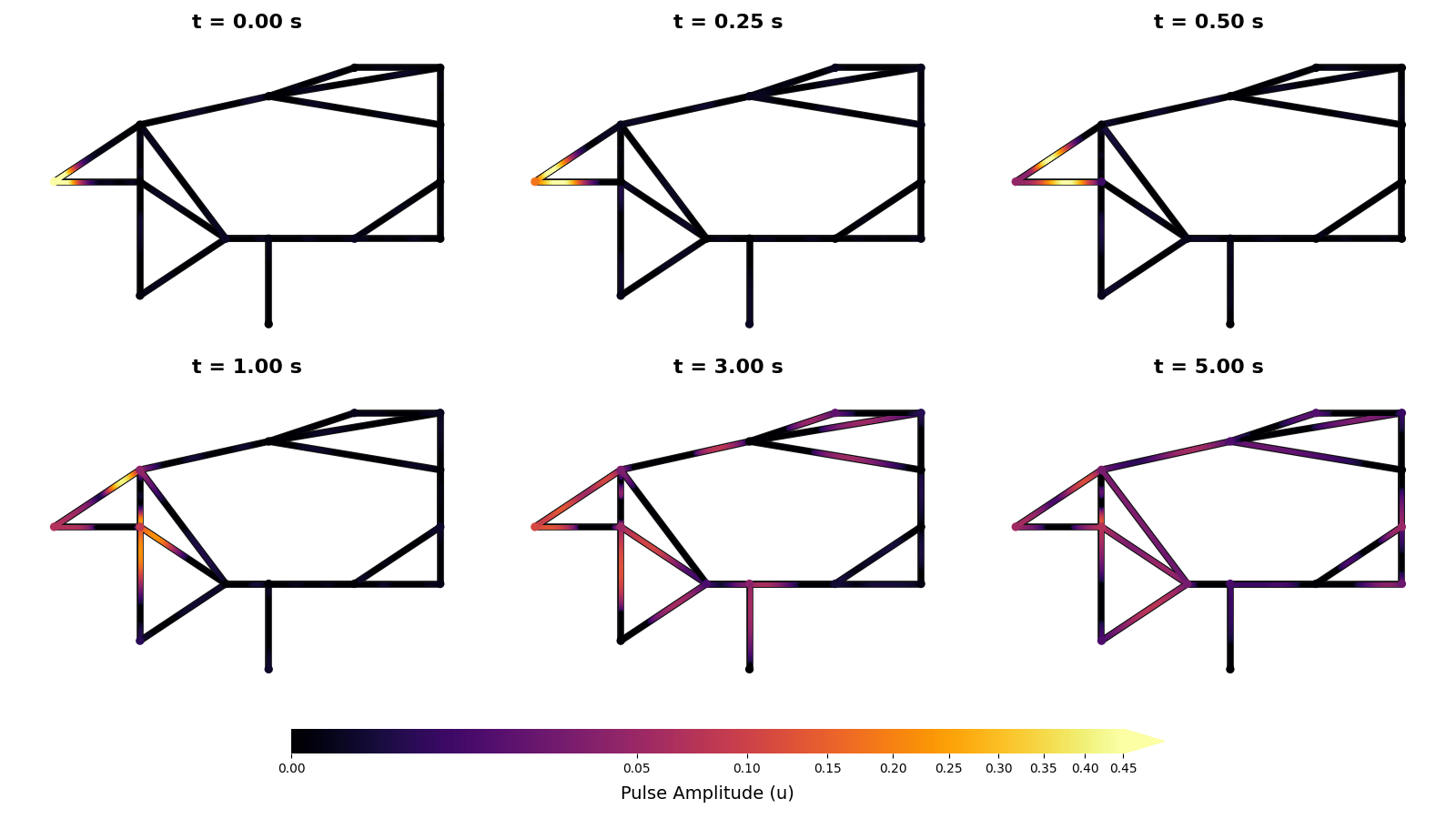}
    \caption{Propogation of the voltage surge at time $t$}
    \label{fig:p4_waves_diagram}
\end{figure}

We also emphasise that GPU memory limitations restricted the spatial and temporal point densities on such large topologies. Neverthless, the observed loss trends suggest that increasing the number of validation and training points may further improve the accuracy of the QGPINNs. Figure~\ref{fig:p4_waves_diagram} illustrates the propagation of the voltage spike over time. The spatial arrangement in Figure~\ref{fig:p4_waves_diagram} differs slightly from the graph shown in Figure~\ref{fig:ieee14_network} due to automatic graph layout generated by \texttt{networkx}, which dynamically optimize node distribution and canvas usage. Despite this difference in layout, the movement of the pulse along the graph edges and the vertices can be clearly observed.

\section{Conclusion}

The proposed QGPINN framework provides a physics-informed deep learning approach for computing the numerical solutions to nonlocal differential equations on quantum graphs. While fractional differential equations have been widely studied on Euclidean domains, their extension to quantum graphs has primarily been addressed through traditional numerical methods. The proposed framework provides a physics-informed alternative by combining edge-wise neural approximations with a unified graph-based loss formulation.

The addition of the various optimization schemes and architectural strategies described in section \ref{main} allows the framework to provide superior accuracy when compared to standard PINN frameworks without substantially increasing the computational cost. The numerical experiments in Section \ref{numerical} demonstrates the robustness of the method on a range of graph structures. The QGPINNs framework successfully computed the numerical solution to fractional differential equations defined on real world complex topologies. Although the present implementation focuses predominantly on elliptic and time-fractional evolution equations given in \eqref{eq:fbvp} and \eqref{eq:parabolic}, respectively, the modular design of the framework remain compatible to a broad class of problem structures and graph topologies, as illustrated in Problem \ref{sec:p5_telegraph}. This modularity also allows users to solve such systems without requiring extensive familiarity with the internal details of PyTorch.

Although the design of the QGPINNs reduces the need for problem specific calibration, the choice of the appropriate adaptive weighting strategy, the number of Fourier dimensions, and the size of the neural network may still depend on the particular application. The framework initializes an edge-wise neural network to each edge instead of a single neural network over the entire graph, which allows it to capture the edge-dependent behaviour and the memory effects associated with nonlocal Caputo derivatives. However, such design may not be computationally viable on extremely large networks. Future work in this direction will focus on incorporating fewer problem specific parameters and also improving the computational efficiency of the framework to allow it to handle large and more complex topologies.

\section*{Acknowledgments}
The first author acknowledges the Anusandhan National Research Foundation (ANRF), Government
of India, for supporting this work through the Prime Minister Early Career Research
Grant (PM-ECRG) under Grant No. ANRF/ECRG/2024/002135/PMS.

\section*{Declaration of generative AI and AI-assisted technologies in the manuscript preparation process}

During the preparation of this work the authors used Gemini 3.1 Pro and Claude Sonnet 5 in order to assist with code implementation and debugging, organization of the manuscript, and language and grammar checks of the manuscript. After using this tool/service, the authors reviewed and edited the content as needed and take full responsibility for the content of the published article.

\bibliographystyle{siamplain}
\bibliography{references}

@inproceedings{rahaman2019spectral,
  title={On the spectral bias of neural networks},
  author={Rahaman, Nasim and Baratin, Aristide and Arpit, Devansh and Draxler, Felix and Lin, Min and Hamprecht, Fred and Bengio, Yoshua and Courville, Aaron},
  booktitle={International conference on machine learning},
  pages={5301--5310},
  year={2019},
  organization={PMLR}
}

@article{tancik2020fourier,
  title={Fourier features let networks learn high frequency functions in low dimensional domains},
  author={Tancik, Matthew and Srinivasan, Pratul and Mildenhall, Ben and Fridovich-Keil, Sara and Raghavan, Nithin and Singhal, Utkarsh and Ramamoorthi, Ravi and Barron, Jonathan and Ng, Ren},
  journal={Advances in neural information processing systems},
  volume={33},
  pages={7537--7547},
  year={2020}
}

@article{li2026novel,
  title={A novel singularity-and discontinuity-capturing PINN for time-fractional diffusion equations involving initial singularities and interfaces on complex curved surfaces},
  author={Li, Hongji and Tan, Zhijun},
  journal={Applied Mathematics and Computation},
  volume={519},
  pages={129917},
  year={2026},
  publisher={Elsevier}
}

@article{wang2021understanding,
  title={Understanding and mitigating gradient flow pathologies in physics-informed neural networks},
  author={Wang, Sifan and Teng, Yujun and Perdikaris, Paris},
  journal={SIAM Journal on Scientific Computing},
  volume={43},
  number={5},
  pages={A3055--A3081},
  year={2021},
  publisher={SIAM}
}

@article{stynes2017error,
  title={Error analysis of a finite difference method on graded meshes for a time-fractional diffusion equation},
  author={Stynes, Martin and O'Riordan, Eugene and Gracia, Jos{\'e} Luis},
  journal={SIAM Journal on Numerical Analysis},
  volume={55},
  number={2},
  pages={1057--1079},
  year={2017},
  publisher={SIAM}
}

@book{berkolaiko2013introduction,
  title={Introduction to Quantum Graphs},
  author={Berkolaiko, Gregory and Kuchment, Peter},
  volume={186},
  year={2013},
  publisher={American Mathematical Society},
  address={Providence, RI}
}

@article{lagaris1998artificial,
  title={Artificial neural networks for solving ordinary and partial differential equations},
  author={Lagaris, Isaac E and Likas, Aristidis and Fotiadis, Dimitrios I},
  journal={IEEE transactions on neural networks},
  volume={9},
  number={5},
  pages={987--1000},
  year={1998},
  publisher={IEEE}
}

@article{sukumar2022exact,
  title={Exact imposition of boundary conditions with distance functions in physics-informed deep neural networks},
  author={Sukumar, Natarajan and Srivastava, Ankit},
  journal={Computer Methods in Applied Mechanics and Engineering},
  volume={389},
  pages={114333},
  year={2022},
  publisher={Elsevier}
}

@article{paszke2019pytorch,
  title={Pytorch: An imperative style, high-performance deep learning library},
  author={Paszke, Adam and Gross, Sam and Massa, Francisco and Lerer, Adam and Bradbury, James and Chanan, Gregory and Killeen, Trevor and Lin, Zeming and Gimelshein, Natalia and Antiga, Luca and others},
  journal={Advances in neural information processing systems},
  volume={32},
  year={2019}
}

@article{lin2007finite,
  title={Finite difference/spectral approximations for the time-fractional diffusion equation},
  author={Lin, Yumin and Xu, Chuanju},
  journal={Journal of computational physics},
  volume={225},
  number={2},
  pages={1533--1552},
  year={2007},
  publisher={Elsevier}
}

@misc{saketgithub,
  author       = {S. Ramchandra and Vaibhav Mehandiratta},
  title        = {{QGPINN}s-Framework-to-solve-FDEs-on-Quantum-Graphs},
  year         = {2026},
  publisher    = {GitHub},
  url          = {https://github.com/Saket2006/QGPINNs-Framework-to-solve-FDEs-on-Quantum-Graphs},
  note         = {GitHub repository}
}

@article{nochetto2015pde,
  title={A {PDE} approach to fractional diffusion in general domains: a priori error analysis},
  author={Nochetto, R. H. and Ot{\'a}rola, E. and Salgado, A. J.},
  journal={Foundations of Computational Mathematics},
  volume={15},
  number={3},
  pages={733--791},
  year={2015},
  publisher={Springer}
}

@article{sophiya2025comprehensive,
  title={A comprehensive analysis of {PINN}s: Variants, applications, and challenges},
  author={Sophiya, A. A. and Nair, A. K. and Maleki, S. and Krishnababu, S. K.},
  journal={arXiv preprint arXiv:2505.22761},
  year={2025}
}

@article{Arioli2017FEM,
  author = {Arioli, M. and Benzi, M.},
  title = {A finite element method for quantum graphs},
  journal = {IMA J. Numer. Anal.},
  volume = {38},
  year = {2018},
  number = {3},
  pages = {1119--1163},
  issn = {0272-4979},
  doi = {10.1093/imanum/drx029}
}

@article{bolin2024gaussian,
  title={Gaussian {W}hittle--{M}at{\'e}rn fields on metric graphs},
  author={Bolin, D. and Simas, A. B. and Wallin, J.},
  journal={Bernoulli},
  volume={30},
  number={2},
  pages={1611--1639},
  year={2024},
  publisher={Bernoulli Society for Mathematical Statistics and Probability}
}

@article{mehandiratta2021optimal,
  title={Optimal control problems driven by time-fractional diffusion equations on metric graphs: optimality system and finite difference approximation},
  author={Mehandiratta, V. and Mehra, M. and Leugering, G.},
  journal={SIAM Journal on Control and Optimization},
  volume={59},
  number={6},
  pages={4216--4242},
  year={2021},
  publisher={SIAM}
}

@article{kovacs2021stochastic,
  title={Stochastic reaction--diffusion equations on networks},
  author={Kov{\'a}cs, M. and Sikolya, E.},
  journal={Journal of Evolution Equations},
  volume={21},
  number={4},
  pages={4213--4260},
  year={2021},
  publisher={Springer}
}

@article{bolin2026statistical,
  title={Statistical inference for {G}aussian {W}hittle--{M}at{\'e}rn fields on metric graphs},
  author={Bolin, D. and Simas, A. B. and Wallin, J.},
  journal={Journal of the Royal Statistical Society Series B: Statistical Methodology},
  year={2026},
  publisher={Oxford University Press UK}
}

@book{Berkolaiko2013,
  author = {Berkolaiko, G. and Kuchment, P.},
  title = {Introduction to quantum graphs},
  series = {Mathematical Surveys and Monographs},
  volume = {186},
  publisher = {American Mathematical Society, Providence, RI},
  year = {2013},
  pages = {xiv+270},
  isbn = {978-0-8218-9211-4},
  doi = {10.1090/surv/186}
}

@article{von1988classical,
  title={Classical solvability of linear parabolic equations on networks},
  author={Below, J. V.},
  journal={Journal of differential equations},
  volume={72},
  number={2},
  pages={316--337},
  year={1988},
  publisher={Elsevier}
}

@article{odvzak2019weyl,
  title={On the {W}eyl law for quantum graphs},
  author={Od{\v{z}}ak, A. and {\v{S}}{\'c}eta, L.},
  journal={Bulletin of the Malaysian Mathematical Sciences Society},
  volume={42},
  number={1},
  pages={119--131},
  year={2019},
  publisher={Springer}
}

@inproceedings{band2017quantum,
  title={Quantum graphs which optimize the spectral gap},
  author={Band, R. and L{\'e}vy, G.},
  booktitle={Annales Henri Poincar{\'e}},
  volume={18},
  pages={3269--3323},
  year={2017},
  organization={Springer}
}

@article{lagnese1993control,
  title={Control of planar networks of {T}imoshenko beams},
  author={Lagnese, J. E. and Leugering, G. and Schmidt, E.},
  journal={SIAM journal on control and optimization},
  volume={31},
  number={3},
  pages={780--811},
  year={1993},
  publisher={SIAM}
}

@article{mehandiratta2023well,
  title={Well-posedness, optimal control and discretization for time-fractional parabolic equations with time-dependent coefficients on metric graphs},
  author={Mehandiratta, V. and Mehra, M. and Leugering, G.},
  journal={Asian Journal of Control},
  volume={25},
  number={3},
  pages={2360--2377},
  year={2023},
  publisher={Wiley Online Library}
}

@article{kumari2025finite,
  title={Finite difference approximation of time-fractional advection-diffusion equation on a metric star graph},
  author={Kumari, S. and Mehra, M. and Mehandiratta, V.},
  journal={International Journal of Computer Mathematics},
  volume={102},
  number={12},
  pages={2032--2050},
  year={2025},
  publisher={Taylor \& Francis}
}

@article{bolin2026numerical,
  title={Numerical approximation of fractional diffusion equations on metric graphs},
  author={Bolin, D. and Riera-Segura, L. and Simas, A. B.},
  journal={arXiv preprint arXiv:2608.01932},
  year={2026}
}

@article{bolin2024regularity,
  title={Regularity and numerical approximation of fractional elliptic differential equations on compact metric graphs},
  author={Bolin, D. and Kov{\'a}cs, M. and Kumar, V. and Simas, A. B.},
  journal={Mathematics of Computation},
  volume={93},
  number={349},
  pages={2439--2472},
  year={2024}
}

@article{singh2024non,
  title={Non-local physics informed neural networks for forward and inverse problems containing non-local operators},
  author={Singh, A. K. and Mehra, M. and Pulch, R.},
  journal={Neural Comput. Appl},
  volume={37},
  pages={1--22},
  year={2024}
}

@article{pang2019fpinns,
  title={f{PINN}s: Fractional physics-informed neural networks},
  author={Pang, G. and Lu, L. and Karniadakis, G. E.},
  journal={SIAM Journal on Scientific Computing},
  volume={41},
  number={4},
  pages={A2603--A2626},
  year={2019},
  publisher={SIAM}
}

@article{zhao2022deep,
  title={A deep learning approach to solve forward differential problems on graphs},
  author={Zhao, Y. and Pasini, M. L.},
  journal={arXiv preprint arXiv:2210.03746},
  year={2022}
}

@article{laczko2025transferable,
  title={A transferable {PINN}-based method for quantum graphs with unseen structure},
  author={Laczk{\'o}, C. L. and V{\'a}ghy, M. A. and Kov{\'a}cs, M.},
  journal={IFAC-PapersOnLine},
  volume={59},
  number={1},
  pages={67--72},
  year={2025},
  publisher={Elsevier}
}

@article{mehandiratta2019existence,
  title={Existence and uniqueness results for a nonlinear {C}aputo fractional boundary value problem on a star graph},
  author={Mehandiratta, V. and Mehra, M. and Leugering, G.},
  journal={Journal of Mathematical Analysis and Applications},
  volume={477},
  number={2},
  pages={1243--1264},
  year={2019},
  publisher={Elsevier}
}

@article{sobirov2021green,
  title={Green's function method for time-fractional diffusion equation on the star graph with equal bonds},
  author={Sobirov, Z. and Rakhimov, K. U. and Ergashov, R. E.},
  journal={Nanosystems: Physics, Chemistry, Mathematics},
  volume={12},
  number={3},
  pages={271--278},
  year={2021}
}

@article{faheem2023collocation,
  title={A collocation method for time-fractional diffusion equation on a metric star graph with $\eta$ edges},
  author={Faheem, M. and Khan, A.},
  journal={Mathematical Methods in the Applied Sciences},
  volume={46},
  number={8},
  pages={8895--8914},
  year={2023},
  publisher={Wiley Online Library}
}

@article{wang2024local,
  title={Local error analysis of {L}1 scheme for time-fractional diffusion equation on a star-shaped pipe network},
  author={Wang, J. and Yu, Y. and Hou, J. and Meng, X.},
  journal={Physica Scripta},
  volume={99},
  number={12},
  pages={125274},
  year={2024},
  publisher={IOP Publishing}
}

@article{vats2026time,
  title={Time-fractional cable equation for anomalous electrodiffusion on a metric star graph with existence, uniqueness and numerical solution},
  author={Vats, Y. and Mehra, M. and Oelz, D.},
  journal={Mathematics and Computers in Simulation},
  year={2026},
  publisher={Elsevier}
}

@article{mehandiratta2020difference,
  title={A difference scheme for the time-fractional diffusion equation on a metric star graph},
  author={Mehandiratta, Vaibhav and Mehra, Mani},
  journal={Applied Numerical Mathematics},
  volume={158},
  pages={152--163},
  year={2020},
  publisher={Elsevier}
}

@article{guo2022monte,
  title={Monte {C}arlo f{PINN}s: {D}eep learning method for forward and inverse problems involving high dimensional fractional partial differential equations},
  author={Guo, L. and Wu, H. and Yu, X. and Zhou, T.},
  journal={Computer Methods in Applied Mechanics and Engineering},
  volume={400},
  pages={115523},
  year={2022},
  publisher={Elsevier}
}

@article{alikhanov2015new,
  title={A new difference scheme for the time fractional diffusion equation},
  author={Alikhanov, A. A.},
  journal={Journal of Computational Physics},
  volume={280},
  pages={424--438},
  year={2015},
  publisher={Elsevier}
}

@article{rumelhart1986learning,
  title={Learning representations by back-propagating errors},
  author={Rumelhart, D. E. and Hinton, G. E. and Williams, R. J.},
  journal={nature},
  volume={323},
  number={6088},
  pages={533--536},
  year={1986},
  publisher={Nature Publishing Group UK London}
}

@article{baydin2018automatic,
  title={Automatic differentiation in machine learning: a survey},
  author={Baydin, A. G. and Pearlmutter, B. A. and Radul, A. A. and Siskind, J. M.},
  journal={Journal of machine learning research},
  volume={18},
  number={153},
  pages={1--43},
  year={2018}
}

@article{liao2018second,
  title={A second-order scheme with nonuniform time steps for a linear reaction-sudiffusion problem},
  author={Liao, H. L. and McLean, W. and Zhang, J.},
  journal={Communications in Computational Physics},
  volume={30},
  pages={567-601},
  year={2021}
}

@article{yoshioka2014burgers,
  title={Burgers type equation models on connected graphs and their application to open channel hydraulics},
  author={Yoshioka, H. and Unami, K. and Fujihara, M.},
  journal={RIMS K\^{o}ky\^{u}roku},
  volume={1890},
  pages={160--171},
  year={2014},
  publisher={Research Institute for Mathematical Sciences, Kyoto University}
}

@book{mainardi2022fractional,
  title={Fractional calculus and waves in linear viscoelasticity},
  author={Mainardi, F.},
  year={2022},
  publisher={World Scientific Publishing Company}
}

\end{document}